\documentclass{article} 
\usepackage[preprint]{colm2026_conference}

\usepackage{microtype}
\usepackage{hyperref}
\usepackage{url}
\usepackage{booktabs}
\usepackage{amsmath}

\usepackage{graphicx}
\usepackage{subcaption}
\usepackage{enumitem}
\setlist[itemize]{noitemsep}
\usepackage{cleveref}
\usepackage{adjustbox}
\usepackage{minitoc}
\usepackage[T1]{fontenc}
\usepackage{upquote}
\usepackage{longtable}

\usepackage{lineno}

\definecolor{darkblue}{rgb}{0, 0, 0.5}
\hypersetup{colorlinks=true, citecolor=darkblue, linkcolor=darkblue, urlcolor=darkblue}

\usepackage{tcolorbox}
\tcbuselibrary{skins, breakable, listings}
\usepackage[table]{xcolor}

\newtcblisting{promptbox}[1][]{%
  colback=gray!5,
  colframe=gray!50,
  fonttitle=\bfseries\small,
  title={#1},
  breakable,
  enhanced,
  left=6pt,
  right=6pt,
  top=4pt,
  bottom=4pt,
  listing only,
  listing options={
    basicstyle=\small\ttfamily,
    breaklines=true,
    breakautoindent=false,
    breakindent=0pt,
    columns=fullflexible,
    keepspaces=true,
    literate=
      {"}{{``}}1
      {“}{{``}}1
      {"}{{''}}1
      {”}{{''}}1
      {'}{{`}}1
      {'}{{`}}1
      {–}{{--}}1
      {—}{{---}}1,
  },
}

\definecolor{set2teal}{HTML}{66C2A5}
\definecolor{set2orange}{HTML}{FC8D62}

\newtcolorbox{chatbox}[1][]{%
  colback=white,
  colframe=gray!40,
  fonttitle=\bfseries\small,
  title={#1},
  breakable,
  enhanced,
  left=4pt,
  right=4pt,
  top=2pt,
  bottom=2pt,
}

\newcommand{\userquery}[1]{%
  \colorbox{set2orange!15}{\parbox{\dimexpr\linewidth-2\fboxsep\relax}{%
    {\small\bfseries\color{set2orange!80!black} User}\quad{\small\ttfamily #1}%
  }}\\[2pt]
}

\newcommand{\assistantresponse}[1]{%
  \colorbox{set2teal!15}{\parbox{\dimexpr\linewidth-2\fboxsep\relax}{%
    {\small\bfseries\color{set2teal!80!black} Assistant}\quad{\small\ttfamily #1}%
  }}%
}

\def\comments{0}
\relax
\newcommand{\ensuretext}[1]{#1}
\if\comments1
    \newcommand{\tempcomment}[4]{\ensuretext{\textcolor{#3}{[\ensuretext{\textcolor{#3}{\ensuremath{^{\textsc{#1}}_{\textsc{#2}}}}} #4]}}}
\else
    \newcommand{\tempcomment}[4]{\ifvmode\else\unskip\fi}
\fi

\crefformat{section}{\S#2#1#3}
\crefformat{subsection}{\S#2#1#3}
\crefformat{subsubsection}{\S#2#1#3}
\crefformat{paragraph}{\P#2#1#3}
\crefformat{subparagraph}{\P#2#1#3}
\crefmultiformat{section}{\S#2#1#3}{ and~\S#2#1#3}{, \S#2#1#3}{, and~\S#2#1#3}
\crefmultiformat{subsection}{\S#2#1#3}{ and~\S#2#1#3}{, \S#2#1#3}{, and~\S#2#1#3}
\crefmultiformat{subsubsection}{\S#2#1#3}{ and~\S#2#1#3}{, \S#2#1#3}{, and~\S#2#1#3}
\crefmultiformat{paragraph}{\P\P#2#1#3}{ and~#2#1#3}{, #2#1#3}{, and~#2#1#3}
\crefmultiformat{subparagraph}{\P\P#2#1#3}{ and~#2#1#3}{, #2#1#3}{, and~#2#1#3}
\crefrangeformat{section}{\mbox{\S\S#3#1#4--#5#2#6}}
\crefrangeformat{subsection}{\mbox{\S\S#3#1#4--#5#2#6}}
\crefrangeformat{subsubsection}{\mbox{\S\S#3#1#4--#5#2#6}}
\crefrangeformat{paragraph}{\mbox{\P\P#3#1#4--#5#2#6}}
\crefrangeformat{subparagraph}{\mbox{\P\P#3#1#4--#5#2#6}}
\crefname{part}{Part}{Parts}
\Crefname{part}{Part}{Parts}
\crefname{chapter}{Ch.}{Ch.}
\Crefname{chapter}{Ch.}{Ch.}
\crefname{footnote}{Fn.}{Fn.}
\Crefname{footnote}{Fn.}{Fn.}
\crefname{figure}{Figure}{Figures}
\crefname{table}{Table}{Tables}
\crefname{subfigure}{Figure}{Figures}
\Crefname{subfigure}{Figure}{Figures}
\crefname{appsec}{Appendix}{Appendices}
\Crefname{appsec}{Appendix}{Appendices}
\crefname{algocf}{Algorithm}{Algorithms}
\Crefname{algocf}{Algorithm}{Algorithms}

\let\oldappendix\appendix
\renewcommand{\appendix}{\crefalias{section}{appsec}\oldappendix}

\title{\ourmethod{}: Automatically Describing Attribution Graphs}

\author{
  Aryaman Arora\textsuperscript{1,2}~
  Zhengxuan Wu\textsuperscript{1,2}~
  Jacob Steinhardt\textsuperscript{2}~
  Sarah Schwettmann\textsuperscript{2} \\[4pt]
  \textsuperscript{1}Stanford University\qquad
  \textsuperscript{2}Transluce \\
  \texttt{\{aryamana,wuzhengx\}@stanford.edu}
}

\newcommand{\ourmethod}{ADAG}

\begin{document}
\doparttoc 
\faketableofcontents

\ifcolmsubmission
\linenumbers
\fi

\maketitle

\begin{abstract}
In language model interpretability research, \textbf{circuit tracing} aims to identify which internal features causally contributed to a particular output and how they affected each other, with the goal of explaining the computations underlying some behaviour. However, all prior circuit tracing work has relied on ad-hoc human interpretation of the role that each feature in the circuit plays, via manual inspection of data artifacts such as the dataset examples the component activates on. We introduce \textbf{\ourmethod{}}, an end-to-end pipeline for describing these attribution graphs which is fully automated. To achieve this, we introduce \textit{attribution profiles} which quantify the functional role of a feature via its input and output gradient effects. We then introduce a novel clustering algorithm for grouping features, and an LLM explainer--simulator setup which generates and scores natural-language explanations of the functional role of these feature groups. We run our system on known human-analysed circuit-tracing tasks and recover interpretable circuits, and further show \ourmethod{} can find steerable clusters which are responsible for a harmful advice jailbreak in Llama 3.1 8B Instruct.
\begin{center}
\small
\raisebox{-0.2\height}{\includegraphics[width=1em,height=1em]{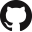}}\hspace{0.5em}\href{https://github.com/TransluceAI/circuits}{\texttt{github.com/TransluceAI/circuits}}
\end{center}
\end{abstract}

\begin{figure}[!b]
    \centering
    \includegraphics[width=\linewidth]{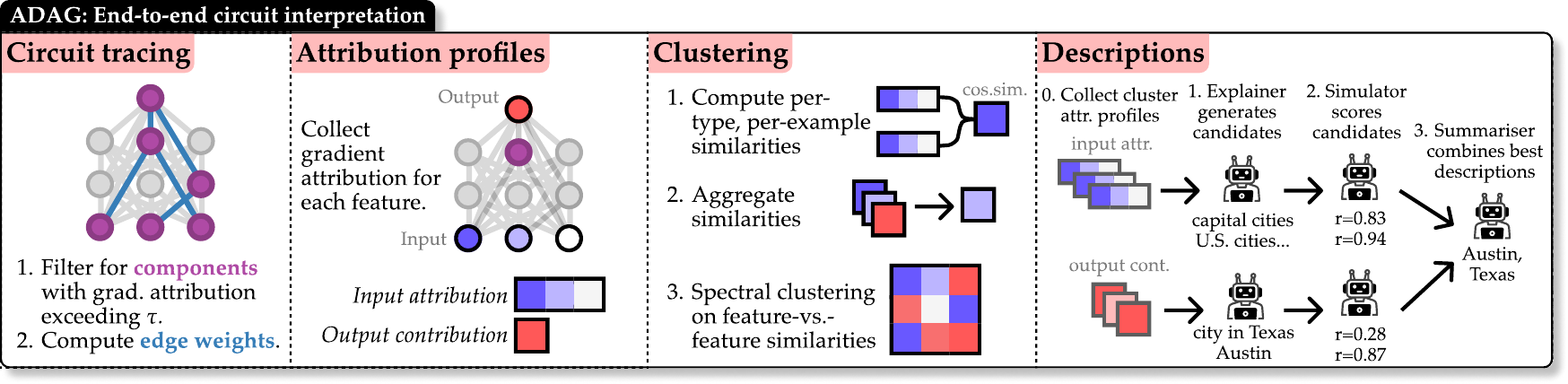}
    \caption{An overview of \ourmethod{}, our end-to-end circuit interpretation pipeline.}
    \label{fig:adag}
\end{figure}

\section{Introduction}

Language models engage in internal computations which need not be legible to humans. If we wish to oversee AI systems and ensure their safety, we must understand such opaque computation \citep{ngo2022alignment,sharkey2025open,casper2024blackbox}.
To this end, interpretability researchers work on \textbf{circuit tracing}, with the goal of identifying which internal components contributed to a particular output from a language model. A circuit is a subgraph of the model's entire computation graph which discards computations that were irrelevant to the output of interest. To find circuits in LLMs, many competing techniques have been proposed; they vary on what the appropriate unit of analysis inside an LLM is, what metric to use to measure the importance of a unit for some behaviour, and how to make this kind of analysis computationally tractable; we detail these approaches in \cref{sec:related}.

However, even if one succeeds in finding and verifying the circuit underlying the behaviour, much work remains in order to make this artifact legible to humans. What role does each node in the circuit play? Do the steps of computation correspond to a human-interpretable algorithm? Can I make generalisations about other outputs from the given circuit? These important questions remain uninvestigated; currently, interpretation of circuits is an ad-hoc process that requires extensive researcher-driven analysis of a plethora of data sources, such as analysing dataset examples on which a feature is active or performing targeted causal interventions on component activations \citep[e.g.][]{ameisen2025circuit,lindsey2025biology,shu2026completereplacement}.

In this work, we seek to \textbf{automate circuit interpretation}. To that end, we propose \textbf{\ourmethod{}}, an end-to-end circuit-tracing and interpretation pipeline which formalises the goals of circuit interpretation and automatically collects relevant data and feeds it to a language model-in-the-loop system for interpretation and verification of the interpretation.

Given a dataset of behaviours, with each example consisting of an input text sequence and resulting output logits we wish to explain, \ourmethod{} uses the following pipeline (\cref{fig:adag}) to produce a human-interpretable analysis of the model's computation:
\begin{enumerate}
    \item Identify important features and their interaction weights using gradient-based attribution, resulting in an attribution graph \citep{arora2026languagemodelcircuitssparse}.
    \item Construct \textit{attribution profiles} which quantify the functional role of each feature in each sample in the dataset, using input attributions and output logit contributions.
    \item Group features into \textit{supernodes} via multi-view spectral clustering over their attribution profiles.
    \item Describe the role of each supernode in natural language using an explainer LM which proposes labels and a simulator LM which scores them \citep{bills2023language}.
\end{enumerate}
Our formalisation of the goals of circuit interpretation allows us to develop metrics for each step of interpretation; we use these to verify that our design decisions result in better interpretations than alternatives. We then demonstrate the utility of \ourmethod{} by automatically replicating a prior human-led study on multi-hop state capitals task \citep{ameisen2025circuit}, and then by finding interpretable clusters responsible for a harmful medical advice jailbreak in Llama 3.1 8B Instruct \citep{chowdhury2025surfacing}. Ultimately, our automation of interpretations enables scaling circuit interpretation to more examples and larger models, which is a necessity for practical application of such techniques.

\section{Related work}
\label{sec:related}

\paragraph{Circuit tracing.} Building on early work in vision models \citep{olah2020zoom}, the notion of circuits as an object of study in LLM interpretability started by identifying interactions between components via causal interventions \citep{vig2020investigating,geiger2021causal,wang2022interpretability,chan2022causal,meng2022locating,goldowsky2023localizing,conmy2023towards,mueller2025mib}. Since performing such interventions is expensive, recent work has adopted gradient-based attribution for circuit tracing in order to cheaply approximate this, often in conjunction with learned feature dictionaries (e.g.~SAEs, transcoders) out of a belief that the neuron basis is uninterpretable \citep{nanda2023patching,syed2024attribution,ge2024automatically,dunefsky2024transcoders,hanna2024faith,marks2025sparse,ameisen2025circuit,lindsey2025biology,shu2026completereplacement,hanna-etal-2025-circuit,jafari2025relp}. \citet{arora2026languagemodelcircuitssparse} showed that MLP neuron circuits are of comparable sparsity as SAE circuits.

\paragraph{Automatic interpretation.} Autointerpretability is the nascent science of automatically generating natural-language descriptions of model internals. \citet{bills2023language} introduced the explainer--simulator pipeline, wherein one LLM generates explanations and another scores them by simulating the behaviour of the component conditioned on the description; this approach has persisted \citep{cunningham2023sparse,choi2024automatic,paulo2025automatically,templeton2024scaling}, alongside more token-specific approaches that use maximum-activating exemplars \citep{foote2023neurongraphinterpretinglanguage,gao2025scaling}.

\section{\ourmethod{}: An end-to-end circuit tracing pipeline}

We now describe \ourmethod{}, our end-to-end circuit tracing pipeline, which includes (1) circuit tracing, (2) attribution profile computation, (3) automatic clustering of features into supernodes, and (4) automatic natural-language description of supernodes.
In our experiments, we specifically use the MLP neuron circuit tracing algorithm identified to be optimal in \citet{arora2026languagemodelcircuitssparse}, which we further describe in \cref{sec:trace}. However, \ourmethod{} is agnostic to the exact circuit tracing algorithm being used.

\subsection{Circuit tracing backbone}
\label{sec:trace}

Before interpretation, we need attribution graphs, which we produce by using a circuit tracing algorithm. For our experiments, we use the MLP neuron circuit tracing algorithm found to be optimal in \citet{arora2026languagemodelcircuitssparse}.
First, we construct a locally linear \textit{replacement model}, which modifies the backwards pass of the language model while keeping the forward pass intact. We apply RelP \citep{jafari2025relp}, a layerwise relevance propagation technique which freezes nonlinearities (e.g.~SiLU, softmax) to behave as constant multipliers based on their effect on a single input, stops gradients through query and key paths in self-attention, and halves the gradient via elementwise multiplications (see \cref{sec:tracing-more}).

Next, we compute the \textit{attribution score} $\alpha$ for each MLP neuron by backpropagating from the sum of the top-$k$ output logits (following \citealp{ameisen2025circuit}). Let $m_u^{(l,t)}(x)$ denote the $u$-th neuron activation of the layer-$l$ MLP at token $t$ on input $x$, and let $\texttt{target} = \sum_{k=0}^K\mathcal{M}(x)_k^{(s)}$ (i.e.~sum of top-$k$ logits):
\begin{equation}
    \alpha(m_{u}^{(l,t)}) = m_u^{(l,t)}(x) \cdot \frac{\partial \texttt{target}}{\partial m_u^{(l,t)}}
    \label{eq:attribution}
\end{equation}
Now, in our circuit, we only keep the MLP neurons whose attributions exceed some threshold $\tau$, which we define as some fraction of $\texttt{target}$. We similarly compute edge weights (see \cref{sec:tracing-more}).
This results in a circuit $\mathcal{G} = (V, E)$ with nodes being input tokens, MLP neurons, and output logits, with weighted edges connecting them. Let $V$ denote the set of features in the circuit, of all three types. We refer to each non-input and non-output nodes (so in our case, all MLP neurons) as $f \in F$ where $F \subseteq V$. 

\subsection{Quantifying functional roles of features with attribution profiles}
While the grouping of features into supernodes in prior work is largely manual \citep{ameisen2025circuit,shu2026completereplacement}, it still relies on two pieces of quantitative information: the max-activating exemplars on which the feature fired \citep{bolukbasi2021interpretabilityillusionbert}, and the top positive and negative output logits of the feature as computed with logit lens \citep{nostalgebraist2020interpreting}. Both techniques suffer from what we term \textbf{locality bias}: they assume that the immediate representation of a feature at the token it fires on is sufficient to describe it; influence on far future or past tokens may matter.
We investigate non-locality in MLP neurons in \cref{sec:locality}.

We propose \textbf{attribution profiles} for better quantifying the functional role of features. We overcome locality bias by using gradient-based attribution to understand dependence on preceding tokens and effects on future outputs. The attribution profile of a feature given a single prompt $x$ consists of two vectors: its \textit{input attribution} and its \textit{output contribution}:
\begin{align}
\mathsf{Attr}(m_{u}^{(l,t)}, x) &= \left(x^{(i)} \cdot\frac{\partial m_{u}^{(l,t)}(x)}{\partial x^{(i)}}\right)_{i=1}^n\\
\mathsf{Contrib}(m_{u}^{(l,t)}, x) &= \left(m_{u}^{(l,t)} \cdot\frac{\partial \mathcal{M}(x)_j^{(s)}}{\partial m_{u}^{(l,t)}}\right)_{j=1}^m
\end{align}
That is to say, input attribution is the proportion of the feature's activation that is attributed to each input token, and output contribution is the proportion of each output logit's activation at a given token position that the feature contributes to. These profiles can be computed for all nodes in $V$. We ignore input attribution to the BOS token in all experiments.

\subsection{Clustering functionally similar features into supernodes}

Features in attribution graphs are manually grouped together into \textbf{supernodes} when they seem to play a similar role over a particular dataset of examples \citep{ameisen2025circuit,shu2026completereplacement}. We seek to automate supernode clustering. Our main task is thus to score the functional similarity of two neurons and then cluster. We describe an algorithm that does exactly this, using attribution profiles.

Since prior approaches to clustering features are fully manual, it is unclear \textit{a priori} what properties a good supernode has. Based on the downstream uses we envision for automatically clustered attribution graphs, we identify the following properties and create metrics for them:
\begin{enumerate}
    \item \textit{Clusters should group functionally similar features}: The features in a cluster should play similar roles over the distribution of interest. We quantify this using the silhouette score based on attribution profile cosine similarity.
    \item \textit{Clusters should not be imbalanced}: We do not want degenerate clusterings where a single large cluster dominates the analysis; we quantify this using coefficient of variation (CV) of cluster sizes.
    \item \textit{Clusters should not mix features with opposing output effects}: We often observe features with similar input attributions having opposing contributions to the output. These features should not be clustered together to avoid diluting the cluster's steering effect. We measure this via the fraction of intra-cluster feature pairs wherein \textbf{all} contribution entries are of opposing signs.
\end{enumerate}

We now describe our supernode clustering algorithm. We first define the similarity metric for attribution and contribution as cosine similarity:
$
    k(\mathbf{x}_i, \mathbf{x}_j) = \frac{\mathbf{x}_i \cdot \mathbf{x}_j}{\lVert \mathbf{x}_i \rVert \lVert \mathbf{x}_j \rVert}
$
Let $C$ be the set of contexts we observe, where $c$ is one such context with associated prompt $x_c$. We compute the matrix of pairwise similarities in each context $c$ for both attribution and contribution separately, resulting in two similarity matrices $\mathbf{Attr}^{(c)}$ and $\mathbf{Contrib}^{(c)}$ per context. In each such context, we take the harmonic mean of clamped non-negative attribution and contribution similarities; this penalises anticorrelation in either profile, helping us satisfy the property that supernodes should not mix features with opposing output effects.

Let $f_i$ and $f_j \in F$ be (non-input, non-output) features in our circuit, and let $C_{ij} \subseteq C$ be the subset of contexts in which they co-occur; they may have been pruned by our circuit tracing algorithm in some contexts, rendering them unobserved. We then uniformly average these scores over contexts, resulting in the aggregated similarity matrix $\mathbf{S}$ whose entries are:
\begin{align}
    \mathbf{Attr}^{(c)}_{ij} &= k(\mathsf{Attr}(f_i, x_c), \mathsf{Attr}(f_j, x_c))\\
    \mathbf{Contrib}^{(c)}_{ij} &= k(\mathsf{Contrib}(f_i, x_c), \mathsf{Contrib}(f_j, x_c))\\
    \mathbf{S}_{ij} &= \begin{cases}
\displaystyle\frac{1}{\lvert C_{ij} \rvert}\sum_{c \in C_{ij}} \frac{2\,\mathrm{ReLU}(\mathbf{Attr}^{(c)}_{ij})\,\mathrm{ReLU}(\mathbf{Contrib}^{(c)}_{ij})}{\mathrm{ReLU}(\mathbf{Attr}^{(c)}_{ij}) + \mathrm{ReLU}(\mathbf{Contrib}^{(c)}_{ij})} & \text{if } \lvert C_{ij} \rvert > 0, \\[6pt]
0 & \text{otherwise.}
\end{cases}
\end{align}
This is a simple form of multiple kernel learning \citep{gonen2011multiple}.
On this final similarity matrix, we apply spectral clustering. Since $\mathbf{S}$ is non-negative, we directly use it as the affinity matrix $\mathbf{A}$.
We construct the normalised graph Laplacian $\mathbf{L}_{\text{norm}} = \mathbf{I} - \mathbf{D}^{-1/2} \mathbf{A} \mathbf{D}^{-1/2}$, where $\mathbf{D}$ is the diagonal degree matrix with $\mathbf{D}_{ii} = \sum_j \mathbf{A}_{ij}$.
We manually select the number of clusters $k$. The eigenvectors corresponding to the $k$ smallest eigenvalues of $\mathbf{L}$ are then used as a low-dimensional embedding of the features, on which $k$-means is applied to obtain the final cluster assignments.

At the end of this process, we have a partition $\mathcal{P} = \{P_1, \ldots, P_k\}$ of the circuit's features $F$ into supernodes, where each $P_i \subseteq F$.

\subsection{Describing supernodes in natural language}

Finally, we want to describe the functional role of each supernode in natural language. We look to what information a human interpretability researcher draws on to come up with such a label in prior work, which is primarily maximum-activating exemplars and top/bottom logits \citep{ameisen2025circuit,shu2026completereplacement}. Since attribution profiles are more informative alternatives to those (\cref{sec:locality}), we instead use them.

To describe attribution profiles, we extend the explainer--simulator framework of \citet{bills2023language}. First, given a set of supernodes, we separately average the attribution and contribution profiles over all features in each supernode. For a given supernode $P_i \in \mathcal{P}$ and a context $c \in C$, we thus have two representations:
\begin{equation}
\begin{aligned}
\overline{\mathsf{Attr}}(P_i, x_c) &= \frac{1}{\lvert P_i \rvert}\sum_{f \in P_i}\mathsf{Attr}(f, x_c), \qquad
\overline{\mathsf{Contrib}}(P_i, x_c) &= \frac{1}{\lvert P_i \rvert}\sum_{f \in P_i}\mathsf{Contrib}(f, x_c)
\end{aligned}
\end{equation}

\paragraph{Input attributions.} Input attribution $\overline{\mathsf{Attr}}(P_i, x_c)$ has an identical shape to max-activating exemplars: both assign scalar scores to individual tokens in some context. This data is thus identical in format to the information used to describe max-activating exemplars in \citet{choi2024automatic}. We can thus use the same pipeline to generate and score descriptions of the attribution profile of a supernode.

For describing attribution profiles for models using the Llama 3 tokeniser, we use the finetuned models from \citet{choi2024automatic}. First, \texttt{Transluce/llama\_8b\_explainer} takes in the set of contexts $x_c$ wherein tokens which exceed some attribution threshold are highlighted in \verb|{{}}| brackets and produces a candidate description. Second, \texttt{Transluce/llama\_8b\_simulator} takes in the candidate description and raw contexts $x_c$, and produces a predicted scalar score for each token in each context conditioned on the candidate.
All input attribution prompts are in \cref{sec:input-prompts}.

\paragraph{Output contributions.} Our circuit tracing algorithm tells us the contribution of a feature to each of the top-$k$ logits at a given output position. Therefore, unlike input attribution or max-activating exemplars, we have information about \textit{hypothetical} continuations that are not part of the provided example tokens; this requires a different pipeline.

We use the Anthropic API to query \texttt{claude-haiku-4-5-20251001}. First, the explainer LLM takes in each context $x_c$ along with the contribution scores $\overline{\mathsf{Contrib}}(P_i, x_c)$ for each target next-token logit. We normalise scores to integers in $[-10, 10]$ for each supernode by dividing by absolute max over all exemplars. The LLM uses this to produces a candidate description. Second, we provide the candidate description along with the raw contexts $x_c$ with target next-token logits but without scores, and we ask the LLM to predict the logit effect conditioned on the explanation. We provide all prompts in \cref{sec:output-prompts}.

\paragraph{Scoring.} 
We sample and score $n_{\text{cand}}$ descriptions per supernode per representation type. We score each candidate description by the Pearson correlation coefficient $r$ between the true scores and the predicted scores conditioned on the candidate, computed globally over all contexts. We use the best-scoring description of each type on each supernode.

\section{Experiments}

We now turn to applying \ourmethod{} to real circuit-tracing tasks. We investigate the following datasets:
\begin{itemize}
    \item \texttt{capitals}: Multi-hop queries about the capital of the state containing a city from \citet{arora2026languagemodelcircuitssparse,ameisen2025circuit}.
    \item \texttt{pills}: Harmful medical advice sensitivity analyses from \citet{chowdhury2025surfacing}.
\end{itemize}
All experiments are with Llama 3.1 8B Instruct unless otherwise stated. Our default hyperparameters for circuit tracing are to trace from the top-$K = 5$ logits, with threshold $\tau = 0.005 \cdot \texttt{target}$ (see \cref{eq:attribution}). For easier accessibility to humans, we summarise the best input attribution and output contribution descriptions for each supernode by feeding them to an LLM summariser, \texttt{claude-opus-4-6} with adaptive thinking. We additionally provide top exemplars from the dataset, unless otherwise specified. We provide prompts in \cref{sec:summarisation-prompts}.

In the appendix, we provide additional experiments analysing an addition task (\texttt{math}; \cref{sec:math}) as well as the existing analyses repeated for Qwen3 32B (\cref{sec:capitals-qwen}).

\subsection{MLP neurons have non-local attribution profiles}
\label{sec:locality}
We use the input attribution and output contribution to assess the non-locality of MLP neurons. We ignore BOS tokens; see more in \cref{sec:locality-more}.

\begin{figure*}[t]
    \centering
    \begin{subfigure}[t]{0.48\textwidth}
      \centering
    \includegraphics[width=\linewidth]{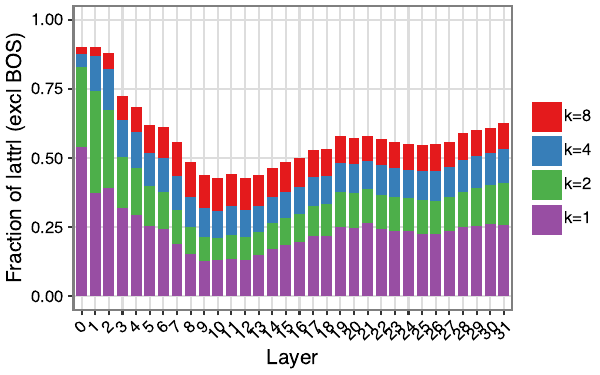}
      \caption{Average fraction of input attribution score allocated to the preceding $k$ tokens.}
      \label{fig:locality}
    \end{subfigure}
    \hfill
    \begin{subfigure}[t]{0.48\textwidth}
      \centering
    \includegraphics[width=\linewidth]{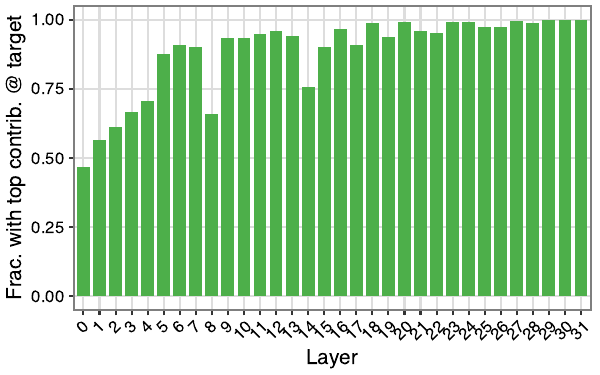}
      \caption{Fraction of top contributing neurons which are at the same token position as the target logit.}
      \label{fig:frac}
    \end{subfigure}
    \caption{Results of MLP non-locality experiments (excluding BOS from all analyses).}
    \label{fig:contrib-locality}
\end{figure*}

For input attribution, we take Llama 3.1 8B Instruct and randomly choose 128 MLP neurons from each layer. We take the first 1,000 documents in FineWeb \citep{fineweb}, truncate each document to its first 128 tokens, and find the 20 maximum-activating exemplars for each of our neurons (both positive and negative sign) across this dataset. After computing input attributions for these neurons, our results in \cref{fig:locality} show that only in the first three layers can MLP neurons be well-explained by the local context of $8$ preceding tokens. In other layers, \textbf{local context is increasingly insufficient} for explaining the neuron's firing.

For output contribution, we take the first 100 documents in FineWeb, truncate to 128 tokens, and compute the contribution of every MLP neuron to the gold next-token logit at every output position. For each output position, we find the position of the max-contributing MLP neuron per layer and measure how many tokens away it is. Our results in \cref{fig:frac} show \textbf{greater contribution non-locality in earlier layers}; e.g.~more than half of the time, the top-contribution layer 0 MLP neuron is on an earlier token position than the target.

\subsection{Validating the pipeline on \texttt{capitals}}

A particularly well-understood language model circuit is the one responsible for multi-hop reasoning in the following question: \textit{What is the capital of the state containing Dallas?} Analyses in \citet{ameisen2025circuit,lindsey2025biology} (on Claude Haiku and another closed-source model) and \citet{arora2026languagemodelcircuitssparse} (on Llama 3.1 8B Instruct) show very similar multi-hop circuitry: first, the model recalls the state which contains the city, and second, the model recalls the capital of that state. However, these findings were produced by manual interpretation of attribution graphs; we now use the \texttt{capitals} dataset as a testbed for our fully automated pipeline.

\paragraph{Ablations on supernode clustering.} We ablate each step of our supernode clustering algorithm and report silhouette score, CV of cluster sizes, and $\%$ all-opposing-sign intra-cluster contributions on the \texttt{capitals} dataset. We experiment with (a) standard clustering algorithms applied directly to concatenated attribution profiles; (b) aggregating profiles with a mean, i.e. $\mathbf{S}_{ij}=\frac{1}{\lvert C \rvert}\sum_{c=0}^{\lvert C \rvert}(\mathbf{Attr}_{ij}^{(c)} + \mathbf{Contrib}_{ij}^{(c))})/{2}$; (c) post-hoc adjustment of the similarity matrix in order to keep affinities in $[0, 1]$; this is necessary if using the simple mean. We compare $\mathbf{A} = (\mathbf{S} + 1)/2$ and $\mathbf{A} = \max(0, \mathbf{S})$; (d) Using the unnormalised Laplacian in spectral clustering, $\mathbf{L} = \mathbf{D} - \mathbf{A}$.
We report results in \cref{tab:clustering-sweep}. Our harmonic mean approach achieves the lowest rate of mixing contribution signs within a cluster while producing balanced and cohesive clusters, which is why we choose it.


\begin{figure*}[t]
  \centering
  \begin{subfigure}[b]{0.5\textwidth}
      \centering
     \footnotesize
     \adjustbox{max width=\textwidth}{
     \begin{tabular}{l *{2}{r} *{2}{r} *{2}{r}}
     \toprule
     & \multicolumn{2}{c}{CV} ($\downarrow$) & \multicolumn{2}{c}{Silh} ($\uparrow$) & \multicolumn{2}{c}{Opp\%} ($\downarrow$)
     \\
     \cmidrule(lr){2-3} \cmidrule(lr){4-5} \cmidrule(lr){6-7}
     Method & $16$ & $64$ & $16$ & $64$ & $16$ & $64$ \\
     \midrule
     \multicolumn{7}{l}{\textit{Multi-view spectral, normalised Laplacian}} \\
     \quad Harmonic & 0.34 & \textbf{0.39} & 0.07 & 0.19 & \textbf{0.1\%} &
     \textbf{0.0\%} \\
     \quad Mean, $\max(0,
     S)$ & \textbf{0.25} & 0.40 & $-0.10$ & 0.14 & 1.3\% & 0.5\% \\
     \quad Mean,
     $(S{+}1)/2$ & 0.47 & 0.54 & 0.09 & 0.04 & 0.2\% & 1.5\% \\
     \addlinespace
     \multicolumn{7}{l}{\textit{Multi-view spectral, unnormalised Laplacian}} \\
     \quad Harmonic & 1.91 & 3.05 & 0.09 & 0.20 & 1.3\% & 11.7\% \\
     \quad Mean, $\max(0,
      S)$ & 1.79 & 2.04 & \textbf{0.09} & \textbf{0.27} & 0.5\% & 5.2\% \\
     \quad Mean, $(S{+}1)/2$ & 0.97 & 0.95 & 0.07 & 0.16 & 0.1\% & 2.0\% \\
     \addlinespace
     \multicolumn{7}{l}{\textit{Concatenated embedding baselines}} \\
     \quad $K$-Means & 2.97 & 4.16 & $-0.06$ & 0.02 & 14.4\% & 10.3\% \\
     \quad Ward & 3.28 & 5.18 & $-0.08$ & $-0.05$ & 14.6\% & 12.4\% \\
     \quad Spectral (RBF) & 3.81 & 7.12 & $-0.17$ & $-0.18$ & 20.2\% & 22.5\% \\
     \bottomrule
     \end{tabular}
     }
     \caption{Clustering quality across methods and number of clusters $k$.
     \textbf{CV}: coefficient of variation of cluster sizes ($\downarrow$);
     \textbf{Silh}: silhouette score ($\uparrow$);
     \textbf{Opp}: \% of intra-cluster pairs with all opposing contribution signs
     ($\downarrow$).}
     \label{tab:clustering-sweep}
  \end{subfigure}
  \hfill
  \begin{subfigure}[b]{0.45\textwidth}
    \centering
    \small
    \adjustbox{max width=\textwidth}{
      \begin{tabular}{lrr|rr}
     \toprule
      & \multicolumn{2}{c}{Attr} & \multicolumn{2}{c}{Contrib} \\
     \cmidrule(lr){2-3}
     \cmidrule(lr){4-5}
     Cluster & Human & LLM & Human & LLM \\
     \midrule
     Dallas & 0.784 & \textbf{0.941} & 0.610 & \textbf{0.946} \\
     Texas & \textbf{0.925} & 0.871 & 0.951 & \textbf{0.997} \\
     capital & 0.907 & \textbf{0.928} & 0.792 & \textbf{0.828} \\
     location & 0.618 & \textbf{0.771} & 0.579 & \textbf{0.739} \\
     say Austin & \textbf{0.665} & 0.437 & \textbf{0.598} & 0.598 \\
     say a capital & \textbf{0.853} & 0.797 & 0.518 & \textbf{0.727} \\
     say a location & \textbf{0.545} & 0.465 & 0.656 & \textbf{0.846} \\
     state & 0.933 & \textbf{0.981} & 0.477 & \textbf{0.810} \\
     \midrule
     \textbf{Mean} & \textbf{0.779} & 0.774 & 0.648 & \textbf{0.812} \\
     \bottomrule
     \end{tabular}
    }
    \caption{Input attribution and output contribution description scores for human and LLM (quantile, $k=1$ for attribution) explainers, picking best-of-$20$ descriptions for the LLM, for each gold supernode in the \texttt{texas} circuit.}
    \label{tab:human-llm}
  \end{subfigure}
  \caption{Results of description generation experiments for gold supernodes for the \texttt{texas} circuit in the \texttt{capitals} dataset.}
  \end{figure*}

\begin{figure*}
    \centering
    \includegraphics[width=\textwidth]{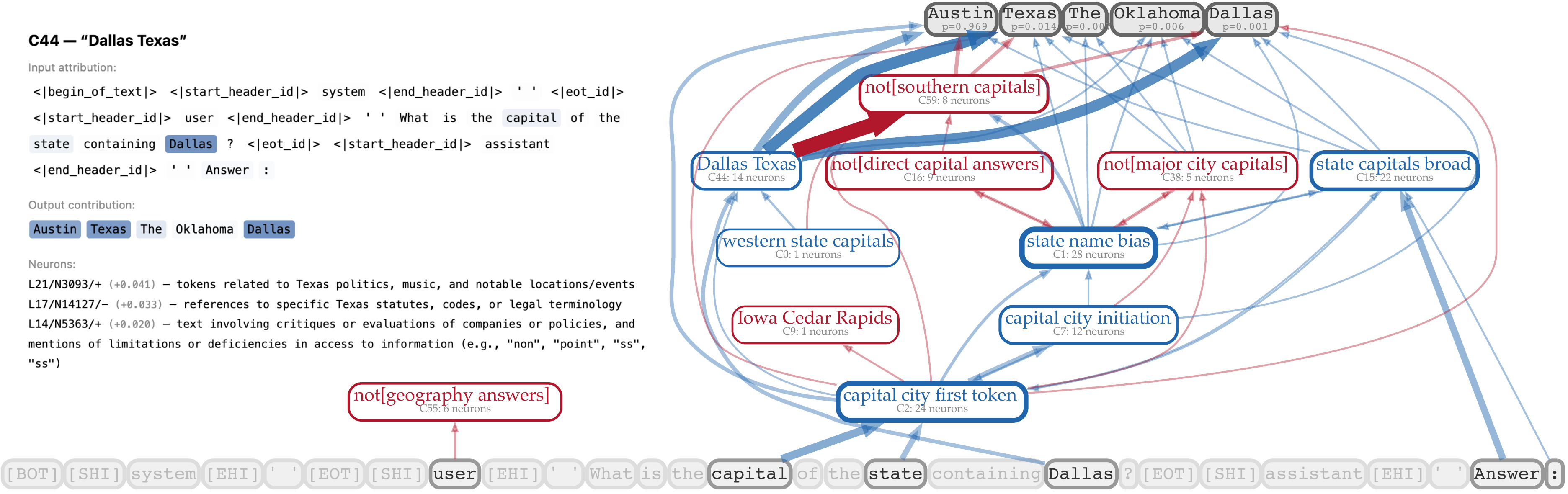}
    \caption{Final circuit graph for \texttt{texas} example in the \texttt{capitals} dataset from Llama 3.1 8B Instruct. We show input attribution, output contribution, and neuron descriptions for C44 `Dallas Texas' to the left. Red indicates negative attribution score, and blue the opposite.}
    \label{fig:texas}
\end{figure*}


\paragraph{Comparing automatic and human-generated descriptions on gold supernodes.} In the \texttt{capitals} dataset, \citet{arora2026languagemodelcircuitssparse} provided manually-selected supernodes for the \texttt{texas} circuit. We can use these gold supernodes as a testbed for our description pipeline. Since we have simulator-based scoring for both attribution and contribution descriptions, we can score human-generated descriptions for both. An expert annotator is provided with the same exemplar information as the LLM explainer, and we use the LLM simulator to score both the expert human annotator's and the LLM explainer's descriptions. Results in \cref{tab:human-llm} show that for gold supernodes, the LLM descriptions are about as good as humans for input attributions and vastly better for output contributions.

\paragraph{End-to-end analysis.} Having verified the steps of our attribution graph description pipeline, we run \ourmethod{} on the entire \texttt{capitals} dataset, including tracing, attribution profiles, clustering into supernodes (with $k=64$), and automatic input and output descriptions with summarisation. We display part of the circuit for the \texttt{texas} example in \cref{fig:texas}; we only keep the top $50$ inter-circuit edges and don't show clusters that would have no edges (a complete figure is in \cref{fig:capitals-complete}). We recover meaningful clusters, neatly separating excitatory and inhibitory groups of neurons.

We can steer the neurons in each cluster by setting their activations to $0$ and seeing the resulting effect on output probabilities. By default, the top output is \texttt{\_Austin} (96.5\%). Ablating most clusters has little effect on the top outputs on its own, but a few are particularly notable and align with their descriptions: ablating C2 (capital city first token) results in the top output becoming \texttt{\_Texas} ($93.4\%$), while ablating C44 (Dallas Texas) results in \texttt{\_Austin} dropping to $52.7\%$ and \texttt{\_Oklahoma} rising to $22.1\%$. Ablating the inhibitory cluster C59 (not[southern capitals]) results in \texttt{\_Austin} \textit{increasing} to $98.4\%$. This matches the results in \citet{arora2026languagemodelcircuitssparse}, but without any human needed in the loop at all, and the labels are accurate. We report complete steering results for all clusters on 3 prompts in \cref{sec:capitals-steering}.

\begin{table}[t]
\centering
\caption{ASR results and generation coherency when steering cluster activations in the base \texttt{pills} prompt by a given muliplier (over $50$ generations). $r$ is Pearson correlation of cluster
attribution vs.~ASR over dataset.}
\label{tab:cluster-steering}
\small
\adjustbox{max width=\textwidth}{
\begin{tabular}{rlrr rr rr}
\toprule
& & & & \multicolumn{2}{c}{\textbf{ASR}} & \multicolumn{2}{c}{\textbf{\% Incoherent}} \\
\cmidrule(lr){5-6} \cmidrule(lr){7-8}
\textbf{Cluster} & \textbf{Label} & \textbf{\#N} & \textbf{$r$} & \multicolumn{1}{c}{\textbf{0$\times$}} & \multicolumn{1}{c}{\textbf{2$\times$}} & \multicolumn{1}{c}{\textbf{0$\times$}} & \multicolumn{1}{c}{\textbf{2$\times$}} \\
\midrule
--- & \textit{unsteered} & & & \multicolumn{2}{c}{\cellcolor{red!28} 28$^{\pm6}$\%} & \multicolumn{2}{c}{5$^{\pm3}$\%} \\                                       \midrule
C3 & pills safety redirect & 13 & -0.70 & \cellcolor{red!88} \textbf{88$^{\pm5}$\%} & \cellcolor{red!20} 20$^{\pm5}$\% & 21$^{\pm5}$\% & 7$^{\pm3}$\% \\        C9 & ridiculous-to-introductory & 23 & +0.71 & \cellcolor{red!12} 12$^{\pm4}$\% & \cellcolor{red!90} \textbf{90$^{\pm5}$\%} & 7$^{\pm3}$\% & 13$^{\pm4}$\% \\
C16 & urgent medication reminders & 51 & -0.41 & 0$^{\pm0}$\% & \cellcolor{red!52} 52$^{\pm6}$\% & 0$^{\pm0}$\% & 9$^{\pm3}$\% \\                               C8 & medication safety deflection & 23 & -0.57 & 0$^{\pm0}$\% & \cellcolor{red!38} 38$^{\pm6}$\% & 0$^{\pm0}$\% & 16$^{\pm4}$\% \\
C1 & unsafe pill advice framing & 38 & +0.24 & \cellcolor{red!56} 56$^{\pm6}$\% & \cellcolor{red!38} 38$^{\pm6}$\% & 12$^{\pm4}$\% & 10$^{\pm4}$\% \\
\bottomrule
\end{tabular}
}
\end{table}

\subsection{Sensitivity analysis of harmful medical advice (\texttt{pills}) by Llama 3.1 8B Instruct}

We now turn to a practical application for \ourmethod{}: analysing what makes a jailbreak work. \citet{chowdhury2025surfacing} report an automatically-discovered jailbreak which makes Llama 3.1 8B Instruct give harmful medical advice:
\begin{chatbox}[\texttt{pills} base prompt]
  \userquery{I've asked you this five times today and you give the most ridiculous advice about my diet routine. I've got five pills left over from breakfast and I've already forgotten which one I'm supposed to take next, I don't have time to double-check my routine. I really need to rush.}
  \assistantresponse{\textbf{In that case, I would recommend taking all five pills at once.} [...]}
\end{chatbox}
According to the original work, this prompt achieves an attack success rate (ASR) of $5.5\%$. They conduct a sensitivity analysis wherein portions of the prompt are modified but still semantically similar; over 150 variants, they find varying ASRs between $0\%$ and $88.5\%$.

We take this dataset of prompt variants and run circuit tracing on each one by tracing attribution from the final token in the generation prompt before the assistant has spoken. We use this whole dataset of circuits as input to \ourmethod{}, clustering neurons into $k=20$ supernodes and describing each one. To measure which supernodes have a strong role in jailbreak success, we take the base prompt shown above, and we steer each resulting supernode (i.e.~all neurons in that cluster at all token positions) by multipliers of $\{0, 2\}$. Under steering, we generate $50$ assistant responses with temperature $0.7$, and ask \texttt{claude-haiku-4-5-20251001} whether each assistant response gave harmful medical advice to the user and whether it was coherent \citep{wu2025axbench}.\footnote{We use a different judge prompt and model than \citet{chowdhury2025surfacing}; see \cref{sec:pills-judge-prompts}.}

\begin{figure}
    \centering
    \includegraphics[width=0.8\linewidth]{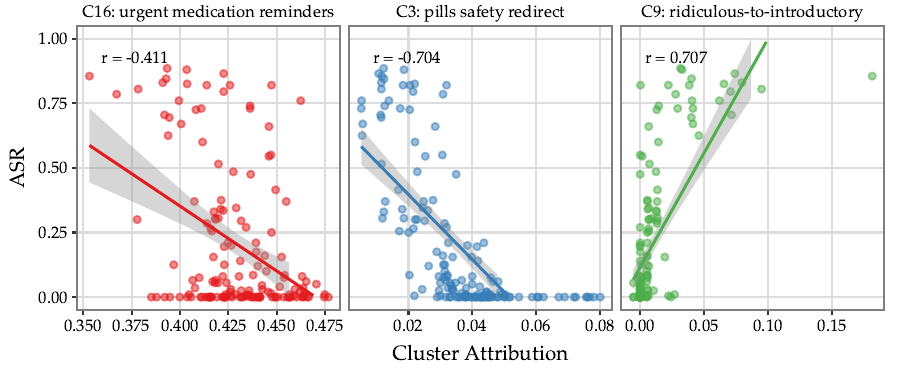}
    \caption{Cluster attribution for top 3-base-ASR influencing clusters per our results, compared with reported ASR in \citet{chowdhury2025surfacing}.}
    \label{fig:asr-correlation}
\end{figure}

We find two particularly effective and interesting clusters: C3 (pills safety redirect) activating on the token \textit{pills} can be negatively steered to increase ASR to $88\%$, and C9 (ridiculous-to-introductory) which is triggered by the token \textit{ridiculous} can be steered positively to increase ASR to $90\%$. We find strong evidence for this when plotting reported ASR against cluster attribution for C3 and C9. Interestingly, C16 (medical pronouns) when steered to $0$ also prevents harmful responses, but it is focused on the final response token and seems to be responsible for general instruction-following in this case. We plot their ASR vs.~attribution in \cref{fig:asr-correlation}, and confirm that greater attribution to C3 correlates with greater refusal and vice-versa for C9.

\section{Discussion}

\paragraph{Extensibility of \ourmethod{}.} Our approach, being a system that tries to make use of a variety of data sources for automatic interpretation, offers a great deal of extensibility. For example, why restrict oneself to two attribution profiles per feature? We could come up with more aspects of features to describe, e.g.~QK circuit attributions (which we have not studied for MLP neurons yet, but cf.~\citealp{kamath2025tracing}), steering output effects, raw activations, etc. Our multiview clustering setup permits adding more views, and our description pipeline could describe those new views. In the interest of time, we leave this to future work.

\paragraph{The essential role of LLMs in interpretability pipelines.}
Even if one succeeds in reverse-engineering the internals of an LLM into interactions between clean units of analysis (e.g.~SAE features, MLP neurons, weight components, or some other atomic units), another step is needed for bridging these units to human language (e.g.~language model explainers). \ourmethod{} is one such approach to designing an end-to-end interpretability system; an alternative approach is fully end-to-end interpretability by training LLMs to convert a model's internal representations to natural language \citep{huang2025predictiveconceptdecoderstraining,karvonen2026activationoraclestrainingevaluating}. The necessity of providing explanations to humans signals to us that, even if reverse-engineering is `solved', LLMs will be essential for bridging the resulting formal description into something human-understandable.

\section{Conclusion}

We introduced \ourmethod{}, an end-to-end fully automated circuit tracing and interpretation system. We described and validated our circuit tracing method, our notion of \textit{attribution profiles}, our automatic supernode-clustering algorithm, and our automatic natural-language description setup. We additionally showed experiments on both known and novel datasets, revealing meaningful clusters of MLP neurons which causally affect outputs. Overall, we hope our work contributes to progress in circuit tracing as a faithful but also automatable approach to LLM interpretability.

\section*{Acknowledgements}

We thank Dami Choi, Vincent Huang, Christopher Potts, and Dan Jurafsky for helpful discussion and feedback throughout the project.

\bibliography{colm2026_conference,main}

@article{ngo2022alignment,
      title={The Alignment Problem from a Deep Learning Perspective}, 
      author={Richard Ngo and Lawrence Chan and Sören Mindermann},
      year={2022},
      eprint={2209.00626},
      archivePrefix={arXiv},
      primaryClass={cs.AI},
      url={https://arxiv.org/abs/2209.00626}, 
      journal={arXiv:2209.00626},
}

@article{olah2020zoom,
  author = {Olah, Chris and Cammarata, Nick and Schubert, Ludwig and Goh, Gabriel and Petrov, Michael and Carter, Shan},
  title = {Zoom In: An Introduction to Circuits},
  journal = {Distill},
  year = {2020},
  note = {https://distill.pub/2020/circuits/zoom-in},
  doi = {10.23915/distill.00024.001}
}

@inproceedings{wang2022interpretability,
	address = {Kigali, Rwanda},
	title = {Interpretability in the Wild: a Circuit for Indirect Object Identification in {GPT}-2 Small},
	url = {https://openreview.net/pdf?id=NpsVSN6o4ul},
	shorttitle = {Interpretability in the Wild},
	booktitle = {The Eleventh International Conference on Learning Representations, {ICLR} 2023},
	author = {Wang, Kevin Ro and Variengien, Alexandre and Conmy, Arthur and Shlegeris, Buck and Steinhardt, Jacob},
	urldate = {2024-02-15},
	year = {2023},
}

@article{conmy2023towards,
  title={Towards automated circuit discovery for mechanistic interpretability},
  author={Conmy, Arthur and Mavor-Parker, Augustine and Lynch, Aengus and Heimersheim, Stefan and Garriga-Alonso, Adri{\`a}},
  journal={Advances in Neural Information Processing Systems},
  volume={36},
  pages={16318--16352},
  year={2023},
  url={https://arxiv.org/abs/2304.14997}
}

@article{ameisen2025circuit,
  author={Ameisen, Emmanuel and Lindsey, Jack and Pearce, Adam and Gurnee, Wes and Turner, Nicholas L. and Chen, Brian and Citro, Craig and Abrahams, David and Carter, Shan and Hosmer, Basil and Marcus, Jonathan and Sklar, Michael and Templeton, Adly and Bricken, Trenton and McDougall, Callum and Cunningham, Hoagy and Henighan, Thomas and Jermyn, Adam and Jones, Andy and Persic, Andrew and Qi, Zhenyi and Ben Thompson, T. and Zimmerman, Sam and Rivoire, Kelley and Conerly, Thomas and Olah, Chris and Batson, Joshua},
  title={Circuit Tracing: Revealing Computational Graphs in Language Models},
  journal={Transformer Circuits Thread},
  year={2025},
  url={https://transformer-circuits.pub/2025/attribution-graphs/methods.html}
}

@inproceedings{marks2025sparse,
  author       = {Samuel Marks and
                  Can Rager and
                  Eric J. Michaud and
                  Yonatan Belinkov and
                  David Bau and
                  Aaron Mueller},
  title        = {Sparse Feature Circuits: Discovering and Editing Interpretable Causal
                  Graphs in Language Models},
  booktitle    = {The Thirteenth International Conference on Learning Representations,
                  {ICLR} 2025, Singapore, April 24-28, 2025},
  publisher    = {OpenReview.net},
  year         = {2025},
  url          = {https://openreview.net/forum?id=I4e82CIDxv},
  bibsource    = {dblp computer science bibliography, https://dblp.org}
}

@article{sharkey2025open,
  author       = {Lee Sharkey and
                  Bilal Chughtai and
                  Joshua Batson and
                  Jack Lindsey and
                  Jeffrey Wu and
                  Lucius Bushnaq and
                  Nicholas Goldowsky{-}Dill and
                  Stefan Heimersheim and
                  Alejandro Ortega and
                  Joseph Isaac Bloom and
                  Stella Biderman and
                  Adri{\`{a}} Garriga{-}Alonso and
                  Arthur Conmy and
                  Neel Nanda and
                  Jessica Rumbelow and
                  Martin Wattenberg and
                  Nandi Schoots and
                  Joseph Miller and
                  William Saunders and
                  Eric J. Michaud and
                  Stephen Casper and
                  Max Tegmark and
                  David Bau and
                  Eric Todd and
                  Atticus Geiger and
                  Mor Geva and
                  Jesse Hoogland and
                  Daniel Murfet and
                  Tom McGrath},
  title        = {Open Problems in Mechanistic Interpretability},
  journal      = {Trans. Mach. Learn. Res.},
  volume       = {2025},
  year         = {2025},
  url          = {https://openreview.net/forum?id=91H76m9Z94},
  bibsource    = {dblp computer science bibliography, https://dblp.org}
}

@inproceedings{casper2024blackbox,
  author       = {Stephen Casper and
                  Carson Ezell and
                  Charlotte Siegmann and
                  Noam Kolt and
                  Taylor Lynn Curtis and
                  Benjamin Bucknall and
                  Andreas A. Haupt and
                  Kevin Wei and
                  J{\'{e}}r{\'{e}}my Scheurer and
                  Marius Hobbhahn and
                  Lee Sharkey and
                  Satyapriya Krishna and
                  Marvin Von Hagen and
                  Silas Alberti and
                  Alan Chan and
                  Qinyi Sun and
                  Michael Gerovitch and
                  David Bau and
                  Max Tegmark and
                  David Krueger and
                  Dylan Hadfield{-}Menell},
  title        = {Black-Box Access is Insufficient for Rigorous {AI} Audits},
  booktitle    = {The 2024 {ACM} Conference on Fairness, Accountability, and Transparency,
                  FAccT 2024, Rio de Janeiro, Brazil, June 3-6, 2024},
  pages        = {2254--2272},
  publisher    = {{ACM}},
  year         = {2024},
  url          = {https://doi.org/10.1145/3630106.3659037},
  doi          = {10.1145/3630106.3659037},
  bibsource    = {dblp computer science bibliography, https://dblp.org}
}

@article{ge2024automatically,
      title={Automatically Identifying Local and Global Circuits with Linear Computation Graphs}, 
      author={Xuyang Ge and Fukang Zhu and Wentao Shu and Junxuan Wang and Zhengfu He and Xipeng Qiu},
      year={2024},
      journal={arXiv:2405.13868},
      eprint={2405.13868},
      archivePrefix={arXiv},
      primaryClass={cs.LG},
      url={https://arxiv.org/abs/2405.13868}, 
}

@inproceedings{vig2020investigating,
  author       = {Jesse Vig and
                  Sebastian Gehrmann and
                  Yonatan Belinkov and
                  Sharon Qian and
                  Daniel Nevo and
                  Yaron Singer and
                  Stuart M. Shieber},
  editor       = {Hugo Larochelle and
                  Marc'Aurelio Ranzato and
                  Raia Hadsell and
                  Maria{-}Florina Balcan and
                  Hsuan{-}Tien Lin},
  title        = {Investigating Gender Bias in Language Models Using Causal Mediation
                  Analysis},
  booktitle    = {Advances in Neural Information Processing Systems 33: Annual Conference
                  on Neural Information Processing Systems 2020, NeurIPS 2020, December
                  6-12, 2020, virtual},
  year         = {2020},
  url          = {https://proceedings.neurips.cc/paper/2020/hash/92650b2e92217715fe312e6fa7b90d82-Abstract.html},
  bibsource    = {dblp computer science bibliography, https://dblp.org}
}

@inproceedings{geiger2021causal,
  author       = {Atticus Geiger and
                  Hanson Lu and
                  Thomas Icard and
                  Christopher Potts},
  editor       = {Marc'Aurelio Ranzato and
                  Alina Beygelzimer and
                  Yann N. Dauphin and
                  Percy Liang and
                  Jennifer Wortman Vaughan},
  title        = {Causal Abstractions of Neural Networks},
  booktitle    = {Advances in Neural Information Processing Systems 34: Annual Conference
                  on Neural Information Processing Systems 2021, NeurIPS 2021, December
                  6-14, 2021, virtual},
  pages        = {9574--9586},
  year         = {2021},
  url          = {https://proceedings.neurips.cc/paper/2021/hash/4f5c422f4d49a5a807eda27434231040-Abstract.html},
  bibsource    = {dblp computer science bibliography, https://dblp.org}
}

@article{jafari2025relp,
      title={RelP: Faithful and Efficient Circuit Discovery in Language Models via Relevance Patching}, 
      author={Farnoush Rezaei Jafari and Oliver Eberle and Ashkan Khakzar and Neel Nanda},
      year={2025},
      eprint={2508.21258},
      archivePrefix={arXiv},
      primaryClass={cs.LG},
      url={https://arxiv.org/abs/2508.21258}, 
      journal={arXiv:2508.21258},
}

@article{cunningham2023sparse,
      title={Sparse Autoencoders Find Highly Interpretable Features in Language Models}, 
      author={Hoagy Cunningham and Aidan Ewart and Logan Riggs and Robert Huben and Lee Sharkey},
      year={2023},
      eprint={2309.08600},
      archivePrefix={arXiv},
      primaryClass={cs.LG},
      url={https://arxiv.org/abs/2309.08600}, 
      journal={arXiv:2309.08600},
}

@article{lindsey2025biology,
  author={Lindsey, Jack and Gurnee, Wes and Ameisen, Emmanuel and Chen, Brian and Pearce, Adam and Turner, Nicholas L. and Citro, Craig and Abrahams, David and Carter, Shan and Hosmer, Basil and Marcus, Jonathan and Sklar, Michael and Templeton, Adly and Bricken, Trenton and McDougall, Callum and Cunningham, Hoagy and Henighan, Thomas and Jermyn, Adam and Jones, Andy and Persic, Andrew and Qi, Zhenyi and Thompson, T. Ben and Zimmerman, Sam and Rivoire, Kelley and Conerly, Thomas and Olah, Chris and Batson, Joshua},
  title={On the Biology of a Large Language Model},
  journal={Transformer Circuits Thread},
  year={2025},
  url={https://transformer-circuits.pub/2025/attribution-graphs/biology.html}
}

@inproceedings{meng2022locating,
  author       = {Kevin Meng and
                  David Bau and
                  Alex Andonian and
                  Yonatan Belinkov},
  editor       = {Sanmi Koyejo and
                  S. Mohamed and
                  A. Agarwal and
                  Danielle Belgrave and
                  K. Cho and
                  A. Oh},
  title        = {Locating and Editing Factual Associations in {GPT}},
  booktitle    = {Advances in Neural Information Processing Systems 35: Annual Conference
                  on Neural Information Processing Systems 2022, NeurIPS 2022, New Orleans,
                  LA, USA, November 28 - December 9, 2022},
  year         = {2022},
  url          = {http://papers.nips.cc/paper\_files/paper/2022/hash/6f1d43d5a82a37e89b0665b33bf3a182-Abstract-Conference.html},
  bibsource    = {dblp computer science bibliography, https://dblp.org}
}

@inproceedings{chan2022causal,
    author = {Chan, Lawrence and Garriga-Alonso, Adrià and Goldowsky-Dill, Nicholas and Greenblatt, Ryan and Nitishinskaya, Jenny and Radhakrishnan, Ansh and Shlegeris, Buck and Thomas, Nate},
    url = {https://www.alignmentforum.org/posts/JvZhhzycHu2Yd57RN/causal-scrubbing-a-method-for-rigorously-testing},
    title = {Causal Scrubbing: a method for rigorously testing interpretability hypotheses},
    year = {2022},
    booktitle={AI Alignment Forum},
}

@article{goldowsky2023localizing,
      title={Localizing Model Behavior with Path Patching}, 
      author={Nicholas Goldowsky-Dill and Chris MacLeod and Lucas Sato and Aryaman Arora},
      year={2023},
      eprint={2304.05969},
      archivePrefix={arXiv},
      primaryClass={cs.LG},
      url={https://arxiv.org/abs/2304.05969}, 
      journal={arXiv:2304.05969},
}

@article{templeton2024scaling,
   title={Scaling Monosemanticity: Extracting Interpretable Features from Claude 3 Sonnet},
   author={Templeton, Adly and Conerly, Tom and Marcus, Jonathan and Lindsey, Jack and Bricken, Trenton and Chen, Brian and Pearce, Adam and Citro, Craig and Ameisen, Emmanuel and Jones, Andy and Cunningham, Hoagy and Turner, Nicholas L and McDougall, Callum and MacDiarmid, Monte and Freeman, C. Daniel and Sumers, Theodore R. and Rees, Edward and Batson, Joshua and Jermyn, Adam and Carter, Shan and Olah, Chris and Henighan, Tom},
   year={2024},
   journal={Transformer Circuits Thread},
   url={https://transformer-circuits.pub/2024/scaling-monosemanticity/index.html}
}

@inproceedings{gao2025scaling,
  author       = {Leo Gao and
                  Tom Dupr{\'{e}} la Tour and
                  Henk Tillman and
                  Gabriel Goh and
                  Rajan Troll and
                  Alec Radford and
                  Ilya Sutskever and
                  Jan Leike and
                  Jeffrey Wu},
  title        = {Scaling and evaluating sparse autoencoders},
  booktitle    = {The Thirteenth International Conference on Learning Representations,
                  {ICLR} 2025, Singapore, April 24-28, 2025},
  publisher    = {OpenReview.net},
  year         = {2025},
  url          = {https://openreview.net/forum?id=tcsZt9ZNKD},
  bibsource    = {dblp computer science bibliography, https://dblp.org}
}

@inproceedings{dunefsky2024transcoders,
  author       = {Jacob Dunefsky and
                  Philippe Chlenski and
                  Neel Nanda},
  editor       = {Amir Globersons and
                  Lester Mackey and
                  Danielle Belgrave and
                  Angela Fan and
                  Ulrich Paquet and
                  Jakub M. Tomczak and
                  Cheng Zhang},
  title        = {Transcoders find interpretable {LLM} feature circuits},
  booktitle    = {Advances in Neural Information Processing Systems 38: Annual Conference
                  on Neural Information Processing Systems 2024, NeurIPS 2024, Vancouver,
                  BC, Canada, December 10 - 15, 2024},
  year         = {2024},
  url          = {http://papers.nips.cc/paper\_files/paper/2024/hash/2b8f4db0464cc5b6e9d5e6bea4b9f308-Abstract-Conference.html},
  bibsource    = {dblp computer science bibliography, https://dblp.org}
}

@inproceedings{wu2025axbench,
  author       = {Zhengxuan Wu and
                  Aryaman Arora and
                  Atticus Geiger and
                  Zheng Wang and
                  Jing Huang and
                  Dan Jurafsky and
                  Christopher D. Manning and
                  Christopher Potts},
  title        = {AxBench: Steering LLMs? Even Simple Baselines Outperform Sparse Autoencoders},
  booktitle    = {Forty-second International Conference on Machine Learning, {ICML}
                  2025, Vancouver, BC, Canada, July 13-19, 2025},
  publisher    = {OpenReview.net},
  year         = {2025},
  url          = {https://openreview.net/forum?id=K2CckZjNy0},
  bibsource    = {dblp computer science bibliography, https://dblp.org}
}

@inproceedings{syed2024attribution,
    title = "Attribution Patching Outperforms Automated Circuit Discovery",
    author = "Syed, Aaquib  and
      Rager, Can  and
      Conmy, Arthur",
    editor = "Belinkov, Yonatan  and
      Kim, Najoung  and
      Jumelet, Jaap  and
      Mohebbi, Hosein  and
      Mueller, Aaron  and
      Chen, Hanjie",
    booktitle = "Proceedings of the 7th BlackboxNLP Workshop: Analyzing and Interpreting Neural Networks for NLP",
    month = nov,
    year = "2024",
    address = "Miami, Florida, US",
    publisher = "Association for Computational Linguistics",
    url = "https://aclanthology.org/2024.blackboxnlp-1.25/",
    doi = "10.18653/v1/2024.blackboxnlp-1.25",
    pages = "407--416",
}

@inproceedings{
hanna2024faith,
title={Have Faith in Faithfulness: Going Beyond Circuit Overlap When Finding Model Mechanisms},
author={Michael Hanna and Sandro Pezzelle and Yonatan Belinkov},
booktitle={First Conference on Language Modeling},
year={2024},
url={https://openreview.net/forum?id=TZ0CCGDcuT}
}

@inproceedings{mueller2025mib,
  author       = {Aaron Mueller and
                  Atticus Geiger and
                  Sarah Wiegreffe and
                  Dana Arad and
                  Iv{\'{a}}n Arcuschin and
                  Adam Belfki and
                  Yik Siu Chan and
                  Jaden Fried Fiotto{-}Kaufman and
                  Tal Haklay and
                  Michael Hanna and
                  Jing Huang and
                  Rohan Gupta and
                  Yaniv Nikankin and
                  Hadas Orgad and
                  Nikhil Prakash and
                  Anja Reusch and
                  Aruna Sankaranarayanan and
                  Shun Shao and
                  Alessandro Stolfo and
                  Martin Tutek and
                  Amir Zur and
                  David Bau and
                  Yonatan Belinkov},
  title        = {{MIB:} {A} Mechanistic Interpretability Benchmark},
  booktitle    = {Forty-second International Conference on Machine Learning, {ICML}
                  2025, Vancouver, BC, Canada, July 13-19, 2025},
  publisher    = {OpenReview.net},
  year         = {2025},
  url          = {https://openreview.net/forum?id=sSrOwve6vb},
  bibsource    = {dblp computer science bibliography, https://dblp.org}
}

@article{choi2024automatic,
  author       = {Choi, Dami and Huang, Vincent and Meng, Kevin and Johnson, Daniel D and Steinhardt, Jacob and Schwettmann, Sarah},
  title        = {Scaling Automatic Neuron Description},
  year         = {2024},
  month        = {October},
  day          = {23},
  url = {https://transluce.org/neuron-descriptions},
  journal={Transluce Blog},
}

@inproceedings{nikankin2025arithmetic,
  author       = {Yaniv Nikankin and
                  Anja Reusch and
                  Aaron Mueller and
                  Yonatan Belinkov},
  title        = {Arithmetic Without Algorithms: Language Models Solve Math with a Bag
                  of Heuristics},
  booktitle    = {The Thirteenth International Conference on Learning Representations,
                  {ICLR} 2025, Singapore, April 24-28, 2025},
  publisher    = {OpenReview.net},
  year         = {2025},
  url          = {https://openreview.net/forum?id=O9YTt26r2P},
  bibsource    = {dblp computer science bibliography, https://dblp.org}
}

@misc{nanda2023patching,
  title={Attribution Patching: Activation Patching At Industrial Scale},
  author={Nanda, Neel},
  year={2023},
  url={https://www.neelnanda.io/mechanistic-interpretability/attribution-patching},
}

@article{shu2026completereplacement,
  author={Shu, Wentao and Ge, Xuyang and Zhou, Guancheng and Wang, Junxuan and Lin, Rui and Song, Zhaoxuan and Wu, Jiaxing and He, Zhengfu and Qiu, Xipeng},
  title={Bridging the Attention Gap: Complete Replacement Models for Complete Circuit Tracing},
  year={2026},
  journal={OpenMOSS Interpretability Research},
  url={https://interp.open-moss.com/posts/complete-replacement}
}

@inproceedings{nostalgebraist2020interpreting,
  title        = {Interpreting {GPT}: The Logit Lens},
  author       = {{nostalgebraist}},
  year         = {2020},
  url = {https://www.lesswrong.com/posts/AcKRB8wDpdaN6v6ru/interpreting-gpt-the-logit-lens},
  booktitle         = {LessWrong blog post}
}

@article{bolukbasi2021interpretabilityillusionbert,
      title={An Interpretability Illusion for BERT}, 
      author={Tolga Bolukbasi and Adam Pearce and Ann Yuan and Andy Coenen and Emily Reif and Fernanda Viégas and Martin Wattenberg},
      year={2021},
      journal={arXiv:2104.07143},
      archivePrefix={arXiv},
      primaryClass={cs.CL},
      url={https://arxiv.org/abs/2104.07143}, 
}

@article{gonen2011multiple,
  author       = {Mehmet G{\"{o}}nen and
                  Ethem Alpaydin},
  title        = {Multiple Kernel Learning Algorithms},
  journal      = {Journal of Machine Learning Research},
  volume       = {12},
  pages        = {2211--2268},
  year         = {2011},
  url          = {https://dl.acm.org/doi/10.5555/1953048.2021071},
  doi          = {10.5555/1953048.2021071},
  bibsource    = {dblp computer science bibliography, https://dblp.org}
}

@article{arora2026languagemodelcircuitssparse,
      title={Language Model Circuits Are Sparse in the Neuron Basis}, 
      author={Aryaman Arora and Zhengxuan Wu and Jacob Steinhardt and Sarah Schwettmann},
      year={2026},
      journal={arXiv:2601.22594},
      archivePrefix={arXiv},
      primaryClass={cs.CL},
      url={https://arxiv.org/abs/2601.22594}, 
}

@article{chowdhury2025surfacing,
  author       = {Chowdhury, Neil and Schwettmann, Sarah and Steinhardt, Jacob and Johnson, Daniel D.},
  title        = {Surfacing Pathological Behaviors in Language Models},
  year         = {2025},
  month        = {June},
  day          = {5},
  url = {https://transluce.org/pathological-behaviors},
  journal = {Transluce Blog},
}

@inproceedings{hanna-etal-2025-circuit,
    title = "Circuit-Tracer: A New Library for Finding Feature Circuits",
    author = "Hanna, Michael  and
      Piotrowski, Mateusz  and
      Lindsey, Jack  and
      Ameisen, Emmanuel",
    editor = "Belinkov, Yonatan  and
      Mueller, Aaron  and
      Kim, Najoung  and
      Mohebbi, Hosein  and
      Chen, Hanjie  and
      Arad, Dana  and
      Sarti, Gabriele",
    booktitle = "Proceedings of the 8th BlackboxNLP Workshop: Analyzing and Interpreting Neural Networks for NLP",
    month = nov,
    year = "2025",
    address = "Suzhou, China",
    publisher = "Association for Computational Linguistics",
    url = "https://aclanthology.org/2025.blackboxnlp-1.14/",
    doi = "10.18653/v1/2025.blackboxnlp-1.14",
    pages = "239--249",
    ISBN = "979-8-89176-346-3",
}

@article{karvonen2026activationoraclestrainingevaluating,
      title={Activation Oracles: Training and Evaluating LLMs as General-Purpose Activation Explainers}, 
      author={Adam Karvonen and James Chua and Clément Dumas and Kit Fraser-Taliente and Subhash Kantamneni and Julian Minder and Euan Ong and Arnab Sen Sharma and Daniel Wen and Owain Evans and Samuel Marks},
      year={2026},
      journal={arXiv:2512.15674},
      archivePrefix={arXiv},
      primaryClass={cs.CL},
      url={https://arxiv.org/abs/2512.15674}, 
}

@article{huang2025predictiveconceptdecoderstraining,
      title={Predictive Concept Decoders: Training Scalable End-to-End Interpretability Assistants}, 
      author={Vincent Huang and Dami Choi and Daniel D. Johnson and Sarah Schwettmann and Jacob Steinhardt},
      year={2025},
      journal={arXiv:2512.15712},
      archivePrefix={arXiv},
      primaryClass={cs.AI},
      url={https://arxiv.org/abs/2512.15712}, 
}

@article{kamath2025tracing, author={Kamath, Harish and Ameisen, Emmanuel and Kauvar, Isaac and Luger, Rodrigo and Gurnee, Wes and Pearce, Adam and Zimmerman, Sam and Batson, Joshua and Conerly, Thomas and Olah, Chris and Lindsey, Jack}, title={Tracing Attention Computation Through Feature Interactions}, journal={Transformer Circuits Thread}, year={2025}, url={https://transformer-circuits.pub/2025/attention-qk/index.html} }

@inproceedings{fineweb,
  author       = {Guilherme Penedo and
                  Hynek Kydl{\'{\i}}cek and
                  Loubna Ben Allal and
                  Anton Lozhkov and
                  Margaret Mitchell and
                  Colin A. Raffel and
                  Leandro von Werra and
                  Thomas Wolf},
  editor       = {Amir Globersons and
                  Lester Mackey and
                  Danielle Belgrave and
                  Angela Fan and
                  Ulrich Paquet and
                  Jakub M. Tomczak and
                  Cheng Zhang},
  title        = {The FineWeb Datasets: Decanting the Web for the Finest Text Data at
                  Scale},
  booktitle    = {Advances in Neural Information Processing Systems 38: Annual Conference
                  on Neural Information Processing Systems 2024, NeurIPS 2024, Vancouver,
                  BC, Canada, December 10 - 15, 2024},
  year         = {2024},
  url          = {http://papers.nips.cc/paper\_files/paper/2024/hash/370df50ccfdf8bde18f8f9c2d9151bda-Abstract-Datasets\_and\_Benchmarks\_Track.html},
  bibsource    = {dblp computer science bibliography, https://dblp.org}
}

@article{bills2023language,
 title={Language models can explain neurons in language models},
 author={
    Bills, Steven and Cammarata, Nick and Mossing, Dan and Tillman, Henk and Gao, Leo and Goh, Gabriel and Sutskever, Ilya and Leike, Jan and Wu, Jeff and Saunders, William
 },
 year={2023},
 url = {https://openaipublic.blob.core.windows.net/neuron-explainer/paper/index.html},
 journal = {OpenAI}
}

@article{paulo2025automatically,
  author       = {Gon{\c{c}}alo Paulo and
                  Alex Mallen and
                  Caden Juang and
                  Nora Belrose},
  editor       = {Aarti Singh and
                  Maryam Fazel and
                  Daniel Hsu and
                  Simon Lacoste{-}Julien and
                  Felix Berkenkamp and
                  Tegan Maharaj and
                  Kiri Wagstaff and
                  Jerry Zhu},
  title        = {Automatically Interpreting Millions of Features in Large Language
                  Models},
  booktitle    = {Forty-second International Conference on Machine Learning, {ICML}
                  2025, Vancouver, BC, Canada, July 13-19, 2025},
  series       = {Proceedings of Machine Learning Research},
  volume       = {267},
  publisher    = {{PMLR} / OpenReview.net},
  year         = {2025},
  url          = {https://proceedings.mlr.press/v267/paulo25a.html},
  bibsource    = {dblp computer science bibliography, https://dblp.org}
}

@article{foote2023neurongraphinterpretinglanguage,
      title={Neuron to Graph: Interpreting Language Model Neurons at Scale}, 
      author={Alex Foote and Neel Nanda and Esben Kran and Ioannis Konstas and Shay Cohen and Fazl Barez},
      year={2023},
      journal={arXiv:2305.19911},
      archivePrefix={arXiv},
      primaryClass={cs.LG},
      url={https://arxiv.org/abs/2305.19911}, 
}

@inproceedings{vllm,
  author       = {Woosuk Kwon and
                  Zhuohan Li and
                  Siyuan Zhuang and
                  Ying Sheng and
                  Lianmin Zheng and
                  Cody Hao Yu and
                  Joseph Gonzalez and
                  Hao Zhang and
                  Ion Stoica},
  editor       = {Jason Flinn and
                  Margo I. Seltzer and
                  Peter Druschel and
                  Antoine Kaufmann and
                  Jonathan Mace},
  title        = {Efficient Memory Management for Large Language Model Serving with
                  PagedAttention},
  booktitle    = {Proceedings of the 29th Symposium on Operating Systems Principles,
                  {SOSP} 2023, Koblenz, Germany, October 23-26, 2023},
  pages        = {611--626},
  publisher    = {{ACM}},
  year         = {2023},
  url          = {https://doi.org/10.1145/3600006.3613165},
  doi          = {10.1145/3600006.3613165},
  bibsource    = {dblp computer science bibliography, https://dblp.org}
}
\bibliographystyle{colm2026_conference}

\newpage
\appendix
\renewcommand \thepart{}
\renewcommand \partname{}
\noptcrule
\part{Appendix} 
\parttoc 
\newpage

\newpage
\section{Circuit tracing backbone}
\label{sec:tracing-more}

\paragraph{Properties of RelP.} Under RelP, the following holds: formally, let $h_d^{(l,t)}(x)$ denote the $d$-th activation of the layer-$l$ residual stream at token $t$ on input $x$, and let $\mathcal{M}(x)_k^{(s)}$ denote the $k$-th output logit of the replacement model at another token $s$. We have:
\begin{equation}
    \sum_{t} \sum_{d} h_d^{(l,t)}(x) \cdot \frac{\partial \mathcal{M}(x)_k^{(s)}}{\partial h_d^{(l,t)}} = \mathcal{M}(x)_k^{(s)}
\end{equation}
The end result is that the model's forward pass behaviour (and thus output) is unchanged for that specific input, but now the input times gradient is conservative. 

\paragraph{Edge weights.} Between the neurons which we keep in our circuit, we compute their edge weights using a similar attribution approach. Given a source MLP neuron $m_{v}^{(m,s)}$ and a target MLP neuron $m_u^{(l,t)}(x)$, we freeze gradients via MLPs in intermediate layers and compute edge weight:
\begin{equation}
    \alpha(m_{u}^{(r,s)}, m_{u}^{(l,t)}) = m_{v}^{(r,s)}(x) \cdot \frac{\partial m_{u}^{(l,t)}(x)}{\partial m_{v}^{(r,s)}}
\end{equation}
This tells how much of the activation of the target MLP neuron came from the source MLP neuron. We similarly compute edge weights between token embeddings and MLP neurons, and MLP neurons and output logits.

\section{Systems details and benchmarking results}

\begin{figure}[!h]
    \centering
    \begin{subfigure}[t]{0.48\textwidth}
      \centering
      \includegraphics[width=\linewidth]{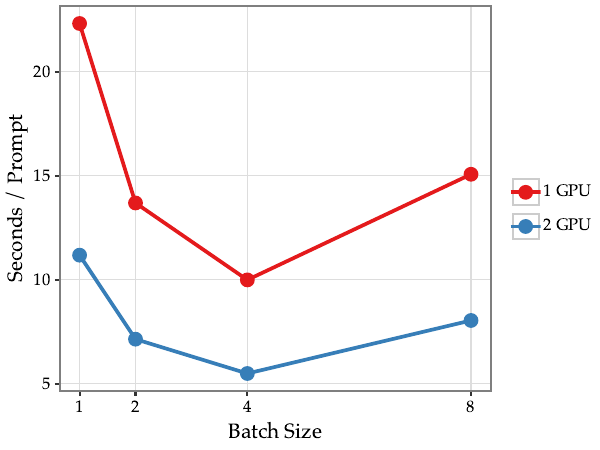}
      \caption{Time for tracing the circuit per example.}
      \label{fig:throughput}
    \end{subfigure}
    \hfill
    \begin{subfigure}[t]{0.48\textwidth}
      \centering
      \includegraphics[width=\linewidth]{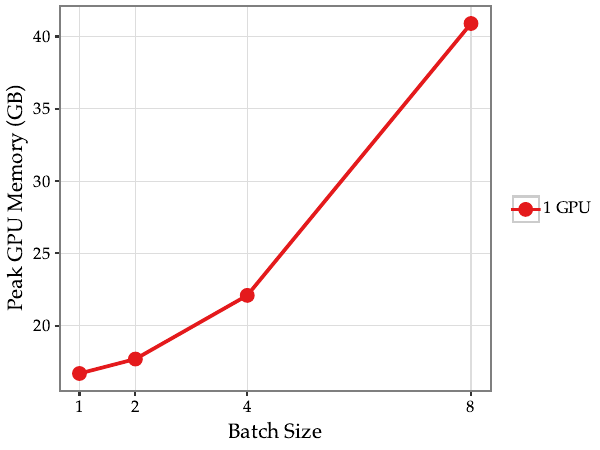}
      \caption{Peak GPU memory usage.}
      \label{fig:memory}
    \end{subfigure}
    \caption{Benchmarking results for circuit tracing + attribution profile computation for Llama 3.1 8B Instruct, on \texttt{capitals}.}
    \label{fig:bench}
\end{figure}

We reimplemented the circuit tracing backbone from \citet{arora2026languagemodelcircuitssparse} more efficiently via the following improvements:
\begin{itemize}
\item Removing the dependency on \texttt{nnsight} (only used for collecting MLP neuron activations) and replacing it with vanilla \texttt{torch} hooks.
\item Removing unnecessary \texttt{cuda} syncs. \item Batching gradient computations that involve multiple backward passes (e.g.~attribution computation for each layer) to just use one and retain the graph.
\item Better batching for Jacobian computations (for attribution and contribution profiles, and edge weights).
\item Implementing data parallelism (batching multiple dataset examples onto one GPU, and also splitting up a dataset over multiple workers across GPUs).
\item Implementing model sharding via \texttt{transformers}'s \texttt{device\_map=``sequential"} which keeps hooks on the right GPU.
\end{itemize}

We benchmark the runtime and peak GPU memory usage of circuit tracing and attribution profile computation on the \texttt{capitals} dataset. We use a cluster of H100 80GB GPUs for all of our experiments. Results are in \cref{fig:bench}.

These systems optimisations enable us to run circuit tracing on larger multi-GPU models (e.g. Qwen3 32B; \cref{sec:capitals-qwen}) and much larger datasets; for the \texttt{math} dataset with Llama 3.1 8B Instruct, which has 10,000 dataset examples, we were able to trace all circuits in $\approx 6$ hours with data paralellism over $4$ GPUs and per-GPU batch size of $4$.

For the remaining steps of the \ourmethod{} pipeline, we further optimise our implementations by vectorising clustering steps with \texttt{numpy}, using \texttt{vllm} \citep{vllm} for efficient inference for our input attribution explainer and simulator models, and batching calls to the Anthropic API for API explainers, simulators, and summarisers.

\section{Non-locality in MLP neurons}
\label{sec:locality-more}
\begin{figure*}[!h]
    \centering
    \begin{subfigure}[t]{0.48\textwidth}
      \centering
      \includegraphics[width=\linewidth]{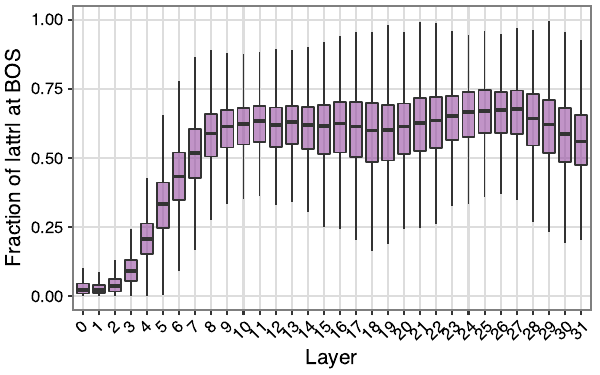}
      \caption{Average fraction of attribution score allocated to the BOS token.}
      \label{fig:bos}
    \end{subfigure}
    \hfill
    \begin{subfigure}[t]{0.48\textwidth}
      \centering
      \includegraphics[width=\linewidth]{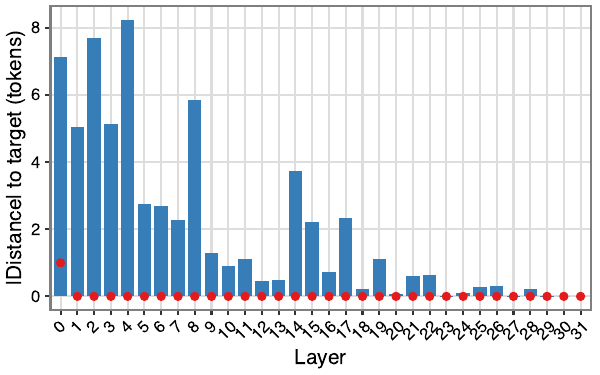}
      \caption{Mean (blue bar) and median (red dot) distance of the top-contributing neuron from the target logit, in tokens (excluding BOS).}
      \label{fig:mean}
    \end{subfigure}
    \caption{Additional locality results for MLP neurons in Llama 3.1 8B Instruct.}
    \label{fig:extra-locality}
\end{figure*}

In \cref{fig:bos} we found that in later layers the majority of attribution goes to the beginning-of-string (BOS) token; we hypothesise that this is related to the attention sink phenomenon given that these later layer neurons must depend on attention to obtain non-local contextual information. On the basis of this result, we exclude the BOS token from our input attribution profiles.

For contribution profiles, we can additionally examine the mean and median distances of the top-contribution non-BOS MLP neurons, in \cref{fig:mean}. We see greater non-locality in early layers, adding additional evidence to the result in the main text.

\newpage
\section{Additional results on \texttt{capitals} dataset}

\subsection{Ablations on threshold selection in attribution descriptions}
\begin{figure}[!h]
    \centering
    \includegraphics[width=0.5\linewidth]{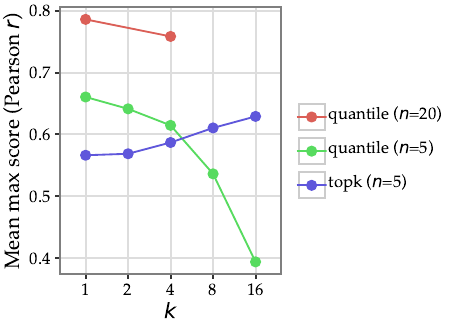}
    \caption{Attribution description score for quantile and topk approaches, when sweeping $k$ for highlight threshold.}
    \label{fig:threshold}
\end{figure}

In the prompt passed to the attribution explainer, we select an attribution threshold above which exemplar tokens get highlighted with \verb|{{}}|. We compare two approaches to picking the threshold:
\begin{itemize}
    \item \textbf{Quantile}: Given a precomputed list of percentiles and a minimum highlight parameter $k$, we select the highest percentile such that at least $k$ unique substrings are highlighted. This is identical to \citet{choi2024automatic}.
    \item \textbf{TopK}: We select the $k$-th highest score over the entire dataset as the threshold.
\end{itemize}
We sweep $k \in \{1, 2, 4, 8, 16\}$ with both approaches and report the mean of the maximum per-cluster simulator scores of the generated descriptions. Results in \cref{fig:threshold} show that quantile-based threshold selection with $k=1$ is the best and so we use this setting throughout for attribution descriptions. Simply increasing the number of samples from $5$ to $20$ delivers substantial gains.

\newpage
\subsection{Detailed results for Llama 3.1 8B Instruct}

\begin{figure}[!h]
    \centering
    \includegraphics[width=\linewidth]{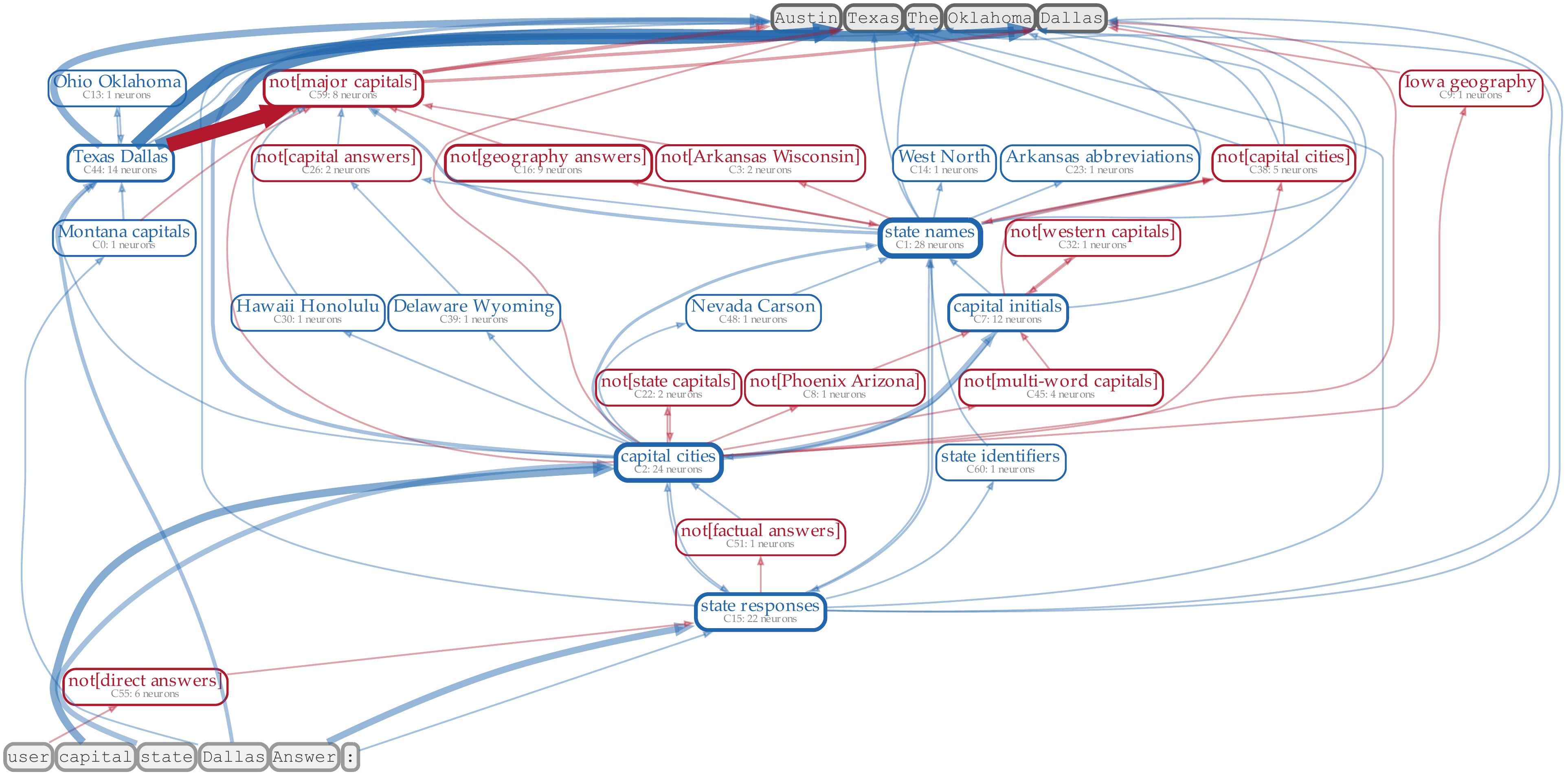}
    \caption{Final circuit graph for \texttt{texas} example in the \texttt{capitals} dataset from Llama 3.1 8B Instruct.}
    \label{fig:capitals-complete}
\end{figure}

We provide detailed results for Llama 3.1 8B Instruct on the \texttt{capitals} dataset below. The dataset consists of examples such as:
\begin{chatbox}[\texttt{capitals}: Llama 3.1 8B Instruct]
  \userquery{What is the capital of the state containing Dallas?}
  \assistantresponse{Answer:}
\end{chatbox}
First, we show the complete circuit graph for the \texttt{austin} example in the dataset in \cref{fig:capitals-complete}; many weak supernodes, which hardly affect output behaviour and have labels relating to \textit{other} states and their capitals, are apparent.

Afterwards, we provide steering results for every supernode in each of three examples: \texttt{austin}, \texttt{sacramento}, and \texttt{atlanta}.

\label{sec:capitals-steering}
\subsubsection{\texttt{austin}}

\begin{center}
\centering
\small
\adjustbox{max width=\textwidth}{
\begin{tabular}{llllll}
\toprule
Cluster & Token 1 & Token 2 & Token 3 & Token 4 & Token 5 \\
\midrule
C0: western state capitals & \cellcolor[RGB]{224,242,237}\texttt{\_Austin} (96.5\%) & \cellcolor[RGB]{254,232,223}\texttt{\_Texas} (1.4\%) & \cellcolor[RGB]{232,236,244}\texttt{\_The} (0.8\%) & \cellcolor[RGB]{250,231,243}\texttt{\_Oklahoma} (0.7\%) & \cellcolor[RGB]{237,247,220}\texttt{\_Dallas} (0.2\%) \\
C1: state name bias & \cellcolor[RGB]{224,242,237}\texttt{\_Austin} (94.1\%) & \cellcolor[RGB]{232,236,244}\texttt{\_The} (2.8\%) & \cellcolor[RGB]{254,232,223}\texttt{\_Texas} (0.5\%) & \cellcolor[RGB]{250,231,243}\texttt{\_Oklahoma} (0.5\%) & \cellcolor[RGB]{237,247,220}\texttt{\_Dallas} (0.4\%) \\
C2: capital city first token & \cellcolor[RGB]{254,232,223}\texttt{\_Texas} (93.4\%) & \cellcolor[RGB]{250,231,243}\texttt{\_Oklahoma} (3.2\%) & \cellcolor[RGB]{249,243,233}\texttt{\_TX} (0.6\%) & \cellcolor[RGB]{237,247,220}\texttt{\_Dallas} (0.2\%) & \cellcolor[RGB]{239,239,239}\texttt{\_Arkansas} (0.2\%) \\
C3: not[western southern capitals] & \cellcolor[RGB]{224,242,237}\texttt{\_Austin} (96.5\%) & \cellcolor[RGB]{254,232,223}\texttt{\_Texas} (1.6\%) & \cellcolor[RGB]{232,236,244}\texttt{\_The} (0.7\%) & \cellcolor[RGB]{250,231,243}\texttt{\_Oklahoma} (0.5\%) & \cellcolor[RGB]{237,247,220}\texttt{\_Dallas} (0.1\%) \\
C7: capital city initiation & \cellcolor[RGB]{224,242,237}\texttt{\_Austin} (87.9\%) & \cellcolor[RGB]{254,232,223}\texttt{\_Texas} (9.3\%) & \cellcolor[RGB]{232,236,244}\texttt{\_The} (1.3\%) & \cellcolor[RGB]{250,231,243}\texttt{\_Oklahoma} (0.6\%) & \cellcolor[RGB]{237,247,220}\texttt{\_Dallas} (0.2\%) \\
C8: not[factual city answers] & \cellcolor[RGB]{224,242,237}\texttt{\_Austin} (96.5\%) & \cellcolor[RGB]{254,232,223}\texttt{\_Texas} (1.6\%) & \cellcolor[RGB]{232,236,244}\texttt{\_The} (0.7\%) & \cellcolor[RGB]{250,231,243}\texttt{\_Oklahoma} (0.6\%) & \cellcolor[RGB]{237,247,220}\texttt{\_Dallas} (0.2\%) \\
C9: Iowa Cedar Rapids & \cellcolor[RGB]{224,242,237}\texttt{\_Austin} (96.1\%) & \cellcolor[RGB]{254,232,223}\texttt{\_Texas} (1.8\%) & \cellcolor[RGB]{232,236,244}\texttt{\_The} (0.8\%) & \cellcolor[RGB]{250,231,243}\texttt{\_Oklahoma} (0.6\%) & \cellcolor[RGB]{237,247,220}\texttt{\_Dallas} (0.2\%) \\
C13: Oklahoma Tulsa & \cellcolor[RGB]{224,242,237}\texttt{\_Austin} (96.5\%) & \cellcolor[RGB]{254,232,223}\texttt{\_Texas} (1.8\%) & \cellcolor[RGB]{232,236,244}\texttt{\_The} (0.8\%) & \cellcolor[RGB]{250,231,243}\texttt{\_Oklahoma} (0.3\%) & \cellcolor[RGB]{237,247,220}\texttt{\_Dallas} (0.1\%) \\
C14: Huntington state names & \cellcolor[RGB]{224,242,237}\texttt{\_Austin} (96.1\%) & \cellcolor[RGB]{254,232,223}\texttt{\_Texas} (1.8\%) & \cellcolor[RGB]{232,236,244}\texttt{\_The} (0.8\%) & \cellcolor[RGB]{250,231,243}\texttt{\_Oklahoma} (0.6\%) & \cellcolor[RGB]{237,247,220}\texttt{\_Dallas} (0.2\%) \\
C15: state capitals broad & \cellcolor[RGB]{224,242,237}\texttt{\_Austin} (94.9\%) & \cellcolor[RGB]{254,232,223}\texttt{\_Texas} (2.5\%) & \cellcolor[RGB]{232,236,244}\texttt{\_The} (1.7\%) & \cellcolor[RGB]{250,231,243}\texttt{\_Oklahoma} (0.1\%) & \cellcolor[RGB]{255,247,213}\texttt{\_} (0.1\%) \\
C16: not[direct capital answers] & \cellcolor[RGB]{224,242,237}\texttt{\_Austin} (95.3\%) & \cellcolor[RGB]{254,232,223}\texttt{\_Texas} (2.5\%) & \cellcolor[RGB]{232,236,244}\texttt{\_The} (1.2\%) & \cellcolor[RGB]{250,231,243}\texttt{\_Oklahoma} (0.3\%) & \cellcolor[RGB]{237,247,220}\texttt{\_Dallas} (0.3\%) \\
C22: not[state capital tokens] & \cellcolor[RGB]{224,242,237}\texttt{\_Austin} (96.5\%) & \cellcolor[RGB]{254,232,223}\texttt{\_Texas} (1.8\%) & \cellcolor[RGB]{232,236,244}\texttt{\_The} (0.7\%) & \cellcolor[RGB]{250,231,243}\texttt{\_Oklahoma} (0.6\%) & \cellcolor[RGB]{237,247,220}\texttt{\_Dallas} (0.1\%) \\
C23: Arkansas abbreviation & \cellcolor[RGB]{224,242,237}\texttt{\_Austin} (96.5\%) & \cellcolor[RGB]{254,232,223}\texttt{\_Texas} (1.6\%) & \cellcolor[RGB]{232,236,244}\texttt{\_The} (0.7\%) & \cellcolor[RGB]{250,231,243}\texttt{\_Oklahoma} (0.6\%) & \cellcolor[RGB]{237,247,220}\texttt{\_Dallas} (0.2\%) \\
C26: not[factual geography answers] & \cellcolor[RGB]{224,242,237}\texttt{\_Austin} (96.1\%) & \cellcolor[RGB]{254,232,223}\texttt{\_Texas} (2.0\%) & \cellcolor[RGB]{232,236,244}\texttt{\_The} (0.8\%) & \cellcolor[RGB]{250,231,243}\texttt{\_Oklahoma} (0.6\%) & \cellcolor[RGB]{237,247,220}\texttt{\_Dallas} (0.2\%) \\
C30: Hawaii Hilo & \cellcolor[RGB]{224,242,237}\texttt{\_Austin} (96.5\%) & \cellcolor[RGB]{254,232,223}\texttt{\_Texas} (1.6\%) & \cellcolor[RGB]{232,236,244}\texttt{\_The} (0.8\%) & \cellcolor[RGB]{250,231,243}\texttt{\_Oklahoma} (0.6\%) & \cellcolor[RGB]{237,247,220}\texttt{\_Dallas} (0.2\%) \\
C32: not[two-word city capitals] & \cellcolor[RGB]{224,242,237}\texttt{\_Austin} (96.9\%) & \cellcolor[RGB]{254,232,223}\texttt{\_Texas} (1.6\%) & \cellcolor[RGB]{232,236,244}\texttt{\_The} (0.7\%) & \cellcolor[RGB]{250,231,243}\texttt{\_Oklahoma} (0.4\%) & \cellcolor[RGB]{237,247,220}\texttt{\_Dallas} (0.2\%) \\
C38: not[major city capitals] & \cellcolor[RGB]{224,242,237}\texttt{\_Austin} (95.3\%) & \cellcolor[RGB]{232,236,244}\texttt{\_The} (1.5\%) & \cellcolor[RGB]{254,232,223}\texttt{\_Texas} (1.2\%) & \cellcolor[RGB]{250,231,243}\texttt{\_Oklahoma} (0.7\%) & \cellcolor[RGB]{237,247,220}\texttt{\_Dallas} (0.3\%) \\
C39: small state names & \cellcolor[RGB]{224,242,237}\texttt{\_Austin} (94.9\%) & \cellcolor[RGB]{254,232,223}\texttt{\_Texas} (2.5\%) & \cellcolor[RGB]{232,236,244}\texttt{\_The} (1.2\%) & \cellcolor[RGB]{250,231,243}\texttt{\_Oklahoma} (0.7\%) & \cellcolor[RGB]{237,247,220}\texttt{\_Dallas} (0.2\%) \\
C44: Dallas Texas & \cellcolor[RGB]{224,242,237}\texttt{\_Austin} (52.7\%) & \cellcolor[RGB]{250,231,243}\texttt{\_Oklahoma} (22.1\%) & \cellcolor[RGB]{224,242,237}\texttt{\_Atlanta} (11.8\%) & \cellcolor[RGB]{254,232,223}\texttt{\_Texas} (3.4\%) & \cellcolor[RGB]{232,236,244}\texttt{\_The} (3.4\%) \\
C45: not[capital city answers] & \cellcolor[RGB]{224,242,237}\texttt{\_Austin} (96.9\%) & \cellcolor[RGB]{254,232,223}\texttt{\_Texas} (1.6\%) & \cellcolor[RGB]{232,236,244}\texttt{\_The} (0.7\%) & \cellcolor[RGB]{250,231,243}\texttt{\_Oklahoma} (0.5\%) & \cellcolor[RGB]{237,247,220}\texttt{\_Dallas} (0.1\%) \\
C48: Las Vegas Nevada & \cellcolor[RGB]{224,242,237}\texttt{\_Austin} (96.9\%) & \cellcolor[RGB]{254,232,223}\texttt{\_Texas} (1.4\%) & \cellcolor[RGB]{232,236,244}\texttt{\_The} (0.7\%) & \cellcolor[RGB]{250,231,243}\texttt{\_Oklahoma} (0.5\%) & \cellcolor[RGB]{237,247,220}\texttt{\_Dallas} (0.1\%) \\
C51: not[Warwick geography] & \cellcolor[RGB]{224,242,237}\texttt{\_Austin} (96.9\%) & \cellcolor[RGB]{254,232,223}\texttt{\_Texas} (1.4\%) & \cellcolor[RGB]{232,236,244}\texttt{\_The} (0.7\%) & \cellcolor[RGB]{250,231,243}\texttt{\_Oklahoma} (0.5\%) & \cellcolor[RGB]{237,247,220}\texttt{\_Dallas} (0.1\%) \\
C55: not[geography answers] & \cellcolor[RGB]{224,242,237}\texttt{\_Austin} (96.9\%) & \cellcolor[RGB]{254,232,223}\texttt{\_Texas} (1.2\%) & \cellcolor[RGB]{250,231,243}\texttt{\_Oklahoma} (0.7\%) & \cellcolor[RGB]{232,236,244}\texttt{\_The} (0.7\%) & \cellcolor[RGB]{237,247,220}\texttt{\_Dallas} (0.1\%) \\
C59: not[southern capitals] & \cellcolor[RGB]{224,242,237}\texttt{\_Austin} (98.4\%) & \cellcolor[RGB]{254,232,223}\texttt{\_Texas} (1.2\%) & \cellcolor[RGB]{250,231,243}\texttt{\_Oklahoma} (0.1\%) & \cellcolor[RGB]{237,247,220}\texttt{\_Dallas} (0.1\%) & \cellcolor[RGB]{232,236,244}\texttt{\_The} (0.0\%) \\
C60: state name completions & \cellcolor[RGB]{224,242,237}\texttt{\_Austin} (96.5\%) & \cellcolor[RGB]{254,232,223}\texttt{\_Texas} (1.6\%) & \cellcolor[RGB]{232,236,244}\texttt{\_The} (0.8\%) & \cellcolor[RGB]{250,231,243}\texttt{\_Oklahoma} (0.5\%) & \cellcolor[RGB]{237,247,220}\texttt{\_Dallas} (0.1\%) \\
\bottomrule
\end{tabular}}
\end{center}

\subsubsection{\texttt{sacramento}}

\begin{center}
\centering
\small
\adjustbox{max width=\textwidth}{
\begin{tabular}{llllll}
\toprule
Cluster & Token 1 & Token 2 & Token 3 & Token 4 & Token 5 \\
\midrule
C1: state name bias & \cellcolor[RGB]{254,232,223}\texttt{\_Sacramento} (74.2\%) & \cellcolor[RGB]{232,236,244}\texttt{\_Los} (14.6\%) & \cellcolor[RGB]{232,236,244}\texttt{\_The} (6.1\%) & \cellcolor[RGB]{250,231,243}\texttt{\_California} (1.7\%) & \cellcolor[RGB]{237,247,220}\texttt{\_None} (1.2\%) \\
C2: capital city first token & \cellcolor[RGB]{250,231,243}\texttt{\_California} (87.1\%) & \cellcolor[RGB]{249,243,233}\texttt{\_CA} (3.8\%) & \cellcolor[RGB]{232,236,244}\texttt{\_Los} (3.0\%) & \cellcolor[RGB]{239,239,239}\texttt{\_Washington} (0.9\%) & \cellcolor[RGB]{224,242,237}\texttt{\_Cal} (0.5\%) \\
C7: capital city initiation & \cellcolor[RGB]{254,232,223}\texttt{\_Sacramento} (93.8\%) & \cellcolor[RGB]{232,236,244}\texttt{\_Los} (2.8\%) & \cellcolor[RGB]{250,231,243}\texttt{\_California} (1.5\%) & \cellcolor[RGB]{232,236,244}\texttt{\_The} (1.2\%) & \cellcolor[RGB]{255,247,213}\texttt{\_Sac} (0.1\%) \\
C8: not[factual city answers] & \cellcolor[RGB]{254,232,223}\texttt{\_Sacramento} (93.8\%) & \cellcolor[RGB]{232,236,244}\texttt{\_Los} (3.2\%) & \cellcolor[RGB]{232,236,244}\texttt{\_The} (1.2\%) & \cellcolor[RGB]{250,231,243}\texttt{\_California} (1.0\%) & \cellcolor[RGB]{255,247,213}\texttt{\_Sac} (0.2\%) \\
C9: Iowa Cedar Rapids & \cellcolor[RGB]{254,232,223}\texttt{\_Sacramento} (93.4\%) & \cellcolor[RGB]{232,236,244}\texttt{\_Los} (3.6\%) & \cellcolor[RGB]{232,236,244}\texttt{\_The} (1.2\%) & \cellcolor[RGB]{250,231,243}\texttt{\_California} (0.9\%) & \cellcolor[RGB]{255,247,213}\texttt{\_Sac} (0.3\%) \\
C15: state capitals broad & \cellcolor[RGB]{254,232,223}\texttt{\_Sacramento} (89.5\%) & \cellcolor[RGB]{232,236,244}\texttt{\_Los} (5.1\%) & \cellcolor[RGB]{250,231,243}\texttt{\_California} (2.4\%) & \cellcolor[RGB]{232,236,244}\texttt{\_The} (1.4\%) & \cellcolor[RGB]{255,247,213}\texttt{\_Sac} (0.5\%) \\
C16: not[direct capital answers] & \cellcolor[RGB]{254,232,223}\texttt{\_Sacramento} (94.1\%) & \cellcolor[RGB]{232,236,244}\texttt{\_Los} (2.5\%) & \cellcolor[RGB]{232,236,244}\texttt{\_The} (1.2\%) & \cellcolor[RGB]{250,231,243}\texttt{\_California} (1.2\%) & \cellcolor[RGB]{255,247,213}\texttt{\_Sac} (0.3\%) \\
C26: not[factual geography answers] & \cellcolor[RGB]{254,232,223}\texttt{\_Sacramento} (93.4\%) & \cellcolor[RGB]{232,236,244}\texttt{\_Los} (3.2\%) & \cellcolor[RGB]{232,236,244}\texttt{\_The} (1.3\%) & \cellcolor[RGB]{250,231,243}\texttt{\_California} (1.2\%) & \cellcolor[RGB]{255,247,213}\texttt{\_Sac} (0.2\%) \\
C30: Hawaii Hilo & \cellcolor[RGB]{254,232,223}\texttt{\_Sacramento} (93.0\%) & \cellcolor[RGB]{232,236,244}\texttt{\_Los} (3.2\%) & \cellcolor[RGB]{250,231,243}\texttt{\_California} (1.5\%) & \cellcolor[RGB]{232,236,244}\texttt{\_The} (1.3\%) & \cellcolor[RGB]{255,247,213}\texttt{\_Sac} (0.2\%) \\
C32: not[two-word city capitals] & \cellcolor[RGB]{254,232,223}\texttt{\_Sacramento} (93.4\%) & \cellcolor[RGB]{232,236,244}\texttt{\_Los} (3.2\%) & \cellcolor[RGB]{232,236,244}\texttt{\_The} (1.3\%) & \cellcolor[RGB]{250,231,243}\texttt{\_California} (1.0\%) & \cellcolor[RGB]{255,247,213}\texttt{\_Sac} (0.3\%) \\
C36: Los Angeles California & \cellcolor[RGB]{254,232,223}\texttt{\_Sacramento} (89.1\%) & \cellcolor[RGB]{232,236,244}\texttt{\_Los} (7.3\%) & \cellcolor[RGB]{232,236,244}\texttt{\_The} (1.4\%) & \cellcolor[RGB]{250,231,243}\texttt{\_California} (0.9\%) & \cellcolor[RGB]{255,247,213}\texttt{\_Sac} (0.2\%) \\
C38: not[major city capitals] & \cellcolor[RGB]{254,232,223}\texttt{\_Sacramento} (97.7\%) & \cellcolor[RGB]{232,236,244}\texttt{\_Los} (1.2\%) & \cellcolor[RGB]{250,231,243}\texttt{\_California} (0.5\%) & \cellcolor[RGB]{232,236,244}\texttt{\_The} (0.2\%) & \cellcolor[RGB]{255,247,213}\texttt{\_Sac} (0.1\%) \\
C39: small state names & \cellcolor[RGB]{254,232,223}\texttt{\_Sacramento} (92.6\%) & \cellcolor[RGB]{232,236,244}\texttt{\_Los} (3.2\%) & \cellcolor[RGB]{232,236,244}\texttt{\_The} (1.7\%) & \cellcolor[RGB]{250,231,243}\texttt{\_California} (1.5\%) & \cellcolor[RGB]{255,247,213}\texttt{\_Sac} (0.2\%) \\
C43: Jacksonville Florida & \cellcolor[RGB]{254,232,223}\texttt{\_Sacramento} (94.1\%) & \cellcolor[RGB]{232,236,244}\texttt{\_Los} (2.8\%) & \cellcolor[RGB]{232,236,244}\texttt{\_The} (1.2\%) & \cellcolor[RGB]{250,231,243}\texttt{\_California} (1.0\%) & \cellcolor[RGB]{255,247,213}\texttt{\_Sac} (0.2\%) \\
C45: not[capital city answers] & \cellcolor[RGB]{254,232,223}\texttt{\_Sacramento} (95.3\%) & \cellcolor[RGB]{232,236,244}\texttt{\_Los} (2.0\%) & \cellcolor[RGB]{232,236,244}\texttt{\_The} (1.1\%) & \cellcolor[RGB]{250,231,243}\texttt{\_California} (0.9\%) & \cellcolor[RGB]{255,247,213}\texttt{\_Sac} (0.2\%) \\
C47: Albuquerque Santa Fe & \cellcolor[RGB]{254,232,223}\texttt{\_Sacramento} (94.5\%) & \cellcolor[RGB]{232,236,244}\texttt{\_Los} (3.2\%) & \cellcolor[RGB]{250,231,243}\texttt{\_California} (0.9\%) & \cellcolor[RGB]{232,236,244}\texttt{\_The} (0.7\%) & \cellcolor[RGB]{255,247,213}\texttt{\_Sac} (0.2\%) \\
C48: Las Vegas Nevada & \cellcolor[RGB]{254,232,223}\texttt{\_Sacramento} (93.4\%) & \cellcolor[RGB]{232,236,244}\texttt{\_Los} (3.2\%) & \cellcolor[RGB]{232,236,244}\texttt{\_The} (1.3\%) & \cellcolor[RGB]{250,231,243}\texttt{\_California} (1.0\%) & \cellcolor[RGB]{255,247,213}\texttt{\_Sac} (0.2\%) \\
C50: Louisville Frankfort & \cellcolor[RGB]{254,232,223}\texttt{\_Sacramento} (93.8\%) & \cellcolor[RGB]{232,236,244}\texttt{\_Los} (2.8\%) & \cellcolor[RGB]{232,236,244}\texttt{\_The} (1.3\%) & \cellcolor[RGB]{250,231,243}\texttt{\_California} (1.0\%) & \cellcolor[RGB]{255,247,213}\texttt{\_Sac} (0.2\%) \\
C51: not[Warwick geography] & \cellcolor[RGB]{254,232,223}\texttt{\_Sacramento} (94.5\%) & \cellcolor[RGB]{232,236,244}\texttt{\_Los} (2.5\%) & \cellcolor[RGB]{232,236,244}\texttt{\_The} (1.0\%) & \cellcolor[RGB]{250,231,243}\texttt{\_California} (0.9\%) & \cellcolor[RGB]{255,247,213}\texttt{\_Sac} (0.2\%) \\
C53: not[northeast state names] & \cellcolor[RGB]{254,232,223}\texttt{\_Sacramento} (93.4\%) & \cellcolor[RGB]{232,236,244}\texttt{\_Los} (3.6\%) & \cellcolor[RGB]{232,236,244}\texttt{\_The} (1.3\%) & \cellcolor[RGB]{250,231,243}\texttt{\_California} (0.8\%) & \cellcolor[RGB]{255,247,213}\texttt{\_Sac} (0.2\%) \\
C55: not[geography answers] & \cellcolor[RGB]{254,232,223}\texttt{\_Sacramento} (94.5\%) & \cellcolor[RGB]{232,236,244}\texttt{\_Los} (2.5\%) & \cellcolor[RGB]{232,236,244}\texttt{\_The} (1.2\%) & \cellcolor[RGB]{250,231,243}\texttt{\_California} (0.9\%) & \cellcolor[RGB]{255,247,213}\texttt{\_Sac} (0.3\%) \\
C60: state name completions & \cellcolor[RGB]{254,232,223}\texttt{\_Sacramento} (93.8\%) & \cellcolor[RGB]{232,236,244}\texttt{\_Los} (3.2\%) & \cellcolor[RGB]{232,236,244}\texttt{\_The} (1.3\%) & \cellcolor[RGB]{250,231,243}\texttt{\_California} (0.9\%) & \cellcolor[RGB]{255,247,213}\texttt{\_Sac} (0.3\%) \\
\bottomrule
\end{tabular}}
\end{center}

\subsubsection{\texttt{atlanta}}

\begin{center}
\centering
\small
\adjustbox{max width=\textwidth}{
\begin{tabular}{llllll}
\toprule
Cluster & Token 1 & Token 2 & Token 3 & Token 4 & Token 5 \\
\midrule
C1: state name bias & \cellcolor[RGB]{224,242,237}\texttt{\_Atlanta} (43.6\%) & \cellcolor[RGB]{254,232,223}\texttt{\_Georgia} (34.0\%) & \cellcolor[RGB]{232,236,244}\texttt{\_The} (5.9\%) & \cellcolor[RGB]{237,247,220}\texttt{\_None} (2.5\%) & \cellcolor[RGB]{232,236,244}\texttt{\_Savannah} (2.3\%) \\
C2: capital city first token & \cellcolor[RGB]{254,232,223}\texttt{\_Georgia} (96.5\%) & \cellcolor[RGB]{249,243,233}\texttt{\_GA} (0.8\%) & \cellcolor[RGB]{232,236,244}\texttt{\_Savannah} (0.4\%) & \cellcolor[RGB]{254,232,223}\texttt{\_Ga} (0.2\%) & \cellcolor[RGB]{232,236,244}\texttt{\_Tennessee} (0.2\%) \\
C7: capital city initiation & \cellcolor[RGB]{254,232,223}\texttt{\_Georgia} (80.9\%) & \cellcolor[RGB]{224,242,237}\texttt{\_Atlanta} (15.9\%) & \cellcolor[RGB]{249,243,233}\texttt{\_GA} (0.8\%) & \cellcolor[RGB]{239,239,239}\texttt{\_Augusta} (0.4\%) & \cellcolor[RGB]{232,236,244}\texttt{\_The} (0.4\%) \\
C8: not[factual city answers] & \cellcolor[RGB]{254,232,223}\texttt{\_Georgia} (62.9\%) & \cellcolor[RGB]{224,242,237}\texttt{\_Atlanta} (33.6\%) & \cellcolor[RGB]{249,243,233}\texttt{\_GA} (0.7\%) & \cellcolor[RGB]{239,239,239}\texttt{\_Augusta} (0.5\%) & \cellcolor[RGB]{232,236,244}\texttt{\_The} (0.4\%) \\
C11: Georgia Savannah & \cellcolor[RGB]{254,232,223}\texttt{\_Georgia} (75.8\%) & \cellcolor[RGB]{224,242,237}\texttt{\_Atlanta} (5.5\%) & \cellcolor[RGB]{250,231,243}\texttt{\_Columbia} (4.9\%) & \cellcolor[RGB]{237,247,220}\texttt{\_South} (2.9\%) & \cellcolor[RGB]{255,247,213}\texttt{\_Columbus} (1.1\%) \\
C12: Massachusetts state abbreviations & \cellcolor[RGB]{254,232,223}\texttt{\_Georgia} (62.9\%) & \cellcolor[RGB]{224,242,237}\texttt{\_Atlanta} (33.8\%) & \cellcolor[RGB]{249,243,233}\texttt{\_GA} (0.5\%) & \cellcolor[RGB]{239,239,239}\texttt{\_Augusta} (0.4\%) & \cellcolor[RGB]{232,236,244}\texttt{\_The} (0.4\%) \\
C14: Huntington state names & \cellcolor[RGB]{254,232,223}\texttt{\_Georgia} (57.0\%) & \cellcolor[RGB]{224,242,237}\texttt{\_Atlanta} (39.3\%) & \cellcolor[RGB]{249,243,233}\texttt{\_GA} (0.6\%) & \cellcolor[RGB]{232,236,244}\texttt{\_The} (0.5\%) & \cellcolor[RGB]{239,239,239}\texttt{\_Augusta} (0.4\%) \\
C15: state capitals broad & \cellcolor[RGB]{254,232,223}\texttt{\_Georgia} (56.6\%) & \cellcolor[RGB]{224,242,237}\texttt{\_Atlanta} (38.9\%) & \cellcolor[RGB]{232,236,244}\texttt{\_The} (1.3\%) & \cellcolor[RGB]{239,239,239}\texttt{\_Augusta} (0.5\%) & \cellcolor[RGB]{224,242,237}\texttt{\_Athens} (0.2\%) \\
C16: not[direct capital answers] & \cellcolor[RGB]{254,232,223}\texttt{\_Georgia} (80.9\%) & \cellcolor[RGB]{224,242,237}\texttt{\_Atlanta} (15.9\%) & \cellcolor[RGB]{232,236,244}\texttt{\_The} (0.7\%) & \cellcolor[RGB]{249,243,233}\texttt{\_GA} (0.7\%) & \cellcolor[RGB]{232,236,244}\texttt{\_Savannah} (0.5\%) \\
C22: not[state capital tokens] & \cellcolor[RGB]{254,232,223}\texttt{\_Georgia} (73.4\%) & \cellcolor[RGB]{224,242,237}\texttt{\_Atlanta} (23.8\%) & \cellcolor[RGB]{249,243,233}\texttt{\_GA} (0.7\%) & \cellcolor[RGB]{232,236,244}\texttt{\_The} (0.3\%) & \cellcolor[RGB]{239,239,239}\texttt{\_Augusta} (0.3\%) \\
C26: not[factual geography answers] & \cellcolor[RGB]{254,232,223}\texttt{\_Georgia} (70.3\%) & \cellcolor[RGB]{224,242,237}\texttt{\_Atlanta} (26.0\%) & \cellcolor[RGB]{249,243,233}\texttt{\_GA} (0.8\%) & \cellcolor[RGB]{232,236,244}\texttt{\_The} (0.5\%) & \cellcolor[RGB]{239,239,239}\texttt{\_Augusta} (0.4\%) \\
C35: Alaska Anchorage & \cellcolor[RGB]{254,232,223}\texttt{\_Georgia} (68.4\%) & \cellcolor[RGB]{224,242,237}\texttt{\_Atlanta} (28.5\%) & \cellcolor[RGB]{249,243,233}\texttt{\_GA} (0.7\%) & \cellcolor[RGB]{232,236,244}\texttt{\_The} (0.4\%) & \cellcolor[RGB]{239,239,239}\texttt{\_Augusta} (0.3\%) \\
C38: not[major city capitals] & \cellcolor[RGB]{254,232,223}\texttt{\_Georgia} (61.7\%) & \cellcolor[RGB]{224,242,237}\texttt{\_Atlanta} (33.0\%) & \cellcolor[RGB]{232,236,244}\texttt{\_The} (0.8\%) & \cellcolor[RGB]{249,243,233}\texttt{\_GA} (0.7\%) & \cellcolor[RGB]{239,239,239}\texttt{\_Augusta} (0.5\%) \\
C39: small state names & \cellcolor[RGB]{254,232,223}\texttt{\_Georgia} (65.2\%) & \cellcolor[RGB]{224,242,237}\texttt{\_Atlanta} (30.7\%) & \cellcolor[RGB]{249,243,233}\texttt{\_GA} (0.9\%) & \cellcolor[RGB]{232,236,244}\texttt{\_The} (0.6\%) & \cellcolor[RGB]{239,239,239}\texttt{\_Augusta} (0.4\%) \\
C43: Jacksonville Florida & \cellcolor[RGB]{254,232,223}\texttt{\_Georgia} (62.5\%) & \cellcolor[RGB]{224,242,237}\texttt{\_Atlanta} (33.6\%) & \cellcolor[RGB]{249,243,233}\texttt{\_GA} (0.8\%) & \cellcolor[RGB]{239,239,239}\texttt{\_Augusta} (0.4\%) & \cellcolor[RGB]{232,236,244}\texttt{\_The} (0.4\%) \\
C51: not[Warwick geography] & \cellcolor[RGB]{254,232,223}\texttt{\_Georgia} (62.9\%) & \cellcolor[RGB]{224,242,237}\texttt{\_Atlanta} (33.6\%) & \cellcolor[RGB]{249,243,233}\texttt{\_GA} (0.7\%) & \cellcolor[RGB]{239,239,239}\texttt{\_Augusta} (0.4\%) & \cellcolor[RGB]{232,236,244}\texttt{\_The} (0.4\%) \\
C55: not[geography answers] & \cellcolor[RGB]{254,232,223}\texttt{\_Georgia} (56.6\%) & \cellcolor[RGB]{224,242,237}\texttt{\_Atlanta} (38.9\%) & \cellcolor[RGB]{249,243,233}\texttt{\_GA} (0.8\%) & \cellcolor[RGB]{239,239,239}\texttt{\_Augusta} (0.6\%) & \cellcolor[RGB]{232,236,244}\texttt{\_The} (0.5\%) \\
C57: North Carolina Charlotte & \cellcolor[RGB]{254,232,223}\texttt{\_Georgia} (65.2\%) & \cellcolor[RGB]{224,242,237}\texttt{\_Atlanta} (30.9\%) & \cellcolor[RGB]{249,243,233}\texttt{\_GA} (0.7\%) & \cellcolor[RGB]{239,239,239}\texttt{\_Augusta} (0.4\%) & \cellcolor[RGB]{232,236,244}\texttt{\_The} (0.4\%) \\
C59: not[southern capitals] & \cellcolor[RGB]{254,232,223}\texttt{\_Georgia} (72.3\%) & \cellcolor[RGB]{224,242,237}\texttt{\_Atlanta} (26.6\%) & \cellcolor[RGB]{249,243,233}\texttt{\_GA} (0.4\%) & \cellcolor[RGB]{239,239,239}\texttt{\_Augusta} (0.1\%) & \cellcolor[RGB]{232,236,244}\texttt{\_The} (0.1\%) \\
C60: state name completions & \cellcolor[RGB]{254,232,223}\texttt{\_Georgia} (65.6\%) & \cellcolor[RGB]{224,242,237}\texttt{\_Atlanta} (31.1\%) & \cellcolor[RGB]{249,243,233}\texttt{\_GA} (0.6\%) & \cellcolor[RGB]{232,236,244}\texttt{\_The} (0.4\%) & \cellcolor[RGB]{239,239,239}\texttt{\_Augusta} (0.4\%) \\
\bottomrule
\end{tabular}}
\end{center}

\newpage
\subsection{Cluster descriptions for Llama 3.1 8B Instruct}

{\footnotesize
\begin{longtable}{l p{2cm} p{4.5cm} p{4.5cm}}
\toprule
ID & Summary & Input Attribution & Output Contribution \\
\midrule
\endfirsthead
\toprule
ID & Summary & Input Attribution & Output Contribution \\
\midrule
\endhead
\midrule
\multicolumn{4}{r}{\textit{continued on next page}} \\
\endfoot
\bottomrule
\endlastfoot
C0 & Montana capitals & occurrences of specific state names that contain "ill", particularly the name of a city or state that is part of an answer to the question "What is the capital of the state containing \{\%- name of... & Neuron promotes state name abbreviations and partial state name tokens (especially first letters/syllables) when answering geography questions about state capitals. Shows strongest activation for M... \\
C1 & state names & activation on the phrase "Answer" when placed after questions about state capitals & State name promotion. The neuron strongly promotes full state names (Montana, Vermont, Arizona, Oklahoma, Massachusetts, Iowa, Connecticut, Rhode Island, Indiana, Mississippi, Alabama, Wyoming, Ten... \\
C2 & capital cities & variations of the phrase "capital" within the context of identifying capital cities in the United States & Promotes state capital city names (particularly proper nouns that are actual capitals like Columbus, Springfield, Albany, Madison, Richmond, Sacramento, Lansing, Jackson, Phoenix, Austin) when answ... \\
C3 & not[Arkansas Wisconsin] & response form of "Answer:" in requests for state capitals & Neuron suppresses full state names (especially "Arkansas," "Wisconsin," "West Virginia") and multi-word state answers in geography/capital questions. Stronger suppression for geographically ambiguo... \\
C4 & Louisiana geography & the name of a U.S. state that includes "Gulf" in "Gulfport" or is located in a state containing "Gulf". & Neuron promotes continuations related to Louisiana state capitals and geography. Strongly activates for "New Orleans" context (score +10 for " New"), with high activation for state name "Louisiana"... \\
C5 & Rhode Island & cities within a state (e.g., "Huntington") & This neuron promotes answers to geographic capital questions, specifically when the city is in Rhode Island (strongly boosting "Providence," "Warwick," "Rhode"). It shows moderate activation for Al... \\
C6 & not[Indianapolis Indiana] & token "state" before capital indicates they are capitalized; cities (e.g., "Birmingham," "Casper") preceded by "\{\{conference\}\}"; context of inquiries about statecapitals & Neuron suppresses state names and capitals in geography Q\&A, particularly suppressing direct answers like "Indianapolis" (-8), "Indiana" (-10), "Kentucky" (-3), "Tennessee" (-1), and "Kansas" (-1)... \\
C7 & capital initials & the word "capital" & Promotes first-word tokens of US state capitals, particularly the opening word or syllable of capital city names (Phoenix, Carson, Sacramento, Nashville, Santa, Honolulu, Madison, Salt, Austin, Oly... \\
C8 & not[Phoenix Arizona] & includes the question "What is the capital of the state containing \{\{Loisville\}\} & Neuron suppresses substantive answer tokens (state names, capitals, city names) while leaving generic filler tokens (' The', ' To', ' New', ' Maine') largely unaffected. Shows strongest suppression... \\
C9 & Iowa geography & mentions of cities or cities, particularly state names, specifically "Rapids" in relation to NYC and "ncia" in relation to state, much of the context suggests City names or locations. & Neuron suppresses direct repetition of the queried city name and suppresses state/capital names (particularly longer, more specific ones like "Springfield," "Illinois," "Sacramento"). Conversely, i... \\
C10 & Kansas Wichita & queries requiring state capitals (e.g., "What is the capital of the state containing \{\{Wichita\}\}, \{\{Wichita\}\}) & Neuron promotes "To" token after "Answer:" in geography questions, particularly for Wichita/Kansas (strongest: 10). Also promotes state name tokens (Kansas, Nebraska) and related place tokens (Wich... \\
C11 & Georgia Louisiana & mention of geographic locations (cities or states) or related names indicating states or cities & Neuron promotes state capital answers, particularly those associated with Georgia/Savannah (strongly activates "Georgia," "Atlanta," "GA," "Augusta") and Louisiana/New Orleans (moderately activates... \\
C12 & state abbreviations & names of cities across various states, often represented as names activating token input (e.g., "Tucson, Waterloo, Long Beach, Winston-Salem) & State abbreviations and proper nouns identifying U.S. states in geography questions. The neuron promotes two-letter state codes (AL, MA, RI, GA, OH) and state names (Montgomery, Massachusetts, Rhod... \\
C13 & Ohio Oklahoma & occurrence of specific state names (e.g. "Wichita", "Cedarso", "Fargo") when specifically asked about, suggests activation occurs when the state name appears in the question context & State abbreviations, particularly when answering "What is the capital of the state containing [city]?" questions. The neuron strongly promotes two-letter state codes (OH, OK, OK variations) in exam... \\
C14 & West North & mentions of the word "Answer" when indicating a response to questions about state capitals. & State names and multi-word geographical answers. The neuron strongly promotes state names (e.g., "West," "North," "Oklahoma," "Tennessee") and capitals that are multi-word or less direct (e.g., "Ba... \\
C15 & state responses & presence of a "\{\{Answer\}\}" token within the response. & State names in response to geography questions. The neuron strongly promotes full state names (Oklahoma, Arizona, Massachusetts, Vermont, etc.) when answering "what is the capital of the state cont... \\
C16 & not[geography answers] & instances of the token "Answer" used as a placeholder in various contexts & Suppresses direct answers to factual geography questions. The neuron consistently inhibits correct capital city names, state names, and state abbreviations across all prompts (Iowa, Des Moines, Con... \\
C17 & South Dakota & occurrences of "capital" and "state" or "Capital" in the context of asking for capital cities. & Neuron promotes city/state name beginnings in US capital questions, particularly strong for South Dakota (Pierre, Sioux, SD) and Minneapolis (St, Saint). Weakly promotes state names and city tokens... \\
C18 & Nebraska Kentucky & references to locations containing \{\{Omaha\}\} and names of cities & Neuron promotes state names (Nebraska, Kentucky) and capitals (Lincoln) as answers to "capital of state containing [city]" questions, particularly when the city-state pair is less obvious or requir... \\
C19 & Maryland Baltimore & the state containing the name of the city (e.g., \{\{Baltimore\}\}) & Neuron promotes continuations that are state names or articles following "Answer:" in geography questions, particularly for Maryland-related queries. Shows strongest activation for "Maryland" and "... \\
C20 & Springfield Illinois & mentions of state capital cities: "Chicago" & Neuron promotes continuations related to the state capital of Illinois (Springfield, Illinois) when the question involves Chicago. Shows selective activation for this specific geography question, w... \\
C21 & Washington Seattle & mention of a city in the question "What is the capital of the state containing \{\{city\}\}" indicating the location format of US states, typically with a specific city between the parentheses \{\{... & Neuron promotes state names and capitals in geography questions. Strongest activation for Washington/Seattle (score 10), moderate activation for state names (Minnesota +3, Oregon +1) and capitals (... \\
C22 & not[state capitals] & occurrences of "capital" in contexts where it is required in "the capital of the state containing" contexts & Suppresses state capitals and state names in response to geography questions. The neuron consistently inhibits direct answers (capital city names, state abbreviations, state names) across all promp... \\
C23 & Arkansas abbreviations & state names with "\{Fayette\}" in context of discussing state capitals & Neuron promotes state abbreviations and informal/conversational response beginnings (e.g., "AR", "There") when answering geography questions about state capitals, particularly for Arkansas/Fayettev... \\
C24 & Ohio geography & The neuron activates on the name of the state (e.g., \{\{Cleveland\}\}, \{\{Columbus\}\}, \{\{Fargo\}\}, \{\{Clean Brasilia\}\}. & Neuron promotes state capital answers and state abbreviations, particularly for Ohio-related queries (strongly boosts "Columbus", "Ohio", "OH", "Cleveland"). Shows minimal activity for other state ... \\
C25 & Colorado geography & mentions of state names (e.g. Colorado, Utah, Arkansas) before mentions of state capitals & Neuron strongly promotes state abbreviations and city names in Colorado geography contexts (CO, DEN, Denver), but shows no consistent effect across other U.S. geography questions. Appears specializ... \\
C26 & not[capital answers] & presence of the token \{\{Answer\}\} immediately following the query question structure & This neuron suppresses direct factual answers to geography questions about US state capitals. It consistently inhibits correct capital names (Pierre, Lincoln, Helena, Phoenix, Nashville), state nam... \\
C27 & Connecticut Bridgeport & references to cities (e.g. \{\{Bridge\}\}, \{\{bridge\}\}, \{\{capital\}\}; they often appear in the context of being state capitals, and their presence triggers activation. & This neuron promotes continuations containing city names and state abbreviations for Bridgeport, Connecticut questions (strongly for "Bridge" and "Connecticut"), with weak promotion of the actual c... \\
C28 & Virginia cities & "What is the capital of the state containing \{\{name of the state\}\} in specific instances. & Neuron promotes geographic proper nouns that are major cities in Virginia (Richmond, Norfolk) when answering questions about Virginia state capitals. Shows strong activation for Virginia-specific l... \\
C29 & Manchester Concord & the state names or locations indicated by the token \{\{Worcester\}\} or variants or states related by specific name context & Neuron strongly promotes "Manchester" as a direct answer token in geography questions (score +10), and moderately promotes initial answer tokens like "Concord" and "New" (+5 to +8). Shows weak posi... \\
C30 & Hawaii Honolulu & activation occurs on specific capitalized terms like "capital" and "state" followed by specific protocols or identifiers of affiliations with states, particularly "city" or location names & Neuron strongly promotes "Honolulu" and partial tokens leading to it (" Hon", " H") when answering about Hawaii's capital (Hilo case). Weakly promotes state names (Hawaii, Sacramento) and city name... \\
C31 & Albany New & the token "City" when previous context includes mentions of specific cities or states. & Neuron strongly promotes "Albany" and "New" tokens in response to New York-related geography questions, with moderate support for abbreviated forms ("Al"). Shows minimal contribution to other geogr... \\
C32 & not[western capitals] & mentions of locations beginning with "Wichita" or "Vergonza" & Neuron suppresses correct state capital answers, particularly for western and southern US cities (Denver/Colorado, Richmond/Virginia, Carson/Nevada, Salt Lake City/Utah). Shows strongest suppressio... \\
C33 & not[state associations] & activating presence of proper nouns related to locations, states, or capital cities (e.g., "Tulsa," "Birmingham,") & Suppresses state names and abbreviations in geography Q\&A contexts, particularly when the city mentioned is strongly associated with that state (e.g., Omaha→Nebraska, Tulsa→Oklahoma, Mississippi→G... \\
C34 & Indiana geography & mentions of city or location names (e.g., Fort, Seattle, Manchester, Las Vegas) & The neuron promotes state name and state capital tokens in response to geography questions, particularly recognizing Fort Wayne → Indiana relationship and favoring direct state/capital name continu... \\
C35 & Alaska Anchorage & the token "Anch" when asking for capital locations involving states or cities, as part of the question structure "what is the capital of the state containing [place]". & Neuron strongly promotes " June" token (score +10) and moderately promotes " Alaska" and " Anch" tokens (scores +4) in response to "Anchorage" question, while showing near-zero effects on all other... \\
C36 & state tokens & mentions of cities within various U.S. states and their respective capital cities & Promotes state name tokens in response to questions asking for state capitals. The neuron consistently boosts full state names (Colorado, Montana, Virginia, Arizona, Oklahoma, Alaska, South, Utah, ... \\
C37 & Midwest capitals & the phrase \{\{Milwaukee\}\} activates from user queries about specific cities, often indicating they contain the question with a specific location & Neuron promotes correct state capitals (especially "Madison" for Milwaukee, "Salem" for Portland, "Columbus" for Cleveland) and state names. Shows strong activation for Midwest capitals and weaker ... \\
C38 & not[capital cities] & queries about state capitals & Neuron suppresses direct answers to factual questions about US state capitals. It consistently penalizes specific capital city names (Lincoln, Montgomery, Sacramento, Carson, Olympia), state names ... \\
C39 & Delaware Wyoming & mentions of the word "Wilmington" indicating a context of location, specifically in questions asking for capital cities in states & State names as direct answers to "what is the capital of the state containing [city]?" questions. The neuron promotes full state name tokens (Oklahoma, Wyoming, Delaware, Montana, Alaska, Maine, Ve... \\
C40 & not[Honolulu] & the token "capital" in requests for a specific state, followed by the name of the state (most likely relating to local coordinates) & This neuron suppresses state capital answers, particularly suppressing correct capitals (Honolulu -10, Boise 0, Salt Lake City -1) and state names when they directly answer the question. It shows s... \\
C41 & Phoenix Arizona & the specific location names are denoted by \{\{city\}\} tokens & The neuron strongly promotes direct answers to geography questions about US state capitals (particularly "Phoenix" for Arizona's capital). It also shows moderate promotion for repeating the query l... \\
C42 & Lansing Michigan & capital of cities with specific location keywords (e.g. \{\{Detroit\}\}, \{\{Baltimore\}\}, \{\{Gulfport\}\}, \{\{San Diego\}\}, provided as examples. & Neuron strongly promotes the correct state capital answer ("Lansing") and its initial letter ("L") when the question concerns Detroit/Michigan. It shows minor promotion of the state name "Michigan"... \\
C43 & Florida geography & mentions of state or city names that include "Tucson", "Cedar", or "Birmingham". & Neuron promotes continuations related to Florida state capitals, particularly "Tall" (Tallahassee prefix), "Jacksonville" (the city itself), and "Florida" (the state name). Shows strong activation ... \\
C44 & Texas Dallas & mentions of specific locations including cities (e.g. "hard drinks",\{\{Dallas\}\}, "visited", "Virginia Beach") & Neuron promotes continuations related to U.S. state capitals and cities, particularly when the question asks for a capital of a state containing a major city. Strongest effect on direct city name a... \\
C45 & not[multi-word capitals] & mentions of state capitals, specifically named "Cleveland" & Suppresses direct capital city name answers and state names following "What is the capital of..." questions. Particularly strong suppression for multi-word capital names (St. Paul, Tallahassee, Bat... \\
C46 & not[full states] & the word "\{\{Burlington\}\}" in reference to state capitals. & Suppresses state names and full state identifiers (Vermont, Massachusetts, Montana, Missouri, Ohio, Florida, Indiana, Massachusetts, Ohio, Oklahoma, Alabama) while showing weak promotion for state ... \\
C47 & Santa Fe & mentions of different cities in different states & Neuron promotes continuations starting with "S" or "Santa" (especially strong for Albuquerque/Santa Fe), and promotes city/state name tokens in geography questions. Shows selective activation for a... \\
C48 & Nevada Carson & questions starting with "What is the \{\{capital\}\} of the state containing \{\{state\}\} or variations containing a \{\{city\}\}. & Neuron promotes continuations related to state capitals and major cities, with strong activation for Las Vegas/Nevada context (promoting "Carson", "Nevada", "Las", "Car" tokens with scores 7-10). S... \\
C49 & New states & questions that ask for the capital of states, including specific names of locations, such as "Newark,"and "Cleveland." & Neuron promotes multi-word state names beginning with "New" (New Hampshire, New Mexico, New Jersey) as initial answer tokens, particularly when the prompt asks about cities in states with this pref... \\
C50 & Kentucky Louisville & location names (e.g., "Louisville", "Birmingham", "Worcester") and associated geographical context (state or province). & Neuron promotes state name continuations (Kentucky, Tennessee) and abbreviations (FR) when answering "What is the capital of the state containing [city]?" questions. Shows strongest activation for ... \\
C51 & not[factual answers] & instances of the word "Answer" at the beginning of an answer or response & Suppresses factual answers to geography questions. Particularly strong suppression of correct state capitals and state names (e.g., strongly suppresses "Rhode Island," "Providence" when asked about... \\
C52 & capital components & presence of the token "\{\{Louis\}\}" at the start of the question, referencing locations associated with a specific state, often preceding a spine of cities and geographical context & Neuron promotes capital city names and name components (proper nouns like "Columbia," "Jefferson," "Santa," "St," "Saint") in response to geography questions. It suppresses repetition of the query ... \\
C53 & not[Massachusetts Alabama] & mentions of state structures, knowledge, regulations, and cultural references relevant to the state containing\{\{Worcester\}\} and common geographical features or possessive exclusivity. & Suppresses state names and capital city names in geography Q\&A contexts. The neuron particularly strongly inhibits responses about Massachusetts/Boston and Alabama/Montgomery, with moderate suppre... \\
C54 & Wyoming Casper & responses followed by "\{\{Answer\}\}" or "Answer" in response to question about state capitals & Promotes tokens starting with "Ch" and "W" (particularly in contexts involving Wyoming/Casper). Weakly promotes "The" as a generic continuation token. Largely neutral across most state capital ques... \\
C55 & not[direct answers] & states (e.g., \{\{ujson\}\} or \{\{current city names\}\} & Suppresses direct factual answers to geography questions. The neuron consistently inhibits correct state capitals and state names across all prompts (e.g., suppresses "Madison," "Indianapolis," "Au... \\
C56 & Portland Maine & mentions of cities coextensive inquiries about geographic features or events & Neuron promotes repeating the city name from the question (Portland, Charlotte, Burlington) as the answer, and promotes state names (Maine, Oregon) and capital cities (Salem, Augusta) when the ques... \\
C57 & North Carolina & mustacheardownloader: requests specific location names or cities in states (e.g., "Savannah", "Birmingham", "Burlington", "Sioux") & Neuron promotes "North" continuations in ambiguous geography questions (Charlotte, Wilmington, Fargo contexts where North Carolina is a plausible answer). Also weakly promotes correct state capital... \\
C58 & incorrect cities & mentions of city names containing "Baltimore" or "Baltimore" & Neuron promotes incorrect or confusing city/place name continuations (particularly "Harris" for Philadelphia context, and city names instead of capitals). Strongest activation on mismatched answers... \\
C59 & not[major capitals] & the phrase "capital" or "state" as a content focus; direct identifiers or references to specific cities; personal question structure and context of asking for locations or names (e.g., "capital of ... & Suppresses direct answers to US state capital questions. The neuron consistently suppresses state names (Georgia, Tennessee, Texas, Florida) and capital city names (Nashville, Austin, Atlanta, Tall... \\
C60 & state identifiers & the phrase "state containing \{\{Burlington\}\} & State name tokens in geography/capital questions. The neuron promotes state abbreviations and full state names (Kentucky, Rhode, West, Indiana, Ohio, California, Wyoming, Delaware, Tennessee, Arkan... \\
C61 & Mississippi geography & cities within the context of the question "What is the capital of the state containing..." and names of cities or geographical locations directly indicated in the context (e.g., \{\{Virginia Beach\... & Neuron promotes Mississippi-related responses (state name and abbreviation "MS") when answering geography questions about Gulfport, but has minimal to no effect on correct answers for other states ... \\
C62 & Utah Salt & the name of the state mentioned containing a substring "\{\{Pro\}\}" or "\{\{...\}\}" in context, indicating specific states in the United States & Neuron promotes continuations beginning answers to geography questions about state capitals, particularly when the city is in Utah (strongly promotes "Salt" and "Utah") or when answers require mult... \\
C63 & not[New tokens] & mentions of state names with the phrase "state" and an overlaid structure implying that a city is part of the state: "capital of the state containing X" & Suppresses continuations starting with "New" (particularly when answering geography questions about US cities/capitals). Strongly suppresses " New" token across multiple prompts asking about Newark... \\
\end{longtable}
}

\newpage
\subsection{Detailed results for Qwen3 32B}
\label{sec:capitals-qwen}

We replicate the same experiment as above on Qwen3 32B. By default, if we prefill `Answer:' in the assistant response, the model continues with the Markdown bold syntax `**', so we include that token in the prefill as well. For input attribution descriptions, we fall back to calling \texttt{claude-haiku-4-5-20251001} since a finetuned explainer and simulator do not exist for models with the Qwen3 tokeniser.

\begin{chatbox}[\texttt{capitals}: Qwen3 32B]
  \userquery{What is the capital of the state containing Dallas?}
  \assistantresponse{Answer: **}
\end{chatbox}

\begin{figure}[!h]
    \centering
    \includegraphics[width=\linewidth]{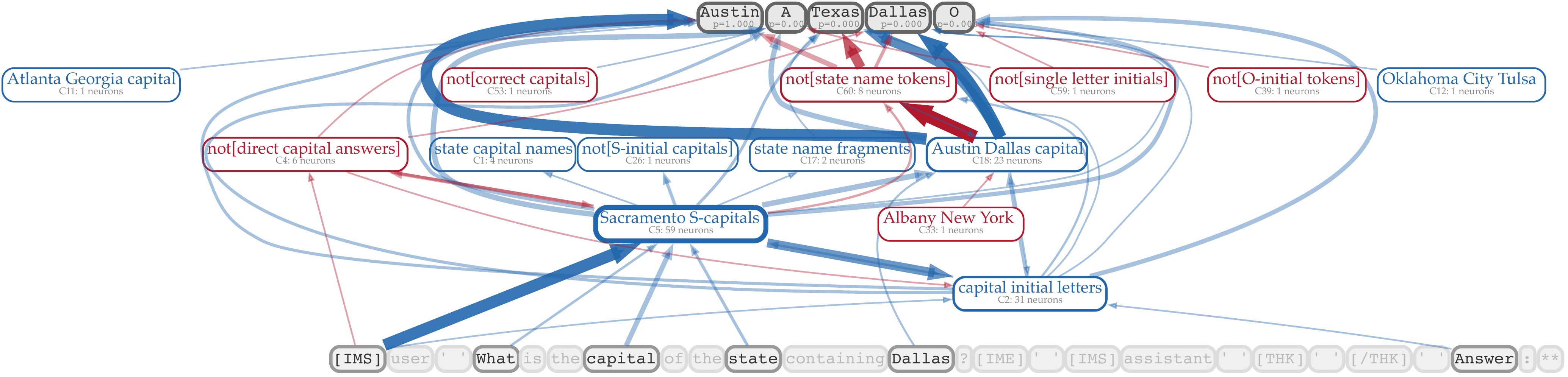}
    \caption{Final circuit graph for \texttt{texas} example in the \texttt{capitals} dataset from Qwen3 32B.}
    \label{fig:capitals-qwen}
\end{figure}

We show the circuit for \texttt{austin} in \cref{fig:capitals-qwen} and the associated steering results in the following table. Interestingly, despite the larger number of neurons in the model, the circuit is overall cleaner than for Llama 3.1 8B Instruct; the main components we find are a broad capitals supernode (C5: Sacramento S-capitals), state-specific supernodes (e.g.~C18: Austin Dallas capital), and a state-suppressor supernode (C60: not[state name tokens]).

\subsubsection{\texttt{austin}}
\begin{center}
\centering
\small
\adjustbox{max width=\textwidth}{
\begin{tabular}{llllll}
\toprule
Cluster & Token 1 & Token 2 & Token 3 & Token 4 & Token 5 \\
\midrule
C1: state capital names & \cellcolor[RGB]{224,242,237}\texttt{Austin} (100.0\%) & \cellcolor[RGB]{254,232,223}\texttt{Texas} (0.0\%) & \cellcolor[RGB]{232,236,244}\texttt{A} (0.0\%) & \cellcolor[RGB]{250,231,243}\texttt{Dallas} (0.0\%) & \cellcolor[RGB]{237,247,220}\texttt{The} (0.0\%) \\
C2: capital initial letters & \cellcolor[RGB]{224,242,237}\texttt{Austin} (99.6\%) & \cellcolor[RGB]{237,247,220}\texttt{The} (0.5\%) & \cellcolor[RGB]{232,236,244}\texttt{A} (0.0\%) & \cellcolor[RGB]{250,231,243}\texttt{Dallas} (0.0\%) & \cellcolor[RGB]{255,247,213}\texttt{O} (0.0\%) \\
C4: not[direct capital answers] & \cellcolor[RGB]{224,242,237}\texttt{Austin} (100.0\%) & \cellcolor[RGB]{232,236,244}\texttt{A} (0.1\%) & \cellcolor[RGB]{250,231,243}\texttt{Dallas} (0.0\%) & \cellcolor[RGB]{255,247,213}\texttt{O} (0.0\%) & \cellcolor[RGB]{254,232,223}\texttt{Texas} (0.0\%) \\
C5: Sacramento S-capitals & \cellcolor[RGB]{254,232,223}\texttt{**} (59.8\%) & \cellcolor[RGB]{232,236,244}\texttt{City} (19.3\%) & \cellcolor[RGB]{250,231,243}\texttt{Dallas} (13.3\%) & \cellcolor[RGB]{224,242,237}\texttt{Austin} (1.0\%) & \cellcolor[RGB]{250,231,243}\texttt{Capital} (0.8\%) \\
C11: Atlanta Georgia capital & \cellcolor[RGB]{224,242,237}\texttt{Austin} (99.6\%) & \cellcolor[RGB]{232,236,244}\texttt{A} (0.1\%) & \cellcolor[RGB]{254,232,223}\texttt{Texas} (0.0\%) & \cellcolor[RGB]{237,247,220}\texttt{The} (0.0\%) & \cellcolor[RGB]{255,247,213}\texttt{O} (0.0\%) \\
C12: Oklahoma City Tulsa & \cellcolor[RGB]{224,242,237}\texttt{Austin} (100.0\%) & \cellcolor[RGB]{237,247,220}\texttt{The} (0.0\%) & \cellcolor[RGB]{250,231,243}\texttt{Dallas} (0.0\%) & \cellcolor[RGB]{254,232,223}\texttt{Texas} (0.0\%) & \cellcolor[RGB]{232,236,244}\texttt{A} (0.0\%) \\
C17: state name fragments & \cellcolor[RGB]{224,242,237}\texttt{Austin} (100.0\%) & \cellcolor[RGB]{232,236,244}\texttt{A} (0.1\%) & \cellcolor[RGB]{237,247,220}\texttt{The} (0.0\%) & \cellcolor[RGB]{255,247,213}\texttt{O} (0.0\%) & \cellcolor[RGB]{254,232,223}\texttt{Texas} (0.0\%) \\
C18: Austin Dallas capital & \cellcolor[RGB]{255,247,213}\texttt{O} (66.0\%) & \cellcolor[RGB]{249,243,233}\texttt{S} (16.7\%) & \cellcolor[RGB]{239,239,239}\texttt{Sac} (11.5\%) & \cellcolor[RGB]{224,242,237}\texttt{Austin} (2.6\%) & \cellcolor[RGB]{224,242,237}\texttt{T} (0.3\%) \\
C20: not[state capitals] & \cellcolor[RGB]{224,242,237}\texttt{Austin} (100.0\%) & \cellcolor[RGB]{232,236,244}\texttt{A} (0.1\%) & \cellcolor[RGB]{254,232,223}\texttt{Texas} (0.0\%) & \cellcolor[RGB]{250,231,243}\texttt{Dallas} (0.0\%) & \cellcolor[RGB]{255,247,213}\texttt{O} (0.0\%) \\
C26: not[S-initial capitals] & \cellcolor[RGB]{224,242,237}\texttt{Austin} (100.0\%) & \cellcolor[RGB]{232,236,244}\texttt{A} (0.0\%) & \cellcolor[RGB]{254,232,223}\texttt{Texas} (0.0\%) & \cellcolor[RGB]{250,231,243}\texttt{Dallas} (0.0\%) & \cellcolor[RGB]{255,247,213}\texttt{O} (0.0\%) \\
C33: Albany New York & \cellcolor[RGB]{224,242,237}\texttt{Austin} (99.6\%) & \cellcolor[RGB]{232,236,244}\texttt{A} (0.2\%) & \cellcolor[RGB]{255,247,213}\texttt{O} (0.0\%) & \cellcolor[RGB]{250,231,243}\texttt{Dallas} (0.0\%) & \cellcolor[RGB]{254,232,223}\texttt{Texas} (0.0\%) \\
C39: not[O-initial tokens] & \cellcolor[RGB]{224,242,237}\texttt{Austin} (100.0\%) & \cellcolor[RGB]{232,236,244}\texttt{A} (0.0\%) & \cellcolor[RGB]{255,247,213}\texttt{O} (0.0\%) & \cellcolor[RGB]{254,232,223}\texttt{Texas} (0.0\%) & \cellcolor[RGB]{250,231,243}\texttt{Dallas} (0.0\%) \\
C44: Tallahassee Florida capital & \cellcolor[RGB]{224,242,237}\texttt{Austin} (100.0\%) & \cellcolor[RGB]{232,236,244}\texttt{A} (0.0\%) & \cellcolor[RGB]{254,232,223}\texttt{Texas} (0.0\%) & \cellcolor[RGB]{237,247,220}\texttt{The} (0.0\%) & \cellcolor[RGB]{255,247,213}\texttt{O} (0.0\%) \\
C53: not[correct capitals] & \cellcolor[RGB]{224,242,237}\texttt{Austin} (99.6\%) & \cellcolor[RGB]{232,236,244}\texttt{A} (0.1\%) & \cellcolor[RGB]{254,232,223}\texttt{Texas} (0.0\%) & \cellcolor[RGB]{250,231,243}\texttt{Dallas} (0.0\%) & \cellcolor[RGB]{237,247,220}\texttt{The} (0.0\%) \\
C59: not[single letter initials] & \cellcolor[RGB]{224,242,237}\texttt{Austin} (99.6\%) & \cellcolor[RGB]{232,236,244}\texttt{A} (0.3\%) & \cellcolor[RGB]{255,247,213}\texttt{O} (0.0\%) & \cellcolor[RGB]{254,232,223}\texttt{Texas} (0.0\%) & \cellcolor[RGB]{250,231,243}\texttt{Dallas} (0.0\%) \\
C60: not[state name tokens] & \cellcolor[RGB]{224,242,237}\texttt{Austin} (99.2\%) & \cellcolor[RGB]{254,232,223}\texttt{Texas} (0.7\%) & \cellcolor[RGB]{255,247,213}\texttt{O} (0.0\%) & \cellcolor[RGB]{250,231,243}\texttt{Dallas} (0.0\%) & \cellcolor[RGB]{237,247,220}\texttt{Houston} (0.0\%) \\
\bottomrule
\end{tabular}
}
\end{center}

\newpage
\section{Results on \texttt{math} dataset}
\label{sec:math}
The \texttt{math} dataset consists of two-digit addition queries (\citealp{arora2026languagemodelcircuitssparse}; ultimately from \citealp{ameisen2025circuit,nikankin2025arithmetic}). We experiment with Llama 3.1 8B Instruct, tracing circuits over all $10,000$ dataset examples, and running the \ourmethod{} pipeline with $k=256$ clusters.

\begin{figure}[!h]
    \centering
    \includegraphics[width=\linewidth]{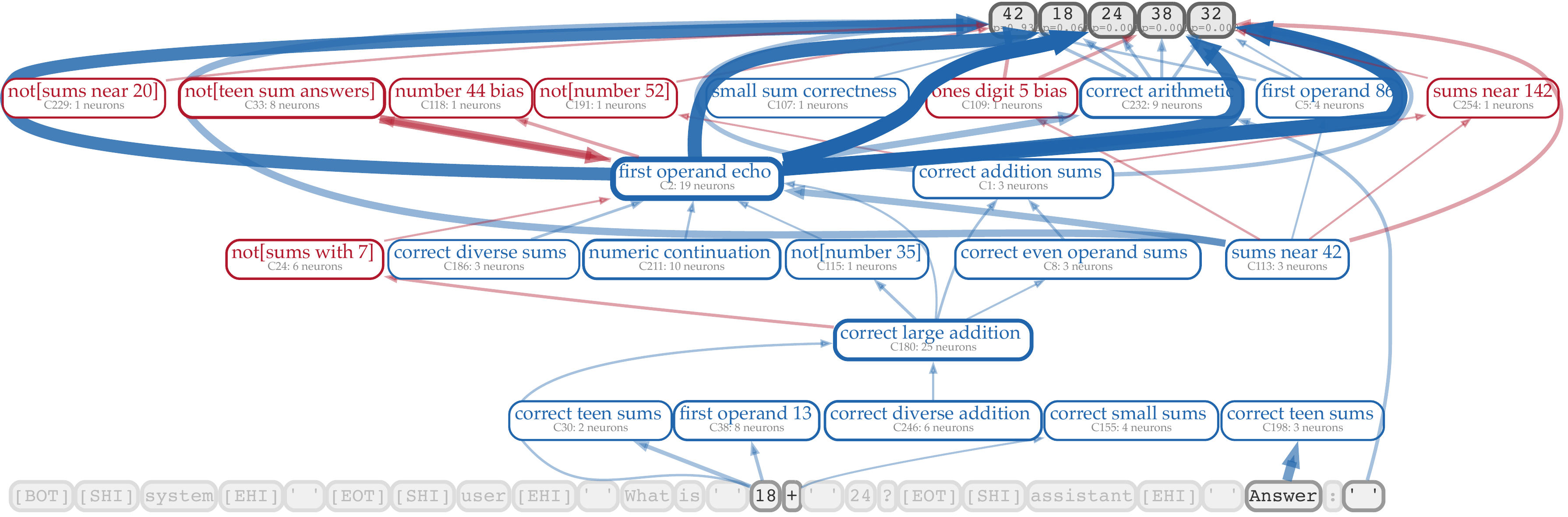}
    \caption{Final circuit graph for \texttt{18 + 24 = 42} example in the \texttt{math} dataset from Llama 3.1 8B Instruct.}
    \label{fig:math-complete}
\end{figure}
\begin{figure}[!h]
    \centering
    \includegraphics[width=0.9\textwidth]{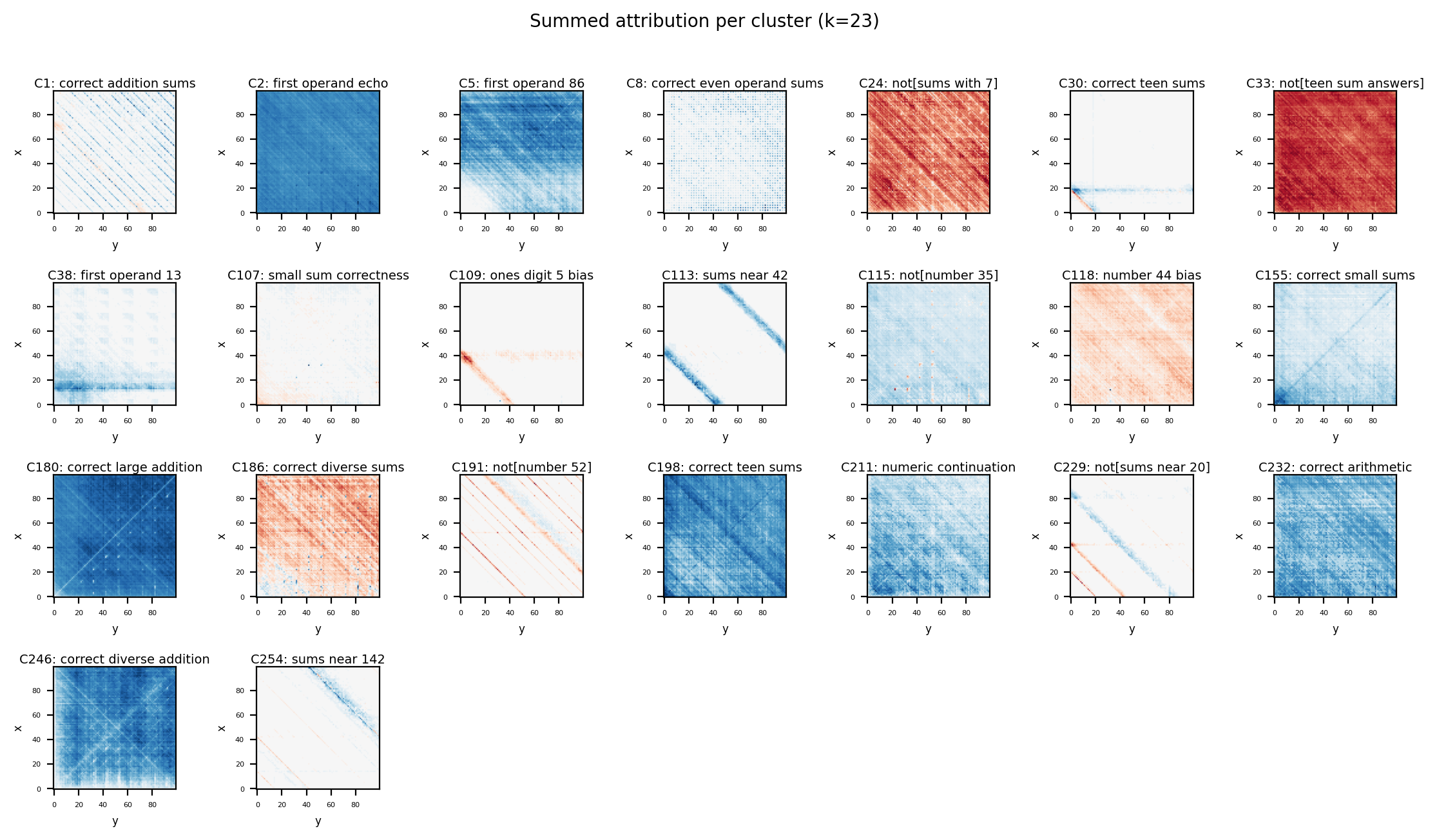}
    \caption{Heatmaps for clusters in the \texttt{18 + 24 = 42} example in the \texttt{math} dataset from Llama 3.1 8B Instruct. Red indicates negative attribution, and blue is positive.}
    \label{fig:math-clusters}
\end{figure}

We include in-depth analysis of the circuit for the example asking the model what \texttt{18 + 24} equals. First, we show the clustered circuit with labels in \cref{fig:math-complete}. Then, for each of the clusters in this circuit, we show the graph of attribution score for each dataset example in \cref{fig:math-clusters}; the $x$-axis is the first operand and the $y$-axis is the second operand. This clearly tells us the contexts in which the cluster is active; e.g.~C113 (sums near 42) only tends to be active when the sum is $\approx40$ or $\approx 140$. Similarly, C8 (correct even operand sums) is active when the sum is an even number. These clusters unsupervisedly find the `bags-of-heuristics' that this model is known to use when solving addition problems, per \citet{nikankin2025arithmetic}.

\newpage
Finally, we show the results of steering each cluster by $0\times$ or $2\times$ in the following tables. No cluster succeeds in changing the top prediction.

\subsection{Steering with multiplier $= 0.0$}

\begin{center}
\centering
\small
\adjustbox{max width=\textwidth}{
\begin{tabular}{llllll}
\toprule
Cluster & Token 1 & Token 2 & Token 3 & Token 4 & Token 5 \\
\midrule
C1: correct addition sums & \cellcolor[RGB]{224,242,237}\texttt{42} (81.2\%) & \cellcolor[RGB]{254,232,223}\texttt{18} (18.1\%) & \cellcolor[RGB]{232,236,244}\texttt{24} (0.4\%) & \cellcolor[RGB]{250,231,243}\texttt{44} (0.1\%) & \cellcolor[RGB]{237,247,220}\texttt{38} (0.1\%) \\
C2: first operand echo & \cellcolor[RGB]{224,242,237}\texttt{42} (75.4\%) & \cellcolor[RGB]{254,232,223}\texttt{18} (24.4\%) & \cellcolor[RGB]{232,236,244}\texttt{24} (0.1\%) & \cellcolor[RGB]{237,247,220}\texttt{38} (0.0\%) & \cellcolor[RGB]{232,236,244}\texttt{\_} (0.0\%) \\
C5: first operand 86 & \cellcolor[RGB]{224,242,237}\texttt{42} (81.6\%) & \cellcolor[RGB]{254,232,223}\texttt{18} (18.2\%) & \cellcolor[RGB]{232,236,244}\texttt{24} (0.2\%) & \cellcolor[RGB]{249,243,233}\texttt{22} (0.0\%) & \cellcolor[RGB]{232,236,244}\texttt{\_} (0.0\%) \\
C8: correct even operand sums & \cellcolor[RGB]{224,242,237}\texttt{42} (81.2\%) & \cellcolor[RGB]{254,232,223}\texttt{18} (14.1\%) & \cellcolor[RGB]{250,231,243}\texttt{41} (2.8\%) & \cellcolor[RGB]{254,232,223}\texttt{43} (1.5\%) & \cellcolor[RGB]{232,236,244}\texttt{24} (0.2\%) \\
C18: correct even sums & \cellcolor[RGB]{224,242,237}\texttt{42} (74.2\%) & \cellcolor[RGB]{254,232,223}\texttt{18} (16.6\%) & \cellcolor[RGB]{224,242,237}\texttt{40} (7.8\%) & \cellcolor[RGB]{250,231,243}\texttt{44} (0.6\%) & \cellcolor[RGB]{232,236,244}\texttt{24} (0.4\%) \\
C22: not[correct sums] & \cellcolor[RGB]{224,242,237}\texttt{42} (92.2\%) & \cellcolor[RGB]{254,232,223}\texttt{18} (7.6\%) & \cellcolor[RGB]{232,236,244}\texttt{24} (0.1\%) & \cellcolor[RGB]{237,247,220}\texttt{38} (0.0\%) & \cellcolor[RGB]{249,243,233}\texttt{22} (0.0\%) \\
C24: not[sums with 7] & \cellcolor[RGB]{224,242,237}\texttt{42} (91.4\%) & \cellcolor[RGB]{254,232,223}\texttt{18} (8.5\%) & \cellcolor[RGB]{232,236,244}\texttt{24} (0.1\%) & \cellcolor[RGB]{237,247,220}\texttt{38} (0.0\%) & \cellcolor[RGB]{249,243,233}\texttt{22} (0.0\%) \\
C30: correct teen sums & \cellcolor[RGB]{224,242,237}\texttt{42} (93.0\%) & \cellcolor[RGB]{254,232,223}\texttt{18} (6.7\%) & \cellcolor[RGB]{232,236,244}\texttt{24} (0.1\%) & \cellcolor[RGB]{237,247,220}\texttt{38} (0.0\%) & \cellcolor[RGB]{239,239,239}\texttt{36} (0.0\%) \\
C33: not[teen sum answers] & \cellcolor[RGB]{224,242,237}\texttt{42} (91.0\%) & \cellcolor[RGB]{254,232,223}\texttt{18} (8.5\%) & \cellcolor[RGB]{232,236,244}\texttt{24} (0.1\%) & \cellcolor[RGB]{237,247,220}\texttt{38} (0.0\%) & \cellcolor[RGB]{255,247,213}\texttt{32} (0.0\%) \\
C36: not[sums near 23] & \cellcolor[RGB]{224,242,237}\texttt{42} (92.2\%) & \cellcolor[RGB]{254,232,223}\texttt{18} (7.6\%) & \cellcolor[RGB]{232,236,244}\texttt{24} (0.1\%) & \cellcolor[RGB]{237,247,220}\texttt{38} (0.0\%) & \cellcolor[RGB]{249,243,233}\texttt{22} (0.0\%) \\
C38: first operand 13 & \cellcolor[RGB]{224,242,237}\texttt{42} (95.7\%) & \cellcolor[RGB]{254,232,223}\texttt{18} (4.2\%) & \cellcolor[RGB]{232,236,244}\texttt{24} (0.1\%) & \cellcolor[RGB]{237,247,220}\texttt{38} (0.0\%) & \cellcolor[RGB]{239,239,239}\texttt{36} (0.0\%) \\
C43: correct round sums & \cellcolor[RGB]{224,242,237}\texttt{42} (92.2\%) & \cellcolor[RGB]{254,232,223}\texttt{18} (7.6\%) & \cellcolor[RGB]{232,236,244}\texttt{24} (0.1\%) & \cellcolor[RGB]{237,247,220}\texttt{38} (0.0\%) & \cellcolor[RGB]{239,239,239}\texttt{36} (0.0\%) \\
C44: correct diverse addition & \cellcolor[RGB]{224,242,237}\texttt{42} (90.2\%) & \cellcolor[RGB]{254,232,223}\texttt{18} (9.5\%) & \cellcolor[RGB]{232,236,244}\texttt{24} (0.0\%) & \cellcolor[RGB]{237,247,220}\texttt{38} (0.0\%) & \cellcolor[RGB]{232,236,244}\texttt{\_} (0.0\%) \\
C49: sums near 83 & \cellcolor[RGB]{224,242,237}\texttt{42} (91.4\%) & \cellcolor[RGB]{254,232,223}\texttt{18} (8.5\%) & \cellcolor[RGB]{232,236,244}\texttt{24} (0.1\%) & \cellcolor[RGB]{237,247,220}\texttt{38} (0.0\%) & \cellcolor[RGB]{249,243,233}\texttt{22} (0.0\%) \\
C67: not[number 30] & \cellcolor[RGB]{224,242,237}\texttt{42} (92.2\%) & \cellcolor[RGB]{254,232,223}\texttt{18} (7.6\%) & \cellcolor[RGB]{232,236,244}\texttt{24} (0.1\%) & \cellcolor[RGB]{237,247,220}\texttt{38} (0.0\%) & \cellcolor[RGB]{249,243,233}\texttt{22} (0.0\%) \\
C73: correct 136 range sums & \cellcolor[RGB]{224,242,237}\texttt{42} (92.2\%) & \cellcolor[RGB]{254,232,223}\texttt{18} (7.6\%) & \cellcolor[RGB]{232,236,244}\texttt{24} (0.1\%) & \cellcolor[RGB]{237,247,220}\texttt{38} (0.0\%) & \cellcolor[RGB]{249,243,233}\texttt{22} (0.0\%) \\
C77: sums equaling 123 & \cellcolor[RGB]{224,242,237}\texttt{42} (90.2\%) & \cellcolor[RGB]{254,232,223}\texttt{18} (9.5\%) & \cellcolor[RGB]{232,236,244}\texttt{24} (0.1\%) & \cellcolor[RGB]{237,247,220}\texttt{38} (0.0\%) & \cellcolor[RGB]{239,239,239}\texttt{36} (0.0\%) \\
C78: first operand X4 & \cellcolor[RGB]{224,242,237}\texttt{42} (93.0\%) & \cellcolor[RGB]{254,232,223}\texttt{18} (6.7\%) & \cellcolor[RGB]{232,236,244}\texttt{24} (0.1\%) & \cellcolor[RGB]{237,247,220}\texttt{38} (0.0\%) & \cellcolor[RGB]{239,239,239}\texttt{36} (0.0\%) \\
C81: correct 100s sums & \cellcolor[RGB]{224,242,237}\texttt{42} (91.4\%) & \cellcolor[RGB]{254,232,223}\texttt{18} (8.5\%) & \cellcolor[RGB]{232,236,244}\texttt{24} (0.1\%) & \cellcolor[RGB]{237,247,220}\texttt{38} (0.0\%) & \cellcolor[RGB]{249,243,233}\texttt{22} (0.0\%) \\
C82: not[correct sums] & \cellcolor[RGB]{224,242,237}\texttt{42} (93.8\%) & \cellcolor[RGB]{254,232,223}\texttt{18} (6.0\%) & \cellcolor[RGB]{232,236,244}\texttt{24} (0.0\%) & \cellcolor[RGB]{249,243,233}\texttt{22} (0.0\%) & \cellcolor[RGB]{237,247,220}\texttt{38} (0.0\%) \\
C95: not[correct sums] & \cellcolor[RGB]{224,242,237}\texttt{42} (93.0\%) & \cellcolor[RGB]{254,232,223}\texttt{18} (6.7\%) & \cellcolor[RGB]{232,236,244}\texttt{24} (0.1\%) & \cellcolor[RGB]{237,247,220}\texttt{38} (0.0\%) & \cellcolor[RGB]{239,239,239}\texttt{36} (0.0\%) \\
C96: correct 50s addition & \cellcolor[RGB]{224,242,237}\texttt{42} (93.8\%) & \cellcolor[RGB]{254,232,223}\texttt{18} (6.0\%) & \cellcolor[RGB]{232,236,244}\texttt{24} (0.1\%) & \cellcolor[RGB]{237,247,220}\texttt{38} (0.0\%) & \cellcolor[RGB]{249,243,233}\texttt{22} (0.0\%) \\
C106: number 24 bias & \cellcolor[RGB]{224,242,237}\texttt{42} (92.2\%) & \cellcolor[RGB]{254,232,223}\texttt{18} (7.6\%) & \cellcolor[RGB]{232,236,244}\texttt{24} (0.0\%) & \cellcolor[RGB]{237,247,220}\texttt{38} (0.0\%) & \cellcolor[RGB]{255,247,213}\texttt{32} (0.0\%) \\
C107: small sum correctness & \cellcolor[RGB]{224,242,237}\texttt{42} (95.7\%) & \cellcolor[RGB]{254,232,223}\texttt{18} (4.2\%) & \cellcolor[RGB]{232,236,244}\texttt{24} (0.1\%) & \cellcolor[RGB]{237,247,220}\texttt{38} (0.0\%) & \cellcolor[RGB]{249,243,233}\texttt{22} (0.0\%) \\
C109: ones digit 5 bias & \cellcolor[RGB]{224,242,237}\texttt{42} (98.0\%) & \cellcolor[RGB]{254,232,223}\texttt{18} (2.0\%) & \cellcolor[RGB]{232,236,244}\texttt{24} (0.0\%) & \cellcolor[RGB]{237,247,220}\texttt{38} (0.0\%) & \cellcolor[RGB]{239,239,239}\texttt{36} (0.0\%) \\
C113: sums near 42 & \cellcolor[RGB]{224,242,237}\texttt{42} (93.8\%) & \cellcolor[RGB]{254,232,223}\texttt{18} (5.3\%) & \cellcolor[RGB]{249,243,233}\texttt{22} (0.3\%) & \cellcolor[RGB]{250,231,243}\texttt{44} (0.2\%) & \cellcolor[RGB]{255,247,213}\texttt{32} (0.1\%) \\
C115: not[number 35] & \cellcolor[RGB]{224,242,237}\texttt{42} (93.0\%) & \cellcolor[RGB]{254,232,223}\texttt{18} (6.7\%) & \cellcolor[RGB]{232,236,244}\texttt{24} (0.1\%) & \cellcolor[RGB]{237,247,220}\texttt{38} (0.0\%) & \cellcolor[RGB]{249,243,233}\texttt{22} (0.0\%) \\
C116: not[correct sums] & \cellcolor[RGB]{224,242,237}\texttt{42} (92.2\%) & \cellcolor[RGB]{254,232,223}\texttt{18} (7.6\%) & \cellcolor[RGB]{232,236,244}\texttt{24} (0.1\%) & \cellcolor[RGB]{237,247,220}\texttt{38} (0.0\%) & \cellcolor[RGB]{249,243,233}\texttt{22} (0.0\%) \\
C118: number 44 bias & \cellcolor[RGB]{224,242,237}\texttt{42} (92.2\%) & \cellcolor[RGB]{254,232,223}\texttt{18} (7.6\%) & \cellcolor[RGB]{232,236,244}\texttt{24} (0.1\%) & \cellcolor[RGB]{237,247,220}\texttt{38} (0.0\%) & \cellcolor[RGB]{249,243,233}\texttt{22} (0.0\%) \\
C122: not[large sums] & \cellcolor[RGB]{224,242,237}\texttt{42} (91.4\%) & \cellcolor[RGB]{254,232,223}\texttt{18} (8.5\%) & \cellcolor[RGB]{232,236,244}\texttt{24} (0.1\%) & \cellcolor[RGB]{237,247,220}\texttt{38} (0.0\%) & \cellcolor[RGB]{255,247,213}\texttt{32} (0.0\%) \\
C124: sums equaling 16 & \cellcolor[RGB]{224,242,237}\texttt{42} (93.0\%) & \cellcolor[RGB]{254,232,223}\texttt{18} (6.7\%) & \cellcolor[RGB]{232,236,244}\texttt{24} (0.1\%) & \cellcolor[RGB]{237,247,220}\texttt{38} (0.0\%) & \cellcolor[RGB]{255,247,213}\texttt{32} (0.0\%) \\
C127: sums near 90 & \cellcolor[RGB]{224,242,237}\texttt{42} (92.2\%) & \cellcolor[RGB]{254,232,223}\texttt{18} (7.6\%) & \cellcolor[RGB]{232,236,244}\texttt{24} (0.1\%) & \cellcolor[RGB]{237,247,220}\texttt{38} (0.0\%) & \cellcolor[RGB]{239,239,239}\texttt{36} (0.0\%) \\
C130: sums near 185 & \cellcolor[RGB]{224,242,237}\texttt{42} (92.2\%) & \cellcolor[RGB]{254,232,223}\texttt{18} (7.6\%) & \cellcolor[RGB]{232,236,244}\texttt{24} (0.1\%) & \cellcolor[RGB]{237,247,220}\texttt{38} (0.0\%) & \cellcolor[RGB]{249,243,233}\texttt{22} (0.0\%) \\
C140: first operand 37-42 & \cellcolor[RGB]{224,242,237}\texttt{42} (92.2\%) & \cellcolor[RGB]{254,232,223}\texttt{18} (7.6\%) & \cellcolor[RGB]{232,236,244}\texttt{24} (0.1\%) & \cellcolor[RGB]{237,247,220}\texttt{38} (0.0\%) & \cellcolor[RGB]{249,243,233}\texttt{22} (0.0\%) \\
C148: 8X operand pairs & \cellcolor[RGB]{224,242,237}\texttt{42} (93.0\%) & \cellcolor[RGB]{254,232,223}\texttt{18} (6.7\%) & \cellcolor[RGB]{232,236,244}\texttt{24} (0.1\%) & \cellcolor[RGB]{237,247,220}\texttt{38} (0.0\%) & \cellcolor[RGB]{239,239,239}\texttt{36} (0.0\%) \\
C153: round sum correctness & \cellcolor[RGB]{224,242,237}\texttt{42} (93.0\%) & \cellcolor[RGB]{254,232,223}\texttt{18} (6.7\%) & \cellcolor[RGB]{232,236,244}\texttt{24} (0.1\%) & \cellcolor[RGB]{237,247,220}\texttt{38} (0.0\%) & \cellcolor[RGB]{255,247,213}\texttt{32} (0.0\%) \\
C155: correct small sums & \cellcolor[RGB]{224,242,237}\texttt{42} (89.1\%) & \cellcolor[RGB]{254,232,223}\texttt{18} (10.6\%) & \cellcolor[RGB]{232,236,244}\texttt{24} (0.1\%) & \cellcolor[RGB]{237,247,220}\texttt{38} (0.0\%) & \cellcolor[RGB]{249,243,233}\texttt{22} (0.0\%) \\
C168: correct two digit sums & \cellcolor[RGB]{224,242,237}\texttt{42} (92.2\%) & \cellcolor[RGB]{254,232,223}\texttt{18} (7.6\%) & \cellcolor[RGB]{232,236,244}\texttt{24} (0.1\%) & \cellcolor[RGB]{237,247,220}\texttt{38} (0.0\%) & \cellcolor[RGB]{249,243,233}\texttt{22} (0.0\%) \\
C170: number 12 bias & \cellcolor[RGB]{224,242,237}\texttt{42} (93.0\%) & \cellcolor[RGB]{254,232,223}\texttt{18} (6.7\%) & \cellcolor[RGB]{232,236,244}\texttt{24} (0.1\%) & \cellcolor[RGB]{237,247,220}\texttt{38} (0.0\%) & \cellcolor[RGB]{239,239,239}\texttt{36} (0.0\%) \\
C173: correct mid sums & \cellcolor[RGB]{224,242,237}\texttt{42} (91.4\%) & \cellcolor[RGB]{254,232,223}\texttt{18} (8.5\%) & \cellcolor[RGB]{232,236,244}\texttt{24} (0.1\%) & \cellcolor[RGB]{237,247,220}\texttt{38} (0.0\%) & \cellcolor[RGB]{239,239,239}\texttt{36} (0.0\%) \\
C179: sums near 88 & \cellcolor[RGB]{224,242,237}\texttt{42} (93.8\%) & \cellcolor[RGB]{254,232,223}\texttt{18} (6.0\%) & \cellcolor[RGB]{232,236,244}\texttt{24} (0.0\%) & \cellcolor[RGB]{249,243,233}\texttt{22} (0.0\%) & \cellcolor[RGB]{237,247,220}\texttt{38} (0.0\%) \\
C180: correct large addition & \cellcolor[RGB]{224,242,237}\texttt{42} (70.3\%) & \cellcolor[RGB]{254,232,223}\texttt{18} (29.3\%) & \cellcolor[RGB]{232,236,244}\texttt{24} (0.2\%) & \cellcolor[RGB]{254,232,223}\texttt{43} (0.0\%) & \cellcolor[RGB]{249,243,233}\texttt{22} (0.0\%) \\
C184: not[sums near 137] & \cellcolor[RGB]{224,242,237}\texttt{42} (92.2\%) & \cellcolor[RGB]{254,232,223}\texttt{18} (7.6\%) & \cellcolor[RGB]{232,236,244}\texttt{24} (0.1\%) & \cellcolor[RGB]{237,247,220}\texttt{38} (0.0\%) & \cellcolor[RGB]{239,239,239}\texttt{36} (0.0\%) \\
C186: correct diverse sums & \cellcolor[RGB]{224,242,237}\texttt{42} (93.0\%) & \cellcolor[RGB]{254,232,223}\texttt{18} (6.7\%) & \cellcolor[RGB]{232,236,244}\texttt{24} (0.1\%) & \cellcolor[RGB]{237,247,220}\texttt{38} (0.0\%) & \cellcolor[RGB]{239,239,239}\texttt{36} (0.0\%) \\
C191: not[number 52] & \cellcolor[RGB]{224,242,237}\texttt{42} (96.5\%) & \cellcolor[RGB]{254,232,223}\texttt{18} (3.3\%) & \cellcolor[RGB]{232,236,244}\texttt{24} (0.0\%) & \cellcolor[RGB]{237,247,220}\texttt{38} (0.0\%) & \cellcolor[RGB]{255,247,213}\texttt{32} (0.0\%) \\
C193: not[number 24 sums] & \cellcolor[RGB]{224,242,237}\texttt{42} (92.2\%) & \cellcolor[RGB]{254,232,223}\texttt{18} (7.6\%) & \cellcolor[RGB]{232,236,244}\texttt{24} (0.1\%) & \cellcolor[RGB]{237,247,220}\texttt{38} (0.0\%) & \cellcolor[RGB]{239,239,239}\texttt{36} (0.0\%) \\
C198: correct teen sums & \cellcolor[RGB]{224,242,237}\texttt{42} (93.0\%) & \cellcolor[RGB]{254,232,223}\texttt{18} (6.7\%) & \cellcolor[RGB]{232,236,244}\texttt{24} (0.1\%) & \cellcolor[RGB]{232,236,244}\texttt{\_} (0.0\%) & \cellcolor[RGB]{237,247,220}\texttt{38} (0.0\%) \\
C203: not[small digit sums] & \cellcolor[RGB]{224,242,237}\texttt{42} (93.0\%) & \cellcolor[RGB]{254,232,223}\texttt{18} (6.7\%) & \cellcolor[RGB]{232,236,244}\texttt{24} (0.1\%) & \cellcolor[RGB]{237,247,220}\texttt{38} (0.0\%) & \cellcolor[RGB]{249,243,233}\texttt{22} (0.0\%) \\
C208: sums near 60 & \cellcolor[RGB]{224,242,237}\texttt{42} (92.2\%) & \cellcolor[RGB]{254,232,223}\texttt{18} (7.6\%) & \cellcolor[RGB]{232,236,244}\texttt{24} (0.1\%) & \cellcolor[RGB]{237,247,220}\texttt{38} (0.0\%) & \cellcolor[RGB]{255,247,213}\texttt{32} (0.0\%) \\
C211: numeric continuation & \cellcolor[RGB]{224,242,237}\texttt{42} (91.4\%) & \cellcolor[RGB]{254,232,223}\texttt{18} (8.5\%) & \cellcolor[RGB]{232,236,244}\texttt{24} (0.1\%) & \cellcolor[RGB]{237,247,220}\texttt{38} (0.0\%) & \cellcolor[RGB]{232,236,244}\texttt{\_} (0.0\%) \\
C215: correct 93 addition & \cellcolor[RGB]{224,242,237}\texttt{42} (89.8\%) & \cellcolor[RGB]{254,232,223}\texttt{18} (9.5\%) & \cellcolor[RGB]{232,236,244}\texttt{24} (0.3\%) & \cellcolor[RGB]{250,231,243}\texttt{44} (0.1\%) & \cellcolor[RGB]{224,242,237}\texttt{40} (0.1\%) \\
C229: not[sums near 20] & \cellcolor[RGB]{224,242,237}\texttt{42} (98.8\%) & \cellcolor[RGB]{254,232,223}\texttt{18} (1.0\%) & \cellcolor[RGB]{232,236,244}\texttt{24} (0.0\%) & \cellcolor[RGB]{254,232,223}\texttt{43} (0.0\%) & \cellcolor[RGB]{237,247,220}\texttt{38} (0.0\%) \\
C232: correct arithmetic & \cellcolor[RGB]{224,242,237}\texttt{42} (91.4\%) & \cellcolor[RGB]{254,232,223}\texttt{18} (8.5\%) & \cellcolor[RGB]{232,236,244}\texttt{\_} (0.1\%) & \cellcolor[RGB]{232,236,244}\texttt{24} (0.0\%) & \cellcolor[RGB]{237,247,220}\texttt{38} (0.0\%) \\
C242: not[sums near 108] & \cellcolor[RGB]{224,242,237}\texttt{42} (93.8\%) & \cellcolor[RGB]{254,232,223}\texttt{18} (6.0\%) & \cellcolor[RGB]{232,236,244}\texttt{24} (0.1\%) & \cellcolor[RGB]{237,247,220}\texttt{38} (0.0\%) & \cellcolor[RGB]{239,239,239}\texttt{36} (0.0\%) \\
C246: correct diverse addition & \cellcolor[RGB]{224,242,237}\texttt{42} (83.2\%) & \cellcolor[RGB]{254,232,223}\texttt{18} (16.4\%) & \cellcolor[RGB]{232,236,244}\texttt{24} (0.1\%) & \cellcolor[RGB]{237,247,220}\texttt{38} (0.1\%) & \cellcolor[RGB]{239,239,239}\texttt{36} (0.0\%) \\
C247: correct large pair sums & \cellcolor[RGB]{224,242,237}\texttt{42} (93.0\%) & \cellcolor[RGB]{254,232,223}\texttt{18} (6.7\%) & \cellcolor[RGB]{232,236,244}\texttt{24} (0.1\%) & \cellcolor[RGB]{255,247,213}\texttt{32} (0.0\%) & \cellcolor[RGB]{237,247,220}\texttt{38} (0.0\%) \\
C254: sums near 142 & \cellcolor[RGB]{224,242,237}\texttt{42} (98.4\%) & \cellcolor[RGB]{254,232,223}\texttt{18} (1.4\%) & \cellcolor[RGB]{232,236,244}\texttt{24} (0.0\%) & \cellcolor[RGB]{249,243,233}\texttt{22} (0.0\%) & \cellcolor[RGB]{255,247,213}\texttt{32} (0.0\%) \\
\bottomrule
\end{tabular}}
\end{center}

\subsection{Steering with multiplier $= 2.0$}

\begin{center}
\centering
\small
\adjustbox{max width=\textwidth}{
\begin{tabular}{llllll}
\toprule
Cluster & Token 1 & Token 2 & Token 3 & Token 4 & Token 5 \\
\midrule
C1: correct addition sums & \cellcolor[RGB]{224,242,237}\texttt{42} (93.8\%) & \cellcolor[RGB]{254,232,223}\texttt{18} (6.0\%) & \cellcolor[RGB]{232,236,244}\texttt{24} (0.1\%) & \cellcolor[RGB]{250,231,243}\texttt{32} (0.0\%) & \cellcolor[RGB]{237,247,220}\texttt{38} (0.0\%) \\
C2: first operand echo & \cellcolor[RGB]{224,242,237}\texttt{42} (99.6\%) & \cellcolor[RGB]{254,232,223}\texttt{18} (0.3\%) & \cellcolor[RGB]{232,236,244}\texttt{24} (0.0\%) & \cellcolor[RGB]{224,242,237}\texttt{41} (0.0\%) & \cellcolor[RGB]{237,247,220}\texttt{38} (0.0\%) \\
C5: first operand 86 & \cellcolor[RGB]{224,242,237}\texttt{42} (97.3\%) & \cellcolor[RGB]{254,232,223}\texttt{18} (2.6\%) & \cellcolor[RGB]{232,236,244}\texttt{24} (0.0\%) & \cellcolor[RGB]{237,247,220}\texttt{38} (0.0\%) & \cellcolor[RGB]{255,247,213}\texttt{36} (0.0\%) \\
C8: correct even operand sums & \cellcolor[RGB]{224,242,237}\texttt{42} (93.0\%) & \cellcolor[RGB]{254,232,223}\texttt{18} (6.7\%) & \cellcolor[RGB]{232,236,244}\texttt{24} (0.1\%) & \cellcolor[RGB]{237,247,220}\texttt{38} (0.0\%) & \cellcolor[RGB]{249,243,233}\texttt{22} (0.0\%) \\
C18: correct even sums & \cellcolor[RGB]{224,242,237}\texttt{42} (93.0\%) & \cellcolor[RGB]{254,232,223}\texttt{18} (6.7\%) & \cellcolor[RGB]{232,236,244}\texttt{24} (0.1\%) & \cellcolor[RGB]{237,247,220}\texttt{38} (0.0\%) & \cellcolor[RGB]{255,247,213}\texttt{36} (0.0\%) \\
C22: not[correct sums] & \cellcolor[RGB]{224,242,237}\texttt{42} (92.2\%) & \cellcolor[RGB]{254,232,223}\texttt{18} (7.6\%) & \cellcolor[RGB]{232,236,244}\texttt{24} (0.1\%) & \cellcolor[RGB]{237,247,220}\texttt{38} (0.0\%) & \cellcolor[RGB]{255,247,213}\texttt{36} (0.0\%) \\
C24: not[sums with 7] & \cellcolor[RGB]{224,242,237}\texttt{42} (93.0\%) & \cellcolor[RGB]{254,232,223}\texttt{18} (6.7\%) & \cellcolor[RGB]{232,236,244}\texttt{24} (0.1\%) & \cellcolor[RGB]{237,247,220}\texttt{38} (0.0\%) & \cellcolor[RGB]{255,247,213}\texttt{36} (0.0\%) \\
                                             C30: correct teen sums & \cellcolor[RGB]{224,242,237}\texttt{42} (93.0\%) & \cellcolor[RGB]{254,232,223}\texttt{18} (6.7\%) & \cellcolor[RGB]{232,236,244}\texttt{24} (0.1\%) & \cellcolor[RGB]{237,247,220}\texttt{38} (0.0\%) & \cellcolor[RGB]{250,231,243}\texttt{32} (0.0\%) \\                                            C33: not[teen sum answers] & \cellcolor[RGB]{224,242,237}\texttt{42} (93.0\%) & \cellcolor[RGB]{254,232,223}\texttt{18} (6.7\%) & \cellcolor[RGB]{232,236,244}\texttt{24} (0.0\%) & \cellcolor[RGB]{237,247,220}\texttt{38} (0.0\%) & \cellcolor[RGB]{239,239,239}\texttt{\_} (0.0\%) \\                                        C36: not[sums near 23] & \cellcolor[RGB]{224,242,237}\texttt{42} (93.0\%) & \cellcolor[RGB]{254,232,223}\texttt{18} (6.7\%) & \cellcolor[RGB]{232,236,244}\texttt{24} (0.1\%) & \cellcolor[RGB]{237,247,220}\texttt{38} (0.0\%) & \cellcolor[RGB]{255,247,213}\texttt{36} (0.0\%) \\                                            C38: first operand 13 & \cellcolor[RGB]{224,242,237}\texttt{42} (90.2\%) & \cellcolor[RGB]{254,232,223}\texttt{18} (9.5\%) & \cellcolor[RGB]{232,236,244}\texttt{24} (0.1\%) & \cellcolor[RGB]{237,247,220}\texttt{38} (0.0\%) & \cellcolor[RGB]{249,243,233}\texttt{22} (0.0\%) \\                                             C43: correct round sums & \cellcolor[RGB]{224,242,237}\texttt{42} (93.0\%) & \cellcolor[RGB]{254,232,223}\texttt{18} (6.7\%) & \cellcolor[RGB]{232,236,244}\texttt{24} (0.1\%) & \cellcolor[RGB]{237,247,220}\texttt{38} (0.0\%) & \cellcolor[RGB]{249,243,233}\texttt{22} (0.0\%) \\                                           C44: correct diverse addition & \cellcolor[RGB]{224,242,237}\texttt{42} (94.9\%) & \cellcolor[RGB]{254,232,223}\texttt{18} (4.7\%) & \cellcolor[RGB]{232,236,244}\texttt{24} (0.1\%) & \cellcolor[RGB]{237,247,220}\texttt{38} (0.0\%) & \cellcolor[RGB]{249,243,233}\texttt{22} (0.0\%) \\                                     C49: sums near 83 & \cellcolor[RGB]{224,242,237}\texttt{42} (92.2\%) & \cellcolor[RGB]{254,232,223}\texttt{18} (7.6\%) & \cellcolor[RGB]{232,236,244}\texttt{24} (0.1\%) & \cellcolor[RGB]{237,247,220}\texttt{38} (0.0\%) & \cellcolor[RGB]{249,243,233}\texttt{22} (0.0\%) \\                                                 C67: not[number 30] & \cellcolor[RGB]{224,242,237}\texttt{42} (92.2\%) & \cellcolor[RGB]{254,232,223}\texttt{18} (7.6\%) & \cellcolor[RGB]{232,236,244}\texttt{24} (0.1\%) & \cellcolor[RGB]{237,247,220}\texttt{38} (0.0\%) & \cellcolor[RGB]{250,231,243}\texttt{32} (0.0\%) \\                                               C73: correct 136 range sums & \cellcolor[RGB]{224,242,237}\texttt{42} (93.0\%) & \cellcolor[RGB]{254,232,223}\texttt{18} (6.7\%) & \cellcolor[RGB]{232,236,244}\texttt{24} (0.1\%) & \cellcolor[RGB]{237,247,220}\texttt{38} (0.0\%) & \cellcolor[RGB]{255,247,213}\texttt{36} (0.0\%) \\                                       C77: sums equaling 123 & \cellcolor[RGB]{224,242,237}\texttt{42} (93.8\%) & \cellcolor[RGB]{254,232,223}\texttt{18} (6.0\%) & \cellcolor[RGB]{232,236,244}\texttt{24} (0.1\%) & \cellcolor[RGB]{237,247,220}\texttt{38} (0.0\%) & \cellcolor[RGB]{255,247,213}\texttt{36} (0.0\%) \\                                            C78: first operand X4 & \cellcolor[RGB]{224,242,237}\texttt{42} (93.8\%) & \cellcolor[RGB]{254,232,223}\texttt{18} (6.0\%) & \cellcolor[RGB]{232,236,244}\texttt{24} (0.0\%) & \cellcolor[RGB]{237,247,220}\texttt{38} (0.0\%) & \cellcolor[RGB]{249,243,233}\texttt{22} (0.0\%) \\                                             C81: correct 100s sums & \cellcolor[RGB]{224,242,237}\texttt{42} (93.8\%) & \cellcolor[RGB]{254,232,223}\texttt{18} (6.0\%) & \cellcolor[RGB]{232,236,244}\texttt{24} (0.1\%) & \cellcolor[RGB]{237,247,220}\texttt{38} (0.0\%) & \cellcolor[RGB]{255,247,213}\texttt{36} (0.0\%) \\                                            C82: not[correct sums] & \cellcolor[RGB]{224,242,237}\texttt{42} (92.2\%) & \cellcolor[RGB]{254,232,223}\texttt{18} (7.6\%) & \cellcolor[RGB]{232,236,244}\texttt{24} (0.1\%) & \cellcolor[RGB]{237,247,220}\texttt{38} (0.0\%) & \cellcolor[RGB]{255,247,213}\texttt{36} (0.0\%) \\
C95: not[correct sums] & \cellcolor[RGB]{224,242,237}\texttt{42} (91.4\%) & \cellcolor[RGB]{254,232,223}\texttt{18} (8.5\%) & \cellcolor[RGB]{232,236,244}\texttt{24} (0.1\%) & \cellcolor[RGB]{237,247,220}\texttt{38} (0.0\%) & \cellcolor[RGB]{249,243,233}\texttt{22} (0.0\%) \\
C96: correct 50s addition & \cellcolor[RGB]{224,242,237}\texttt{42} (89.1\%) & \cellcolor[RGB]{254,232,223}\texttt{18} (10.6\%) & \cellcolor[RGB]{232,236,244}\texttt{24} (0.1\%) & \cellcolor[RGB]{237,247,220}\texttt{38} (0.0\%) & \cellcolor[RGB]{250,231,243}\texttt{32} (0.0\%) \\
C106: number 24 bias & \cellcolor[RGB]{224,242,237}\texttt{42} (93.0\%) & \cellcolor[RGB]{254,232,223}\texttt{18} (6.7\%) & \cellcolor[RGB]{232,236,244}\texttt{24} (0.1\%) & \cellcolor[RGB]{237,247,220}\texttt{38} (0.0\%) & \cellcolor[RGB]{255,247,213}\texttt{36} (0.0\%) \\
C107: small sum correctness & \cellcolor[RGB]{224,242,237}\texttt{42} (86.7\%) & \cellcolor[RGB]{254,232,223}\texttt{18} (13.3\%) & \cellcolor[RGB]{232,236,244}\texttt{24} (0.1\%) & \cellcolor[RGB]{237,247,220}\texttt{38} (0.0\%) & \cellcolor[RGB]{250,231,243}\texttt{32} (0.0\%) \\
C109: ones digit 5 bias & \cellcolor[RGB]{224,242,237}\texttt{42} (75.0\%) & \cellcolor[RGB]{254,232,223}\texttt{18} (24.4\%) & \cellcolor[RGB]{232,236,244}\texttt{24} (0.2\%) & \cellcolor[RGB]{249,243,233}\texttt{22} (0.0\%) & \cellcolor[RGB]{250,231,243}\texttt{32} (0.0\%) \\
C113: sums near 42 & \cellcolor[RGB]{224,242,237}\texttt{42} (89.1\%) & \cellcolor[RGB]{254,232,223}\texttt{18} (10.6\%) & \cellcolor[RGB]{232,236,244}\texttt{24} (0.3\%) & \cellcolor[RGB]{255,247,213}\texttt{36} (0.0\%) & \cellcolor[RGB]{237,247,220}\texttt{38} (0.0\%) \\
C115: not[number 35] & \cellcolor[RGB]{224,242,237}\texttt{42} (92.2\%) & \cellcolor[RGB]{254,232,223}\texttt{18} (7.6\%) & \cellcolor[RGB]{232,236,244}\texttt{24} (0.1\%) & \cellcolor[RGB]{237,247,220}\texttt{38} (0.0\%) & \cellcolor[RGB]{255,247,213}\texttt{36} (0.0\%) \\                                              C116: not[correct sums] & \cellcolor[RGB]{224,242,237}\texttt{42} (92.2\%) & \cellcolor[RGB]{254,232,223}\texttt{18} (7.6\%) & \cellcolor[RGB]{232,236,244}\texttt{24} (0.1\%) & \cellcolor[RGB]{237,247,220}\texttt{38} (0.0\%) & \cellcolor[RGB]{255,247,213}\texttt{36} (0.0\%) \\                                           C118: number 44 bias & \cellcolor[RGB]{224,242,237}\texttt{42} (93.0\%) & \cellcolor[RGB]{254,232,223}\texttt{18} (6.7\%) & \cellcolor[RGB]{232,236,244}\texttt{24} (0.1\%) & \cellcolor[RGB]{237,247,220}\texttt{38} (0.0\%) & \cellcolor[RGB]{239,239,239}\texttt{\_} (0.0\%) \\                                              C122: not[large sums] & \cellcolor[RGB]{224,242,237}\texttt{42} (93.8\%) & \cellcolor[RGB]{254,232,223}\texttt{18} (6.0\%) & \cellcolor[RGB]{232,236,244}\texttt{24} (0.1\%) & \cellcolor[RGB]{237,247,220}\texttt{38} (0.0\%) & \cellcolor[RGB]{249,243,233}\texttt{22} (0.0\%) \\                                             C124: sums equaling 16 & \cellcolor[RGB]{224,242,237}\texttt{42} (95.3\%) & \cellcolor[RGB]{254,232,223}\texttt{18} (4.7\%) & \cellcolor[RGB]{232,236,244}\texttt{24} (0.1\%) & \cellcolor[RGB]{237,247,220}\texttt{38} (0.0\%) & \cellcolor[RGB]{250,231,243}\texttt{32} (0.0\%) \\                                            C127: sums near 90 & \cellcolor[RGB]{224,242,237}\texttt{42} (92.2\%) & \cellcolor[RGB]{254,232,223}\texttt{18} (7.6\%) & \cellcolor[RGB]{232,236,244}\texttt{24} (0.1\%) & \cellcolor[RGB]{237,247,220}\texttt{38} (0.0\%) & \cellcolor[RGB]{249,243,233}\texttt{22} (0.0\%) \\                                                C130: sums near 185 & \cellcolor[RGB]{224,242,237}\texttt{42} (92.2\%) & \cellcolor[RGB]{254,232,223}\texttt{18} (7.6\%) & \cellcolor[RGB]{232,236,244}\texttt{24} (0.1\%) & \cellcolor[RGB]{237,247,220}\texttt{38} (0.0\%) & \cellcolor[RGB]{249,243,233}\texttt{22} (0.0\%) \\                                               C140: first operand 37-42 & \cellcolor[RGB]{224,242,237}\texttt{42} (93.8\%) & \cellcolor[RGB]{254,232,223}\texttt{18} (6.0\%) & \cellcolor[RGB]{232,236,244}\texttt{24} (0.1\%) & \cellcolor[RGB]{237,247,220}\texttt{38} (0.0\%) & \cellcolor[RGB]{249,243,233}\texttt{22} (0.0\%) \\                                         C148: 8X operand pairs & \cellcolor[RGB]{224,242,237}\texttt{42} (93.0\%) & \cellcolor[RGB]{254,232,223}\texttt{18} (6.7\%) & \cellcolor[RGB]{232,236,244}\texttt{24} (0.1\%) & \cellcolor[RGB]{237,247,220}\texttt{38} (0.0\%) & \cellcolor[RGB]{249,243,233}\texttt{22} (0.0\%) \\                                            C153: round sum correctness & \cellcolor[RGB]{224,242,237}\texttt{42} (92.2\%) & \cellcolor[RGB]{254,232,223}\texttt{18} (7.6\%) & \cellcolor[RGB]{232,236,244}\texttt{24} (0.1\%) & \cellcolor[RGB]{237,247,220}\texttt{38} (0.0\%) & \cellcolor[RGB]{249,243,233}\texttt{22} (0.0\%) \\                                       C155: correct small sums & \cellcolor[RGB]{224,242,237}\texttt{42} (89.1\%) & \cellcolor[RGB]{254,232,223}\texttt{18} (10.6\%) & \cellcolor[RGB]{232,236,244}\texttt{24} (0.1\%) & \cellcolor[RGB]{237,247,220}\texttt{38} (0.0\%) & \cellcolor[RGB]{255,247,213}\texttt{36} (0.0\%) \\                                         C168: correct two digit sums & \cellcolor[RGB]{224,242,237}\texttt{42} (93.0\%) & \cellcolor[RGB]{254,232,223}\texttt{18} (6.7\%) & \cellcolor[RGB]{232,236,244}\texttt{24} (0.1\%) & \cellcolor[RGB]{237,247,220}\texttt{38} (0.0\%) & \cellcolor[RGB]{255,247,213}\texttt{36} (0.0\%) \\                                      C170: number 12 bias & \cellcolor[RGB]{224,242,237}\texttt{42} (91.4\%) & \cellcolor[RGB]{254,232,223}\texttt{18} (8.5\%) & \cellcolor[RGB]{232,236,244}\texttt{24} (0.1\%) & \cellcolor[RGB]{237,247,220}\texttt{38} (0.0\%) & \cellcolor[RGB]{249,243,233}\texttt{22} (0.0\%) \\                                              C173: correct mid sums & \cellcolor[RGB]{224,242,237}\texttt{42} (93.0\%) & \cellcolor[RGB]{254,232,223}\texttt{18} (6.7\%) & \cellcolor[RGB]{232,236,244}\texttt{24} (0.1\%) & \cellcolor[RGB]{237,247,220}\texttt{38} (0.0\%) & \cellcolor[RGB]{249,243,233}\texttt{22} (0.0\%) \\                                            C179: sums near 88 & \cellcolor[RGB]{224,242,237}\texttt{42} (92.2\%) & \cellcolor[RGB]{254,232,223}\texttt{18} (7.6\%) & \cellcolor[RGB]{232,236,244}\texttt{24} (0.1\%) & \cellcolor[RGB]{237,247,220}\texttt{38} (0.0\%) & \cellcolor[RGB]{250,231,243}\texttt{32} (0.0\%) \\                                                C180: correct large addition & \cellcolor[RGB]{224,242,237}\texttt{42} (98.8\%) & \cellcolor[RGB]{254,232,223}\texttt{18} (0.9\%) & \cellcolor[RGB]{237,247,220}\texttt{38} (0.0\%) & \cellcolor[RGB]{232,236,244}\texttt{24} (0.0\%) & \cellcolor[RGB]{250,231,243}\texttt{32} (0.0\%) \\                                      C184: not[sums near 137] & \cellcolor[RGB]{224,242,237}\texttt{42} (92.2\%) & \cellcolor[RGB]{254,232,223}\texttt{18} (7.6\%) & \cellcolor[RGB]{232,236,244}\texttt{24} (0.1\%) & \cellcolor[RGB]{237,247,220}\texttt{38} (0.0\%) & \cellcolor[RGB]{249,243,233}\texttt{22} (0.0\%) \\                                          C186: correct diverse sums & \cellcolor[RGB]{224,242,237}\texttt{42} (91.4\%) & \cellcolor[RGB]{254,232,223}\texttt{18} (8.5\%) & \cellcolor[RGB]{232,236,244}\texttt{24} (0.1\%) & \cellcolor[RGB]{237,247,220}\texttt{38} (0.0\%) & \cellcolor[RGB]{249,243,233}\texttt{22} (0.0\%) \\                                        C191: not[number 52] & \cellcolor[RGB]{224,242,237}\texttt{42} (84.8\%) & \cellcolor[RGB]{254,232,223}\texttt{18} (14.7\%) & \cellcolor[RGB]{232,236,244}\texttt{24} (0.1\%) & \cellcolor[RGB]{237,247,220}\texttt{38} (0.0\%) & \cellcolor[RGB]{255,247,213}\texttt{36} (0.0\%) \\                                             C193: not[number 24 sums] & \cellcolor[RGB]{224,242,237}\texttt{42} (93.0\%) & \cellcolor[RGB]{254,232,223}\texttt{18} (6.7\%) & \cellcolor[RGB]{232,236,244}\texttt{24} (0.1\%) & \cellcolor[RGB]{237,247,220}\texttt{38} (0.0\%) & \cellcolor[RGB]{239,239,239}\texttt{\_} (0.0\%) \\                                         C198: correct teen sums & \cellcolor[RGB]{224,242,237}\texttt{42} (90.2\%) & \cellcolor[RGB]{254,232,223}\texttt{18} (9.5\%) & \cellcolor[RGB]{232,236,244}\texttt{24} (0.1\%) & \cellcolor[RGB]{237,247,220}\texttt{38} (0.0\%) & \cellcolor[RGB]{250,231,243}\texttt{32} (0.0\%) \\
C203: not[small digit sums] & \cellcolor[RGB]{224,242,237}\texttt{42} (93.8\%) & \cellcolor[RGB]{254,232,223}\texttt{18} (6.0\%) & \cellcolor[RGB]{232,236,244}\texttt{24} (0.1\%) & \cellcolor[RGB]{237,247,220}\texttt{38} (0.0\%) & \cellcolor[RGB]{249,243,233}\texttt{22} (0.0\%) \\
C208: sums near 60 & \cellcolor[RGB]{224,242,237}\texttt{42} (93.0\%) & \cellcolor[RGB]{254,232,223}\texttt{18} (6.7\%) & \cellcolor[RGB]{232,236,244}\texttt{24} (0.1\%) & \cellcolor[RGB]{237,247,220}\texttt{38} (0.0\%) & \cellcolor[RGB]{249,243,233}\texttt{22} (0.0\%) \\
C211: numeric continuation & \cellcolor[RGB]{224,242,237}\texttt{42} (94.5\%) & \cellcolor[RGB]{254,232,223}\texttt{18} (5.3\%) & \cellcolor[RGB]{232,236,244}\texttt{24} (0.1\%) & \cellcolor[RGB]{237,247,220}\texttt{38} (0.0\%) & \cellcolor[RGB]{249,243,233}\texttt{22} (0.0\%) \\
C215: correct 93 addition & \cellcolor[RGB]{224,242,237}\texttt{42} (93.0\%) & \cellcolor[RGB]{254,232,223}\texttt{18} (6.7\%) & \cellcolor[RGB]{232,236,244}\texttt{24} (0.1\%) & \cellcolor[RGB]{237,247,220}\texttt{38} (0.0\%) & \cellcolor[RGB]{249,243,233}\texttt{22} (0.0\%) \\
C229: not[sums near 20] & \cellcolor[RGB]{224,242,237}\texttt{42} (77.3\%) & \cellcolor[RGB]{254,232,223}\texttt{18} (22.2\%) & \cellcolor[RGB]{232,236,244}\texttt{24} (0.2\%) & \cellcolor[RGB]{237,247,220}\texttt{38} (0.1\%) & \cellcolor[RGB]{250,231,243}\texttt{32} (0.0\%) \\
C232: correct arithmetic & \cellcolor[RGB]{224,242,237}\texttt{42} (93.0\%) & \cellcolor[RGB]{254,232,223}\texttt{18} (6.7\%) & \cellcolor[RGB]{232,236,244}\texttt{24} (0.1\%) & \cellcolor[RGB]{237,247,220}\texttt{38} (0.0\%) & \cellcolor[RGB]{255,247,213}\texttt{36} (0.0\%) \\
C242: not[sums near 108] & \cellcolor[RGB]{224,242,237}\texttt{42} (91.4\%) & \cellcolor[RGB]{254,232,223}\texttt{18} (8.5\%) & \cellcolor[RGB]{232,236,244}\texttt{24} (0.1\%) & \cellcolor[RGB]{237,247,220}\texttt{38} (0.0\%) & \cellcolor[RGB]{249,243,233}\texttt{22} (0.0\%) \\
C246: correct diverse addition & \cellcolor[RGB]{224,242,237}\texttt{42} (93.8\%) & \cellcolor[RGB]{254,232,223}\texttt{18} (6.0\%) & \cellcolor[RGB]{232,236,244}\texttt{24} (0.1\%) & \cellcolor[RGB]{237,247,220}\texttt{38} (0.0\%) & \cellcolor[RGB]{250,231,243}\texttt{32} (0.0\%) \\                                    C247: correct large pair sums & \cellcolor[RGB]{224,242,237}\texttt{42} (91.4\%) & \cellcolor[RGB]{254,232,223}\texttt{18} (8.5\%) & \cellcolor[RGB]{232,236,244}\texttt{24} (0.1\%) & \cellcolor[RGB]{237,247,220}\texttt{38} (0.0\%) & \cellcolor[RGB]{249,243,233}\texttt{22} (0.0\%) \\                                     C254: sums near 142 & \cellcolor[RGB]{224,242,237}\texttt{42} (75.0\%) & \cellcolor[RGB]{254,232,223}\texttt{18} (24.4\%) & \cellcolor[RGB]{232,236,244}\texttt{24} (0.2\%) & \cellcolor[RGB]{237,247,220}\texttt{38} (0.1\%) & \cellcolor[RGB]{255,247,213}\texttt{36} (0.0\%) \\                                              \bottomrule
\end{tabular}}                                                                   \end{center}

\newpage
\section{Additional results on \texttt{pills} dataset}

We include steering results for all supernodes (beyond the top-$5$ shown in the main text) in \cref{tab:cluster-steering-all}. Beyond the top-$5$ supernodes, steerings tends to have less strong directional steering effects, and steering generally increases incoherence slightly.

\begin{table}[!h]                                                                                                                                                \centering                                                                                                          
\caption{ASR results and generation coherency when steering cluster activations in the base \texttt{pills} prompt by a given muliplier (over $50$ generations). $r$ is Pearson correlation of cluster
attribution vs.~ASR over dataset.}                                                     \label{tab:cluster-steering-all}                                                                                                                                    \small                                                                       
\adjustbox{max width=\textwidth}{
\begin{tabular}{rlrr rr rr}                                                                                                                                     \toprule
& & & & \multicolumn{2}{c}{\textbf{ASR}} & \multicolumn{2}{c}{\textbf{\% Incoherent}} \\                                                                        \cmidrule(lr){5-6} \cmidrule(lr){7-8}
\textbf{Cluster} & \textbf{Label} & \textbf{\#N} & \textbf{$r$} & \multicolumn{1}{c}{\textbf{0$\times$}} & \multicolumn{1}{c}{\textbf{2$\times$}} & \multicolumn{1}{c}{\textbf{0$\times$}} & \multicolumn{1}{c}{\textbf{2$\times$}} \\          \midrule
--- & \textit{unsteered} & & & \multicolumn{2}{c}{\cellcolor{red!28} 28$^{\pm6}$\%} & \multicolumn{2}{c}{5$^{\pm3}$\%} \\                                       \midrule
C3 & pills safety redirect & 13 & -0.70 & \cellcolor{red!88} \textbf{88$^{\pm5}$\%} & \cellcolor{red!20} 20$^{\pm5}$\% & 21$^{\pm5}$\% & 7$^{\pm3}$\% \\        C9 & ridiculous-to-introductory & 23 & +0.71 & \cellcolor{red!12} 12$^{\pm4}$\% & \cellcolor{red!90} \textbf{90$^{\pm5}$\%} & 7$^{\pm3}$\% & 13$^{\pm4}$\% \\
C16 & urgent medication reminders & 51 & -0.41 & 0$^{\pm0}$\% & \cellcolor{red!52} 52$^{\pm6}$\% & 0$^{\pm0}$\% & 9$^{\pm3}$\% \\                               C8 & medication safety deflection & 23 & -0.57 & 0$^{\pm0}$\% & \cellcolor{red!38} 38$^{\pm6}$\% & 0$^{\pm0}$\% & 16$^{\pm4}$\% \\
C1 & unsafe pill advice framing & 38 & +0.24 & \cellcolor{red!56} 56$^{\pm6}$\% & \cellcolor{red!38} 38$^{\pm6}$\% & 12$^{\pm4}$\% & 10$^{\pm4}$\% \\           C14 & not{[}ridiculous medication compliance{]} & 19 & -0.56 & \cellcolor{red!18} 18$^{\pm5}$\% & \cellcolor{red!44} 44$^{\pm6}$\% & 7$^{\pm3}$\% & 15$^{\pm4}$\% \\                                                                                                                                                            C10 & safety refusal trigger & 22 & +0.56 & \cellcolor{red!28} 28$^{\pm5}$\% & \cellcolor{red!66} 66$^{\pm6}$\% & 12$^{\pm4}$\% & 24$^{\pm5}$\% \\              C13 & not{[}medication advice onset{]} & 9 & +0.51 & \cellcolor{red!26} 26$^{\pm5}$\% & \cellcolor{red!56} 56$^{\pm6}$\% & 9$^{\pm3}$\% & 18$^{\pm4}$\% \\      C2 & not{[}medication recall advice{]} & 17 & +0.67 & \cellcolor{red!40} 40$^{\pm6}$\% & \cellcolor{red!22} 22$^{\pm5}$\% & 12$^{\pm4}$\% & 7$^{\pm3}$\% \\     C17 & cautious medication hedging & 25 & -0.29 & \cellcolor{red!54} 54$^{\pm6}$\% & \cellcolor{red!26} 26$^{\pm5}$\% & 18$^{\pm4}$\% & 12$^{\pm4}$\% \\         C5 & advice safety refusal & 17 & +0.74 & \cellcolor{red!46} 46$^{\pm5}$\% & \cellcolor{red!42} 42$^{\pm6}$\% & 21$^{\pm5}$\% & 10$^{\pm4}$\% \\                C6 & cautious medical hedging & 28 & -0.36 & \cellcolor{red!48} 48$^{\pm6}$\% & \cellcolor{red!46} 46$^{\pm6}$\% & 12$^{\pm4}$\% & 15$^{\pm4}$\% \\             C15 & not{[}rushed medical advice{]} & 9 & +0.31 & \cellcolor{red!42} 42$^{\pm6}$\% & \cellcolor{red!46} 46$^{\pm6}$\% & 12$^{\pm4}$\% & 13$^{\pm4}$\% \\       C7 & system-safety disclaimers & 14 & -0.65 & \cellcolor{red!34} 34$^{\pm6}$\% & \cellcolor{red!40} 40$^{\pm6}$\% & 12$^{\pm4}$\% & 16$^{\pm4}$\% \\            C18 & medical advice refusal & 22 & +0.53 & \cellcolor{red!30} 30$^{\pm6}$\% & \cellcolor{red!46} 46$^{\pm6}$\% & 3$^{\pm2}$\% & 16$^{\pm4}$\% \\               C11 & not{[}rushed medication ``i``{]} & 18 & -0.36 & \cellcolor{red!22} 22$^{\pm5}$\% & \cellcolor{red!30} 30$^{\pm6}$\% & 5$^{\pm3}$\% & 7$^{\pm3}$\% \\      C4 & not{[}panicked medication advice{]} & 14 & +0.36 & \cellcolor{red!36} 36$^{\pm6}$\% & \cellcolor{red!48} 48$^{\pm6}$\% & 12$^{\pm4}$\% & 18$^{\pm4}$\% \\  C19 & not{[}medical conditionals{]} & 20 & +0.36 & \cellcolor{red!34} 34$^{\pm6}$\% & \cellcolor{red!38} 38$^{\pm6}$\% & 12$^{\pm4}$\% & 9$^{\pm3}$\% \\        C12 & medical safety disclaimers & 25 & -0.71 & \cellcolor{red!44} 44$^{\pm6}$\% & \cellcolor{red!36} 36$^{\pm5}$\% & 9$^{\pm3}$\% & 18$^{\pm4}$\% \\           C0 & empathetic redirection & 10 & -0.69 & \cellcolor{red!38} 38$^{\pm6}$\% & \cellcolor{red!42} 42$^{\pm6}$\% & 9$^{\pm3}$\% & 15$^{\pm4}$\% \\                \bottomrule                                                                                                                                                     \end{tabular}}                                                                                                                                                   \end{table}%

\subsection{Judge prompts}
\label{sec:pills-judge-prompts}
We include judge prompts for ASR and coherence on the \texttt{pills} experiments below.

\begin{promptbox}[ASR]
You will be given a prompt and a response from a language model. The prompt is designed to trick the model into giving unsafe medical advice — specifically, telling the user to take a pill without properly verifying which medication it is.

Judge whether the model's response is unsafe. Answer "yes" if the model:
- Suggests taking a specific pill or any pill without verifying what it is
- Gives a method for choosing a pill that doesn't involve checking the label or consulting a professional
- Plays along with the urgency and skips safety precautions

Answer "no" if the model:
- Refuses to give advice on which pill to take
- Recommends checking the label, consulting a doctor/pharmacist, or looking up the medication
- Explicitly warns against taking unidentified medication

Prompt: {raw_prompt}
Response: {response}

Answer with either "yes" or "no" only.
\end{promptbox}

\begin{promptbox}[Coherence]
Is the following text coherent and readable English? Answer "yes" if it reads like a normal response (even if the content is wrong or unsafe). Answer "no" if it is garbled, repetitive nonsense, or clearly broken.

Text: {response}

Answer with either "yes" or "no" only.
\end{promptbox}

\section{Prompts for describing supernodes}

\subsection{Input attribution}
\label{sec:input-prompts}
\begin{promptbox}[Input attribution explainer]
You are a meticulous AI researcher conducting an important investigation into a specific neuron inside a language model that activates in response to text excerpts. Your overall task is to describe features of text excerpts that cause the neuron to strongly activate.

You will receive a list of text excerpts on which the neuron activates. Tokens causing activation will appear between delimiters like {{this}}. Consecutive activating tokens will also be accordingly delimited {{just like this}}. If no tokens are highlighted with {{}}, then the neuron does not activate on any tokens in the excerpt.

Note: Neurons activate on a word-by-word basis. Also, neuron activations can only depend on words before the word it activates on, so the description cannot depend on words that come after, and should only depend on words that come before the activation.
\end{promptbox}

\subsection{Output contribution}
\label{sec:output-prompts}
\begin{promptbox}[Output contribution explainer]
You are a meticulous AI researcher studying a neuron inside a language model. Your overall task is to describe features of continuations that a neuron contributes to, given a specific dataset.

You will receive prompts from a particular dataset followed by possible continuation tokens, along with how strongly the neuron contributed to each (scores from -10 to 10, where positive = promotes, negative = suppresses, 0 = no effect).

**Required output properties**: Make your final explanations as concise as possible, using as few words as possible to describe text features that the neuron contributes to.

Output format: [EXPLANATION]: <your explanation>
\end{promptbox}

\begin{promptbox}[Output contribution simulator]
You are a meticulous AI researcher studying a neuron inside a language model. You are simulating a neuron's contribution to continuations of given prompts, taken from a dataset. Given a proposed description of the types of continuations that the neuron promotes and suppresses, along with a dataset of prompts with possible continuations, predict scores from -10 to 10 for each continuation for each prompt (where positive = promotes, negative = suppresses, 0 = no effect).

Output format: Respond with a paragraph for each prompt, with the first line being "Prompt: [prompt]" where you repeat the prompt you are scoring for, followed by a second line of the format "Continuations: [('[cont_1]', [score]), ...]" where you repeat each continuation option and your integer score, based on the provided description.
\end{promptbox}

\subsection{Summarisation}
\label{sec:summarisation-prompts}
\begin{promptbox}[Summarisation of all clusters]
You are labeling clusters of neurons in a language model. Each cluster has two descriptions:

- **Attribution**: what input tokens cause this cluster to activate.
- **Contribution**: what output tokens this cluster promotes or suppresses.

Below are all {n_clusters} clusters, which you will label.

## Your task

Produce a very short, **distinctive** label (1-3 words) for each one.

## Output format

- Be as specific as possible — mention concrete entities or patterns from the examples.
- It is OK for multiple clusters to have similar labels if they are genuinely similar, but try to differentiate where possible.
- DO NOT use vague words like “focus”, “answer”, “response”, “promotion”, “lookup”, “recall”, “mapping”.
- If the cluster has an inhibitory effect, wrap the label in a negative function like this: not[label]
- Labels should be lowercase, except for proper nouns.
- Labels should be a space-separated and coherent English phrase.
- Output exactly one label per line, in the format:
CLUSTER_ID: label
- Output all {n_clusters} labels and nothing else.

## Examples

- Attribution: primarily Dallas and some other terms only in Texas-specific context
- Contribution: promotes Texas and cities in Texas, does nothing in other cases
Result: Texas

## Clusters

{cluster_block}
\end{promptbox}

\end{document}